\documentclass[twoside]{article}

\usepackage[utf8]{inputenc}
\usepackage{amsfonts}
\usepackage{amsmath,amssymb}
\usepackage{nccmath}
\usepackage{mathrsfs}
\usepackage{graphicx}
\usepackage{subfigure}
\usepackage{tabularx}
\usepackage[all]{xy}
\usepackage{color}
\usepackage{multirow}
\usepackage{booktabs}
\usepackage{enumitem}
\usepackage{hyperref}
\usepackage{algpseudocode}
\usepackage{algorithm}
\usepackage{natbib}

\usepackage{wrapfig}

\newtheorem{proposition}{Proposition}
\newtheorem{lemma}[proposition]{Lemma}
\newtheorem{theorem}[proposition]{Theorem}

\newtheorem{remark}[proposition]{Remark}

\newtheorem{corollary}[proposition]{Corollary}

\newenvironment{proof}[1][Proof\ ]{\medskip\noindent{\bf #1}\ }{%
\hfill $\Box$\par\quad\par}

\usepackage{times}
\usepackage{enumitem}
\usepackage{multirow}
\usepackage{setspace} 

\def\mcl#1{\mathcal{#1}}
\def\bracket#1{\left\langle #1\right\rangle}
\def\hil{\mcl{H}}
\def\nn{\nonumber}
\def\opn{\operatorname}
\def\mr{\mathrm}

\def\Bbracket#1{\bigg\langle #1\bigg\rangle}

\def\red#1{\textcolor{black}{#1}}

\newenvironment{mythm}[1][]{\medskip\par\noindent{\bfseries #1}\ \,\,\em}{\medskip\par}

\DeclareSymbolFont{EulerExtension}{U}{euex}{m}{n}
\DeclareMathSymbol{\euintop}{\mathop} {EulerExtension}{"52}
\DeclareMathSymbol{\euointop}{\mathop} {EulerExtension}{"48}

\begin{document}

\title{Deep Koopman-layered Model with Universal Property \\Based on Toeplitz Matrices}

\author{Yuka Hashimoto$^{1,2}$\quad Tomoharu Iwata$^{1}$\medskip\\
{\normalsize 1. NTT Corporation}\\
{\normalsize 2. Center for Advanced Intelligence Project, RIKEN}}

\date{}

\maketitle
\begin{abstract}
We propose deep Koopman-layered models with learnable parameters in the form of Toeplitz matrices for analyzing the transition of the dynamics of time-series data.
The proposed model has both theoretical solidness and flexibility.
By virtue of the universal property of Toeplitz
matrices and the reproducing property underlying the model, we show its universality and generalization property.
In addition, the flexibility of the proposed model enables the model to fit time-series data coming from nonautonomous dynamical systems.
When training the model, we apply Krylov subspace methods for efficient computations, which establish a new connection between Koopman operators and numerical linear algebra.
\red{We also empirically demonstrate that the proposed model outperforms existing methods on eigenvalue estimation of multiple Koopman operators for nonautonomous systems.}
\end{abstract}

\section{Introduction}\label{sec:intro}
Koopman operator has been one of the important tools in machine learning~\cite{kawahara16,ishikawa18,lusch17,Brunton_Kutz19,hashimoto20}.
Koopman operators are linear operators that describe the composition of functions and are applied to analyzing time-series data generated by nonlinear dynamical systems~\citep{koopman31,mezic12,klus17,giannakis20,mezic22}.
For systems with discrete Koopman spectra, by computing the eigenvalues of Koopman operators, we can extract the periodic and decay properties of the underlying dynamical systems.
An important feature of applying Koopman operators is that we can estimate them with given time-series data through fundamental linear algebraic tools such as projection.
A typical approach to estimate Koopman operators is extended dynamic mode decomposition (EDMD)~\citep{williams15}, and 
with EDMD as a starting point, many DMD-based methods are proposed~\citep{kawahara16,colbrook24,schmid22}. 
However, there are two big challenges for DMD-based approaches.
First, we need to choose the dictionary functions to determine the approximation space of the Koopman operator, and what choice of them gives us a better estimation is far from trivial.
Especially, the theoretical analysis regarding the choice of them is challenging.
Second, since we construct the estimation in an analytical way, the model is not flexible enough to incorporate additional information about Koopman operators, such as information coming from other Koopman operators and prior knowledge about the dynamics.

{For autonomous systems, we need to estimate a single Koopman operator.
In this case, \citet{ishikawa24} proposed to choose derivatives of kernel functions as dictionary functions based on the theory of Jet spaces.}
Several works deal with nonautonomous systems.
\citet{macesic18} applied EDMD to estimate a time-dependent Koopman operator for each time window. 
\citet{peitz19} applied EDMD for switching dynamical systems for solving optimal control problems.
However, as far as we know, no existing works show proper choices of dictionary functions for nonautonomous systems based on theoretical analysis \red{such as universality and generalization}.
In addition, in existing approaches, since each Koopman operator for a time window is estimated individually, the model is not flexible enough to take the information of other Koopman operators into account.

To find a proper approximation space and gain the flexibility of the model, neural network-based Koopman methods have been proposed~\citep{lusch17,azencot20,shi22}.
They construct an encoder from the data space to the approximation space for the Koopman operator, and a decoder from the approximation space to the data space.
The encoder and decoder are constructed as deep neural networks, and are trained using data.
Neural network-based Koopman methods for nonautonomous systems have also been proposed.
\citet{liu23} proposed to decompose the Koopman operator into a time-invariant part and a time-variant part.
The time-variant part of the Koopman operator is constructed individually for each time window using EDMD.
\citet{wei24} assumed the ergodicity of the dynamical system and considered time-averaged Koopman operators for nonautonomous dynamical systems.
However, \red{since the approximation space changes over the learning process of the neural networks, their theoretical properties, such as universality and generalization bound, have not been proven.}

In this work, we tackle the above two challenges and propose a framework that estimates multiple Koopman operators {over time} with the Fourier basis approximation space and learnable Toeplitz matrices.
Using our framework, we can estimate multiple Koopman operators simultaneously and can capture the transition of properties of data along time via multiple Koopman operators.
We call each Koopman operator the Koopman layer, and the whole model the deep Koopman-layered model.
The proposed model has both theoretical solidness and flexibility.
We show that the deep Koopman-layered model gives a proper choice of dictionary functions by using the Fourier basis even for nonautonomous dynamical systems.
Here, the proper is in the sense that we can show its theoretical properties such as universality and generalization bound.
In addition, the proposed model has learnable parameters, which makes the model more flexible to fit nonautonomous dynamical systems by incorporating additional information compared to the analytical methods such as EDMD.
The proposed model resolves the issue of theoretical analysis for the neutral network-based methods and that of the flexibility for the analytical methods simultaneously.
\red{The proposed model can be also applied to time-series forecasting by extracting invariant features of the dynamics.}

We show that each Koopman layer is represented by the exponential of a matrix constructed with Toeplitz matrices and diagonal matrices.
This allows us to apply Krylov subspace methods~\citep{saad92,guttel13,hashimoto16} to compute the estimation of Koopman operators {with low computational costs}.
By virtue of the universal property of Toeplitz matrices~\citep{ye16}, we can show the universality of the proposed model with a linear algebraic approach.
We also show a generalization bound of the proposed model using the theory of reproducing kernel Hilbert spaces (RKHSs).


Our contributions are summarized as follows:
\begin{itemize}[nosep,leftmargin=*]
\item We propose a model {for analyzing nonautonomous dynamical systems} that has both theoretical solidness and flexibility.
We show that the proposed model provides a proper choice of dictionary functions, in the sense that we can show the universality and the generalization bound regarding the model.
As for the flexibility, we can learn multiple Koopman operators simultaneously, which enables us to extract the transition of properties of dynamics.
\item We apply Krylov subspace methods to efficiently compute the estimation of Koopman operators.
This establishes a new connection between Koopman operator theoretic approaches and Krylov subspace methods, which opens up future directions for extracting further information about dynamical systems using numerical linear algebraic approaches.
\item 
\red{We empirically demonstrate that the proposed model outperforms existing methods on eigenvalue estimation of multiple Koopman operators and time-series forecasting for nonautonomous systems.}
\end{itemize}

\section{Preliminary}
\paragraph{Notations}
We use a generalized concept of matrices with multi index.
For a finite index set $N\subset \mathbb{Z}^d$ and $a_{j,l}\in\mathbb{C}$ $(j,l\in N)$, we call $A=[a_{j,l}]_{j,l\in N}$ an $N$ by $N$ matrix and denote by $\mathbb{C}^{N\times N}$ the space of all $N$ by $N$ matrices.
We can deal with the generalized matrices in the same way as the standard matrices since we can construct a bijection between them.
In addition, let $\mathcal{X}\subset\mathbb{R}^d$ be a compact space.
We denote by $L^2(\mathcal{X})$ the space of complex-valued square-integrable functions on $\mathcal{X}$, equipped with the Lebesgue measure.

\paragraph{Koopman generator and operator}
Consider an ODE $\frac{\mr{d}x}{\mr{d}t}(t)=f(x(t))$ on $\mathcal{X}$.
Let $g:\mathbb{R}\times \mathcal{X}$ be the flow of the ODE, that is, 
the function $g(\cdot,x)$ is the trajectory of the dynamical system starting at the initial value $x$.
We assume $g$ is continuous and invertible.
We also assume the Jacobian $Jg_t^{-1}$ of $g_t^{-1}$ is bounded for any $t\in\mathbb{R}$, where $g_t=g(t,\cdot)$.
We define the Koopman operator $K^t$ on $L^2(\mathcal{X})$ by the composition with $g(t,\cdot)$ as $K^th(x)=h(g(t,x))$ for $h\in L^2(\mathcal{X})$ and $x\in \mathcal{X}$.
The Koopman operator is a linear operator that maps a function $h$ to the function $h(g(t,\cdot))$.
{Note that the Koopman operator $K^t$ is linear even if $g(t,\cdot)$ is nonlinear.}
Since $K^t$ depends on $t$, we can consider the family of Koopman operators $\{K^t\}_{t\in\mathbb{R}}$.
For $h\in C^1(\mathcal{X})$, where $C^1(\mathcal{X})$ is the space of continuous differentiable functions on $\mathcal{X}$, define a linear operator $L$ by differentiating $K^t$ with respect to $t$ as 
\begin{align*}
    Lh=\lim_{t\to\infty}\frac{K^th-h}{t},
\end{align*}
where the limit is by means of $L^2(\mathcal{X})$.
We call $L$ the Koopman generator.
We write $K^t=\mr{e}^{tL}$.
Note that for the function $h$ defined as $h(t,x)=K^t\tilde{h}(x)$ for $\tilde{h}\in C^1(\mathcal{X})$, we have $\frac{\partial h}{\partial t}=Lh$.
If $L$ is bounded, then it coincides with the standard definition $\mr{e}^{tL}=\sum_{i=1}^{\infty}(tL)^i/i!$.
If $L$ is unbounded, it can be justified by approximating $L$ by a sequence of bounded operators and considering the strong limit of the sequence of the exponential of the bounded operators~\citep{yoshida80}.

\section{Deep Koopman-layered model}\label{sec:deep_koopman}
We propose deep Koopman-layered models based on the Koopman operator theory, which have both theoretical solidness and flexibility.
\subsection{Multiple dynamical systems and Koopman generators}
Consider $J$ ODEs $\frac{\mr{d}x}{\mr{d}t}(t)=f_j(x(t))$ on $\mathcal{X}$ for $j=1,\ldots,J$.
Let $g_j:\mathbb{R}\times \mathcal{X}$ be the flow of the $j$th ODE. 
Let $v\in L^2(\mathcal{X})$ be an auxiliary nonlinear transformation.
Consider the following model:
\begin{align}
G(x)
=v(g_J(t_J,\cdots g_1(t_1,x))).\label{eq:model_flows}
\end{align}
Starting from a point $x$, it is first transformed according to the flow $g_1$, and then $g_2$, and so on.
This model describes a switching dynamical system, and is also regarded as a discrete approximation of a nonautonomous dynamical system.
We set $v$ to adapt the model to the Koopman operator theory.

\begin{remark}\label{rmk:d+1}
Since $v$ is in the complex-valued function space $L^2(\mathcal{X})$, $G$ itself is a complex-valued function.
However, we can easily extend the model to describe only the flow $g_J(t_J,\cdot)\circ \cdots\circ g_1(t_1,\cdot)$, which is a map from $\mathcal{X}$ to $\mathcal{X}$.
Indeed, let $\tilde{g}_j(x,y)=[g_j(t_j,x),y]$ for $x\in\mathcal{X}$ and $y\in [0,d]$.
Let $\tilde{v}$ be a function that satisfies $\tilde{v}(x,k)=x_k$, where $x_k$ is the $k$th element of $x$, and let $G=\tilde{v}\circ \tilde{g}_J\circ \cdots \circ \tilde{g}_1$.
Then, $G(\cdot,k)$ is the $k$th element of $g_J(t_J,\cdot)\circ \cdots\circ g_1(t_1,\cdot)$.
\end{remark}

\begin{remark}\label{rmk:equiv_transform}
In many practical situations, initial values are in a compact domain, and we focus on dynamics in a finite time interval.
In this case, we can regard that the dynamics is in a bounded domain $\mathcal{X}$ in $\mathbb{R}^d$.
In addition, we can reduce the problem on $\mcl{X}$ to that on $\mathbb{T}^d$.
Here, $\mathbb{T}$ is the torus $\mathbb{R}/2\pi\mathbb{Z}$, i.e., the set of real numbers modulo $2\pi$.
The analysis in $\mathbb{T}^d$ enables us to apply the Fourier functions to show the theoretical property of the proposed model.
However, in practical computations, we do not need to identify $\mcl{X}$ or transform the dynamical systems.
See Appendix~\ref{ap:equiv_transform} for more details.
\end{remark}

\subsection{Approximation of Koopman generators using Toeplitz matrices}\label{subsec:approximation_koopman}
Based on Remark~\ref{rmk:equiv_transform}, in the following, we focus on dynamical systems on $\mathbb{T}^d$.
We train the model~\eqref{eq:model_flows} using given time-series data.
For this purpose, we apply the Koopman operator theory. 
Let $L_j$ be the Koopman generator associated with the flow $g_j$.
The Koopman operator $K^{t_j}_j$ of $g_j$ is represented as $\mr{e}^{t_jL_j}$.
Since the model~\eqref{eq:model_flows} is defined using compositions, it is represented as
$G=\mr{e}^{t_1L_1}\cdots \mr{e}^{t_JL_J}v$.

To deal with the Koopman generators defined on the infinite-dimensional space, we approximate them using a finite number of Fourier functions.
For the remaining part of this section, we omit the subscript $j$ for simplicity.
However, in practice, the approximation is computed 
for each layer $j=1,\ldots,J$.
Let $q_n(x)=\mr{e}^{\mr{i}n\cdot x}$ for $n\in\mathbb{Z}^d$ and $x\in\mathbb{T}^d$, where $\mr{i}$ is the imaginary unit.
Let $M_r\subset \mathbb{Z}^d$ be a finite index set for $r=1,\ldots, R$.
We set the $k$th element of the function $f$ in the ODE as 
\begin{align}
\sum_{m_{R}\in M_{R}}a^{k}_{m_{R},R}q_{m_{R}}\cdots \sum_{m_1\in M_1}a^{k}_{m_1,1}q_{m_1}\label{eq:f}
\end{align}
with $a^{k}_{m_r,r}\in\mathbb{C}$,
the product of weighted sums of Fourier functions.
Then, we approximate the Koopman generator $L$ by projecting the input vector onto the finite-dimensional space $V_N:=\opn{Span}\{q_n\,\mid\,n\in N \}$, where $N\subset \mathbb{Z}^d$ is a finite index set, applying $L$, and projecting it back to $V_N$ as $Q_NQ_N^*LQ_NQ_N^*$.
Here, $Q_N:\mathbb{C}^N\to V_N$ is the linear operator defined as $Q_Nc=\sum_{n\in N}c_nq_n$ for $c=(c_n)_{n\in N}\in\mathbb{C}^N$ and $^*$ is the adjoint.
Note that $Q_NQ_N^*$ is the projection onto $V_N$.
Then, the representation matrix $Q_N^*LQ_N$ of the approximated Koopman generator $Q_NQ_N^*LQ_NQ_N^*$ is written as follows.
Throughout the paper, all the proofs are documented in Appendix~\ref{ap:proof}.
\begin{proposition}\label{prop:representation_matrix}
The $(n,l)$-entry of the representation matrix $Q_N^*LQ_N$ 
is
\begin{align}
&\sum_{k=1}^d\sum_{n_{R}-l\in M_{R}}\sum_{n_{R-1}-n_{R}\in M_{R-1}}\cdots \sum_{n_{2}-n_3\in M_{2}}\sum_{n-n_2\in M_{1}}\nn\\
&\qquad\qquad a^{k}_{n_{R}-l,R}a^{k}_{n_{R-1}-n_{R},R-1}\cdots a^{k}_{n_2-n_3,2}a^{k}_{n-n_2,1} \mr{i}l_k,\label{eq:representation_matrix}
\end{align}
where $l_k$ is the $k$th element of the multi index $l\in\mathbb{Z}^d$.
Moreover, we set $n_r=m_{R_j}+\cdots +m_r+l$, thus $n_1=n$, $m_r=n_{r}-n_{r+1}$ for $r=1,\ldots,R-1$, and $m_{R}=n_{R}-l$.
\end{proposition}

Note that since the sum involves the differences of indices, it can be written using Toeplitz matrices, whose $(n,l)$-entry depends only on $n-l$.
We {approximate the sum appearing in Eq.~\eqref{eq:representation_matrix} by} restricting the index $n_r$ to $N$, combining with the information of time $t$, and setting a matrix $\mathbf{L}\in\mathbb{C}^ {N\times N}$ as 
\begin{align}
\mathbf{L}=t\sum_{k=1}^dA_1^{k}\cdots A_{R}^{k} D_k,\label{eq:generator_ap}
\end{align}
where $A_r^{k}$ is the Toeplitz matrix defined as $A_r^{k}=[a_{n-l,r}^{k}]_{n,l\in N}$ and $D_k$ is the diagonal matrix defined as $(D_k)_{l,l}=\mr{i}l_k$ for the multi index $l\in\mathbb{Z}^d$.
We finally regard $Q_N\mathbf{L}Q_N^*$ as an approximation of the Koopman generator $L$.
Then, we construct the approximation $\mathbf{G}$ of $G$ in Eq.~\eqref{eq:model_flows} as follows, which we call the deep Koopman-layered model.
\begin{align}
\mathbf{G}=\mr{e}^{Q_N\mathbf{L}_1Q_N^*}\!\cdots\! \mr{e}^{Q_N\mathbf{L}_JQ_N^*}v
=Q_N\mr{e}^{\mathbf{L}_1}\!\cdots \!\mr{e}^{\mathbf{L}_J}Q_N^*v.\label{eq:koopman_layered}
\end{align}

To compute the product of the matrix exponential $\mr{e}^{\mathbf{L}_j}$ and the vector $\mr{e}^{\mathbf{L}_{j+1}}\cdots \mr{e}^{\mathbf{L}_J}Q_N^*v$, we can use Krylov subspace methods.
If {the number of indices for describing $f$ is smaller than that for describing the whole model}, i.e., $\vert M_r\vert\ll \vert N\vert$, then the Toeplitz matrix $A^{k}_r$ is sparse.
In this case, the matrix-vector product can be computed with the computational cost of $O(\sum_{r=1}^{R}\vert M_r\vert \vert N\vert)$.
Thus, one Krylov iteration costs $O(\sum_{r=1}^{R}\vert M_r\vert \vert N\vert)$, {which makes the computation efficient compared to direct methods without taking the structure of the matrix into account, whose computational cost results in $O(\vert N\vert^3)$.}
We also note that even if the Toeplitz matrices are dense, the computational cost of one iteration of the Krylov subspace method is $O(\vert N\vert \log\vert N\vert)$ if we use the fast Fourier transform.
We numerically show the effect of the number of Krylov iterations on the performance in Appendix~\ref{ap:time_forecast_additional}.

\begin{remark}
To restrict $f$ to be a real-valued map and reduce the number of parameters $a_{m,r}^{k}$, we set $M_r$ as $\{-m_{1,r},\ldots,m_{1,r}\}\times \cdots\times \{-m_{d,r},\ldots,m_{d,r}\}$ for $m_{k,r}\in\mathbb{N}$ for $k=1,\ldots,d$.
In addition, we set $a_{m,r}^{k}=\overline{a_{-m,r}^{k}}$ for $m\in M_r$.
Then, we have $a_{m,r}^{k}q_m=\overline{a_{-m,r}^{k}q_{-m}}$, and $f$ is real-valued.
\end{remark}

\begin{remark}
An advantage of applying Koopman operators is that their spectra describe the properties of dynamical systems.
For example, if the dynamical system is measure preserving, then the corresponding Koopman operator is unitary.
Since each Koopman layer is an estimation of the Koopman operator, we can analyze time-series data coming from nonsutonomous dynamical systems by computing the eigenvalues of the Koopman layers.
We will observe the eigenvalues of Koopman layers numerically in Subsection~\ref{subsec:numexp_eig} and show that Koopman layers perform better than those with general linear operators in Appendix~\ref{ap:eig_general_linear}.
\end{remark}

The proposed model has connections with neural ODE-based models and neural network-based Koopman approaches.
See Appendix~\ref{sec:connection} for more details.

\subsection{Practical implementation}\label{subsec:training}
\begin{figure}
    \centering
    \includegraphics[width=\linewidth]{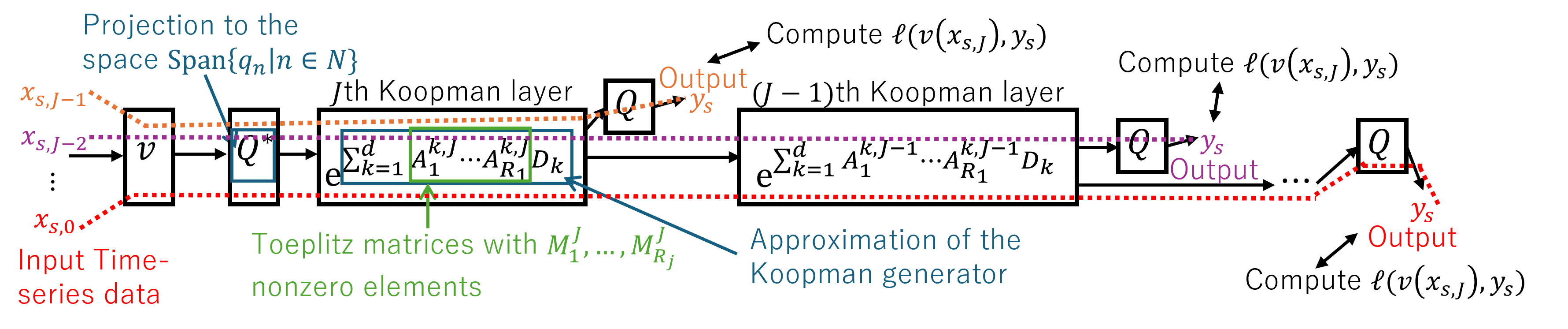}\vspace{-.9cm}
    \caption{Overview of the deep Koopman-layered model.}
    \label{fig:overview}
\end{figure}

We provide a pseudocode of the algorithm of training the deep Koopman-layered model in Algorithm~\ref{al:deep_koopman}.
Here, $\bracket{\cdot,\cdot}$ is the inner product in $L^2(\mathbb{T}^d)$.
Thus, $\bracket{q_n,v}$ means the $n$th Fourier coefficient of a function $v$.
In addition, we put all the learnable parameters $A=[a_{k,j}^{n,r}]_{k=1,\ldots,d,n\in N\bigcap M_r^j,r=1,\ldots,R_j,j=1,\ldots{J}}$.
\red{We input a family of time-series data $\{(x_{s,0},\ldots,x_{s,{J}})\}_{s=1}^S$ to $\mathbf{G}$.
We learn the parameter $A$ by minimizing $\sum_{s=1}^S\ell(v(x_{s,{J}}),\mathbf{G}_j(x_{s,j-1}))$ for $j=1,\ldots,{J}$ for a loss function $\ell$ using an optimization method.}
See Appendix~\ref{ap:pseudocode} for more details.
We also provide an overview of the deep Koopman-layered model in Figure~\ref{fig:overview}.

\begin{algorithm}[t]
\caption{Training deep Koopman-layered model}\label{al:deep_koopman}
 \begin{algorithmic}[1]
 \Require $v\in L^2(\mathbb{T})$, $N\subseteq \mathbb{Z}^d$, $J\in\mathbb{N}$, $R_1,\ldots,R_J\in\mathbb{N}$, $M^j_1,\ldots,M^j_{R_j}\subseteq \mathbb{Z}^d\ (j=1,\ldots,J)$,  $\ell:\mathbb{C}\times \mathbb{C}\to\mathbb{R}_+$, time-series $\{(x_{s,1},\ldots,x_{s,J})\}_{s=1}^S$
 \Ensure Learnable parameter $A$ of the deep Koopman-layered model
 \State Compute a vector $u=[\bracket{q_n,v}]_{n\in N}$.
 \State Set $(D_k)_{l,l}=\mathrm{i}l_k$.
 \State Initialize $A$.
 \For{each epoch}
 \For{each layer $j=J,\ldots,1$}
 \State Compute $u=\mr{e}^{\sum_{k=1}^dA_1^{k,j}\cdots A_{R_j}^{k,j}D_k}u$ using a Krylov subspace method. 
 \State Compute the output $y_{s}=\sum_{n\in N}q_n(x_{s,j-1})u_n$ of $j$th layer for $s=1,\ldots S$.
 \State Compute the loss $H_j=\sum_{s=1}^S\ell(v(x_{s,{J}}),y_{s})$.
 \EndFor
 \State Compute the total loss $H=\sum_{j=1}^{J}H_j$ and the gradient of $H$ with respect to $A$ and apply a gradient method to update the learnable parameter $A$.
 \EndFor
 \end{algorithmic}
\end{algorithm}

\section{Universality}
In this section, we show the universal property of the proposed deep Koopman-layered model.
{We can interpret the model $\mathbf{G}$ as the approximation of the target function by transforming the function $v$ into the target function using the linear operator $Q_N\mr{e}^{\mathbf{L}_1}\cdots \mr{e}^{\mathbf{L}_J}Q_N^*$.}
If we can represent any linear operator by $\mr{e}^{\mathbf{L}_1}\cdots \mr{e}^{\mathbf{L}_J}$, then we can transform $v$ into any target function {in $V_N$, which means we can approximate any function as $N$ goes to the whole set $\mathbb{Z}^d$. }
Thus, this property corresponds to the universality of the model.
In Section~\ref{sec:deep_koopman}, by constructing the model with the matrix $\mr{e}^{\mathbf{L}_1}\cdots \mr{e}^{\mathbf{L}_J}$ based on the Koopman operators with the Fourier functions, we restrict the number of parameters of the linear operator that transforms $v$ into the target function. 
The universality of the model means that this restriction is reasonable in the sense of representing the target functions using the deep Koopman-layered model.

Let $T(N,\mathbb{C})=\{\sum_{k=1}^dA_1^{k}\cdots A_{R_k}^{k} D_k,\,\mid\, R_k\in\mathbb{N},\ A_1^{k}\cdots A_{R_k}^{k}\in\mathbb{C}^{N\times N}\mbox{ : Toeplitz}\}$ be the set of matrices in the form of $\mathbf{L}$ in Eq.~\eqref{eq:generator_ap}. 
Let $L_0^2(\mathbb{T}^d)=\overline{\opn{Span}\{q_n\,\mid\,n\neq 0\}}$ be the space of $L^2$ functions whose average is $0$.
We show the following fundamental result of the universality of the model:
\begin{theorem}~\label{thm:universality}
Assume $v\in L_0^2(\mathbb{T}^d)$ and $v\neq 0$.
For any $f\in L_0^2(\mathbb{T}^d)$ with $f\neq 0$ and for any $\epsilon>0$, there exist a finite set $N\subset \mathbb{Z}\setminus \{0\}$, a positive integer $J$, and matrices $\mathbf{L}_1,\ldots,\mathbf{L}_J\in T(N,\mathbb{C})$ such that $\Vert f-\mathbf{G}\Vert\le \epsilon$ and $\mathbf{G}=Q_N\mr{e}^{\mathbf{L}_1}\cdots \mr{e}^{\mathbf{L}_{J}}Q_N^*v$. 
\end{theorem}

Theorem~\ref{thm:universality} is for a single function $f$, which corresponds to an autonomous system.
According to Theorem~\ref{thm:universality}, we may need more than one layer even for autonomous systems.
This is an effect of the approximation of the generator.
If we can use the true Koopman generator, then we only need one layer for autonomous systems.
However, since we approximated the generator using matrices, we may need more than one layer.

Applying Theorem~\ref{thm:universality} {for each component of $\mathbf{G}$}, we obtain the following result for the flow $g_{\tilde{J}}(t_{\tilde{J}},\cdot)\circ \cdots \circ g_1(t_1,\cdot)$ with $\tilde{J}\in\mathbb{N}$ of a nonautonomous system, which is considered in Eq.~\eqref{eq:model_flows}.
\begin{corollary}\label{cor:universality_time_series}
Assume $v\in L_0^2(\mathbb{T}^d)$ and $v\neq 0$.
For any sequence $g_1(t_1,\cdot),\ldots, g_{\tilde{J}}(t_{\tilde{J}},\cdot)$ of flows satisfying $v\circ g_{\tilde{J}}(t_{\tilde{J}},\cdot)\circ\cdots \circ g_{j}(t_{j},\cdot)\in L^2_0(\mathbb{T}^d)$ and $v\circ g_{\tilde{J}}(t_{\tilde{J}},\cdot)\circ\cdots \circ g_{j}(t_{j},\cdot)\neq 0$ for $j=1,\ldots \tilde{J}$, and for any $\epsilon>0$, there exist a finite set $N\subset \mathbb{Z}\setminus \{0\}$, integers $0<J_1 <\cdots < J_{\tilde{J}}$, and matrices $\mathbf{L}_1,\ldots,\mathbf{L}_{J_{\tilde{J}}}\in T(N,\mathbb{C})$ such that $\Vert v\circ g_{\tilde{J}}(t_{\tilde{J}},\cdot)\circ \cdots \circ g_j(t_j,\cdot)-\mathbf{G}_j\Vert\le \epsilon$ and $\mathbf{G}_j=Q_N\mr{e}^{\mathbf{L}_{J_{j-1}+1}}\cdots \mr{e}^{\mathbf{L}_{J_{\tilde{J}}}}Q_N^*v$ for $j=1,\ldots,\tilde{J}$, where $J_0=1$. 
\end{corollary}

\begin{remark}
The function space $L_0^2(\mathbb{T}^d)$ for the target function is not restrictive. 
By adding one additional learnable parameter $c\in\mathbb{C}$ to the model $\mathbf{G}$ in Theorem~\ref{thm:universality} and consider the model $\mathbf{G}(x)+c$ for an input $x\in\mathbb{T}^d$, we can represent any function in $L^2(\mathbb{T}^d)$.
\end{remark}

\begin{remark}\label{rmk:convergence_rate}
In the same manner as Theorem~\ref{thm:universality}, we can show that we can represent any function in $V_N=\operatorname{Span}\{q_n\,\mid\,n\in N\}$ exactly using the deep Koopman-layered model.
Thus, if the decay rate of the Fourier transform of the target function is $\alpha$, then the convergence rate with respect to $N$ is $O(({1-\alpha^2})^{-d/2})$.
See Appendix~\ref{ap:convergence_rate} for more details.
\end{remark}

The proof of Theorem~\ref{thm:universality} is obtained by a linear algebraic approach.
{By virtue of setting $f_j$ as the product of weighted sums of Fourier functions as explained in Eq.~\eqref{eq:f}, the approximation of the Koopman generator is composed of Toeplitz matrices.} 
As a result, we can apply the following proposition regarding Toeplitz matrices by \citet[Theorem 2]{ye16}.
\begin{proposition}\label{prop:toeplitz}
For any $B\in\mathbb{C}^{N\times N}$, there exists $R=\lfloor \vert N\vert/2 \rfloor+1$ Toeplitz matrices $A_1,\ldots,A_R$ such that $B=A_1\cdots A_R$.
\end{proposition}

We use Proposition~\ref{prop:toeplitz} to show the following lemma regarding the representation with $T(N,\mathbb{C})$.
\begin{lemma}\label{lem:toeplitz_diagonal}
Assume $N\subset \mathbb{Z}^d\setminus\{0\}$.
Then, we have $\mathbb{C}^{N\times N}=T(N,\mathbb{C})$.
\end{lemma}

Since $\mathbb{C}^{N\times N}$ is a Lie algebra and the corresponding Lie group $GL(N,\mathbb{C})$, the group of nonsingular $N$ by $N$ matrices, is connected, we have the following lemma~\citep[Corollary 3.47]{hall15}.
\begin{lemma}\label{lem:lie_algebra}
We have $GL(N,\mathbb{C})=\{\mr{e}^{\mathbf{L}_1}\cdots \mr{e}^{\mathbf{L}_{J}}\,\mid\,J\in \mathbb{N},\ \mathbf{L}_1,\ldots,\mathbf{L}_J\in \mathbb{C}^{N\times N}\}$.
\end{lemma}
We also use the following transitive property of $GL(N,\mathbb{C})$ and finally obtain Theorem~\ref{thm:universality}.
\begin{lemma}\label{lem:transitive}
For any $\mathbf{u},\mathbf{v}\in\mathbb{C}^N\setminus \{0\}$, there exists $A\in GL(N,\mathbb{C})$ such that $\mathbf{u}=A\mathbf{v}$.
\end{lemma}

\section{Generalization bound}
We investigate the generalization property of the proposed deep Koopman-layered model in this section.
Our framework with Koopman operators enables us to derive a generalization bound involving the norms of Koopman operators.

Let $\mcl{G}_N=\{Q_N\mr{e}^{\mathbf{L}_1}\cdots \mr{e}^{\mathbf{L}_J}Q_N^*v\,\mid\, \mathbf{L}_1,\ldots,\mathbf{L}_J\in T(N,\mathbb{C})\}$ be the function class of deep Koopman-layered model~\eqref{eq:koopman_layered}.
Let $\ell(\mcl{G}_N)=\{(x,y)\mapsto\ell(f(x),y)\,\mid\,f\in\mcl{G}_N\}$ for a function $\ell$ bounded by $C>0$.
We have the following result of a generalization bound for the deep Koopman-layered model.
\begin{proposition}\label{prop:generalization}
Let $h\in\ell(\mcl{G}_N)$, $x$ and $y$ be random variables, $S\in \mathbb{N}$, and $x_1,\ldots,x_S$ and $y_1,\ldots,y_S$ be i.i.d. samples drawn from the distributions of $x$ and $y$, respectively.
Let $\tau>0$ and $\alpha=\sum_{j\in\mathbb{Z}^d}\mr{e}^{-2\tau\Vert j\Vert_1}$, where $\Vert [j_1,\ldots,j_d]\Vert_1=\vert j_1\vert+\cdots +\vert j_d\vert$.
For any $\delta>0$, with probability at least $1-\delta$, we have
\begin{align*}
&\mr{E}[h(x,y)]\le \frac{1}{S}\sum_{s=1}^S h(x_n,y_n)
\frac{\alpha}{\sqrt{S}}\max_{j\in N}\mr{e}^{\tau\Vert j\Vert_1}\!\!\!\!\!\!\!\!\!\sup_{\mathbf{L}_1,\ldots,\mathbf{L}_J\in T(N,\mathbb{C})}\!\!\!\!\!\Vert \mr{e}^{\mathbf{L}_1}\Vert\cdots \Vert\mr{e}^{\mathbf{L}_J}\Vert\,\Vert v\Vert
+3C\sqrt{\frac{\log(\delta/2)}{S}}.
\end{align*}  
\end{proposition}

We can see that the complexity of the model depends exponentially on both $N$ and $J$.
We use the Rademacher complexity to derive Proposition~\ref{prop:generalization}. 
For this purpose, we regard the model~\eqref{eq:model_flows} as a function in an RKHS.
See Appendix~\ref{ap:generalization_details} for more details.

\begin{remark}
The exponential dependence of the generalization bound on the number of layers is also typical for standard neural networks~\citep{neyshabur15,bartlett17,golowich18,hashimoto23}.
\end{remark}
\begin{remark}\label{rmk:regularization}
Based on Proposition~\ref{prop:generalization}, we can control the generalization error by adding a regularization term to the loss function to make $\Vert \mr{e}^{\mathbf{L}_1}\Vert\cdots \Vert\mr{e}^{\mathbf{L}_J}\Vert$ smaller.
We note that $\Vert \mr{e}^{\mathbf{L}_j}\Vert$ is expected to be bounded with respect to $N$ since the corresponding Koopman operator is bounded in our setting. See Appendix~\ref{ap:boundedness} for more details.
\end{remark}

\section{Numerical results}\label{sec:num_exp}
We empirically confirm the validity of the proposed deep Koopman-layered model.
The experimental settings are detailed in Appendix~\ref{ap:exp_detail}.


\subsection{Representation power and generalization}\label{subsec:vanderPol}
\begin{wrapfigure}{r}[0pt]{0.3\textwidth}\vspace{-1.9cm}
    \centering
    \subfigure[Without the regularization ($J=1,2$, $S=1000$, without noise)]{\includegraphics[height=2.7cm]{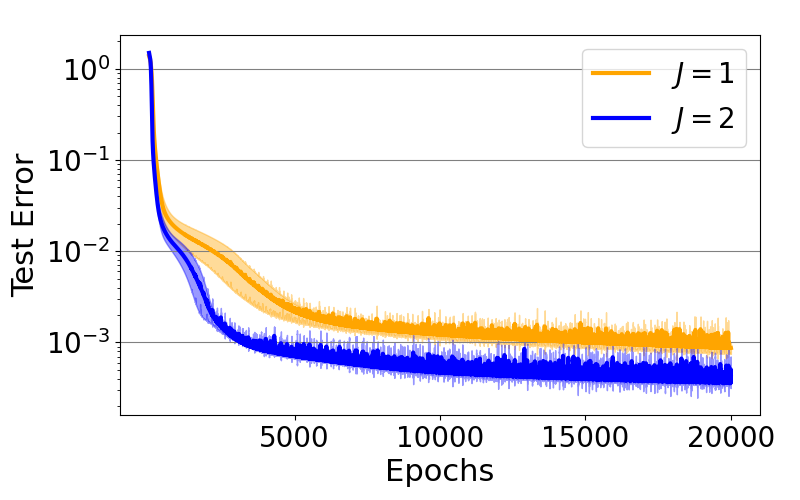}}\vspace{-.3cm}
    \subfigure[With and without the regularization ($J=3$, $S=30$, with noise)]{\includegraphics[height=2.7cm]{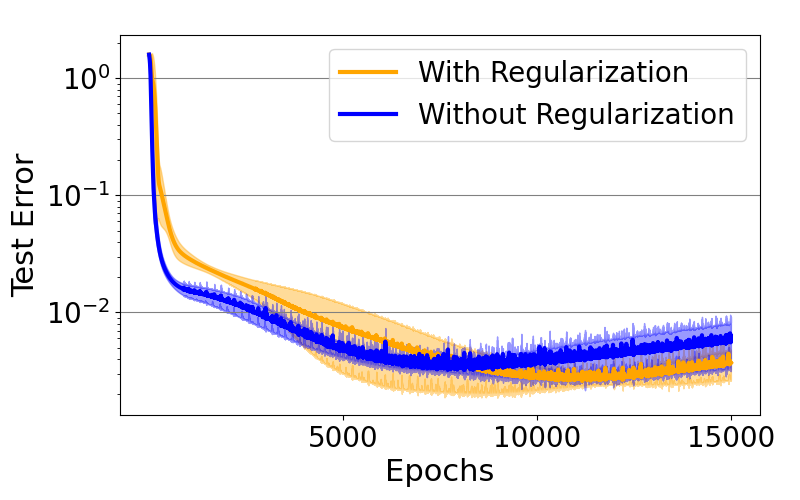}}\vspace{-.4cm}
    \caption{Test errors for different values of $J$ with and without the regularization based on the norms of the Koopman operators. (The average $\pm$ standard deviation of three independent runs.)}
    \label{fig:layers}
\end{wrapfigure}

To confirm the fundamental property of the Koopman layer, we first consider an autonomous system.
Consider the van der Pol oscillator on $\mathbb{T}$:
$\frac{\mr{d}^2x(t)}{\mr{d}t^2}=-\mu (1-x(t)^2) \frac{\mr{d}x(t)}{\mr{d}t}+x(t),$ 
where $\mu=3$.
By setting $\mr{d}x/\mr{d}t$ as a new variable, we regard the system as a first-ordered system on the two-dimensional space.
We generated time-series data according to it and trained the deep-Koopman-layered model.

Figure~\ref{fig:layers} (a) shows the test error for $J=1$ and $J=2$.
We can see that the performance becomes higher when $J=2$ than $J=1$.
Note that Theorem~\ref{thm:universality} is a fundamental result for autonomous systems, and we may need more than one layer even for the autonomous systems.
The result reflects this theoretical result.
In addition, based on Remark~\ref{rmk:regularization}, we added the regularization term $10^{-5}(\Vert \mr{e}^{\mathbf{L}_1}\Vert+\cdots +\Vert \mr{e}^{\mathbf{L}_J}\Vert)$ to reduce the generalization error.
To focus on the generalization property, we reduced the sample size and added more noise to the data.
We set $J=3$ to consider the case where the number of parameters is large.
The results are shown in Figure~\ref{fig:layers} (b).
We can see that with the regularization, we achieve smaller test errors than without the regularization, which implies that with the regularization, the model generalizes well. 

\begin{figure*}[t]\vspace{-.2cm}
\setlength{\tabcolsep}{5pt}
\def\arraystretch{0.8}
\newcolumntype{C}{>{\centering\arraybackslash}X}
\centering
\begin{tabularx}{\textwidth}{Cccccc}
\vspace{-2cm}{Deep Koopman-layerd model}&
\includegraphics[scale=0.17]{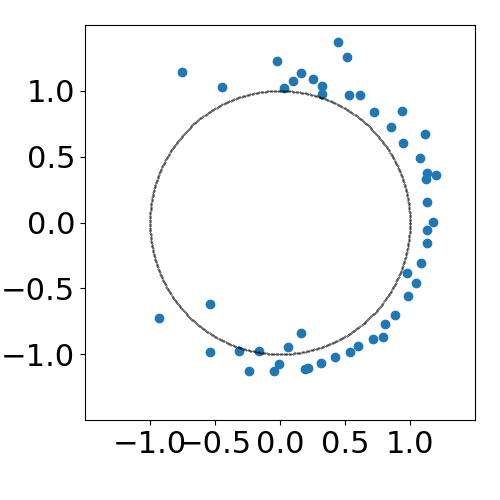}&
\includegraphics[scale=0.17]{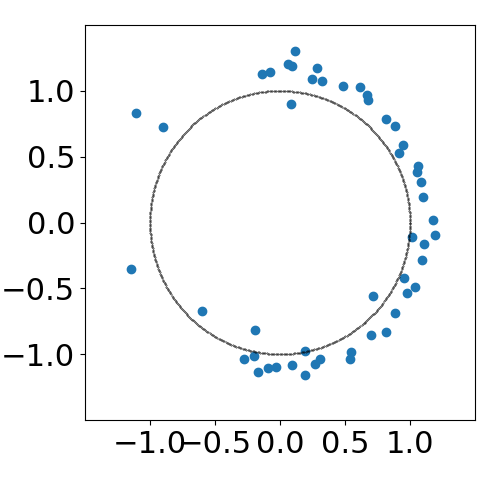}&
\includegraphics[scale=0.17]{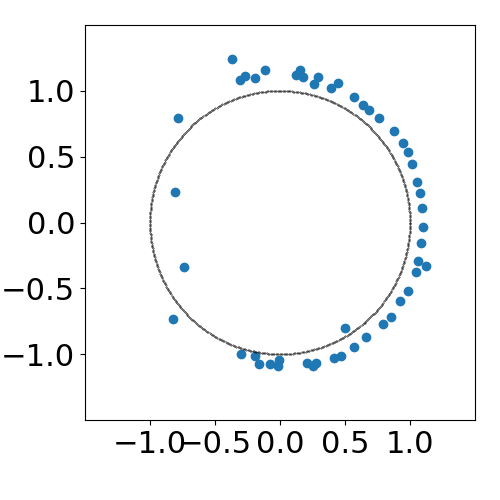}&
\includegraphics[scale=0.17]{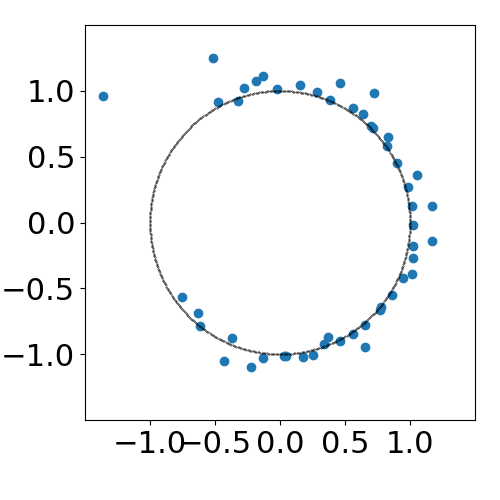}&
\includegraphics[scale=0.17]{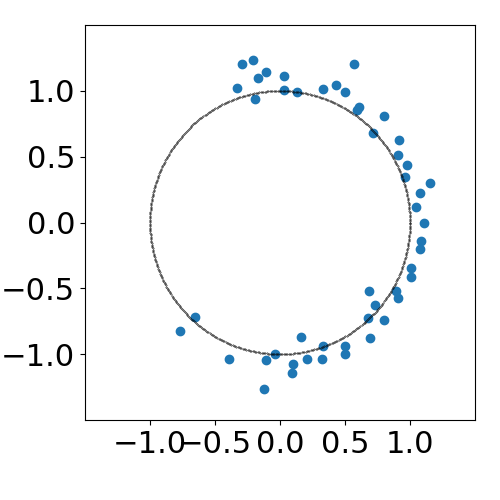}\vspace{-.3cm}\\

\vspace{-2cm}EDMD&
\includegraphics[scale=0.17]{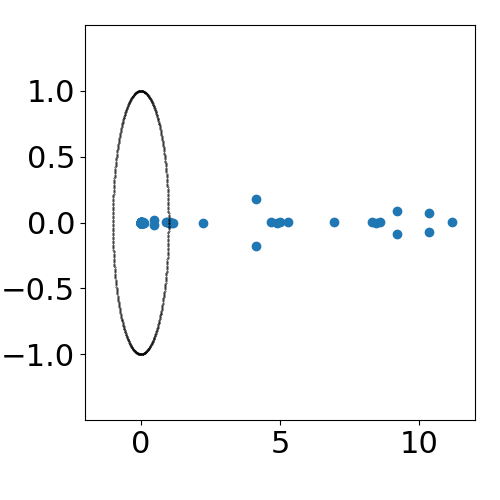}&
\includegraphics[scale=0.17]{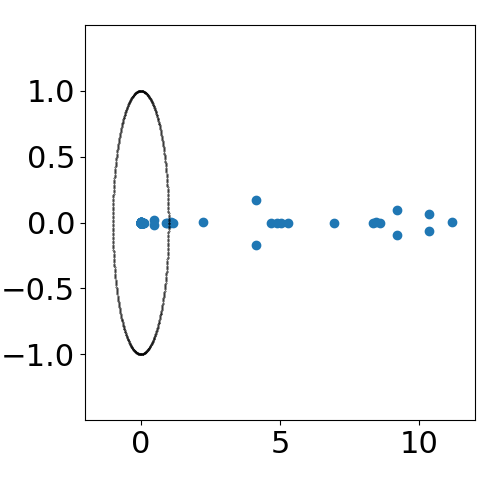}&
\includegraphics[scale=0.17]{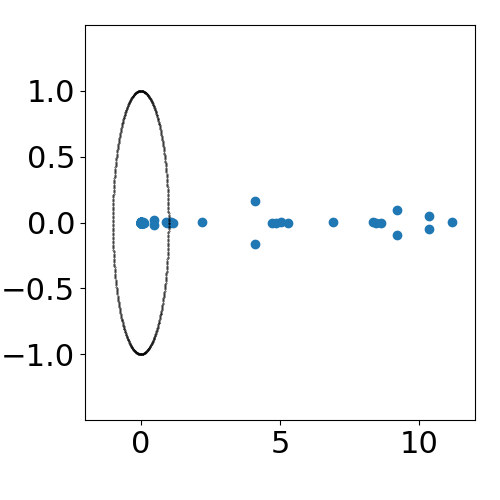}&
\includegraphics[scale=0.17]{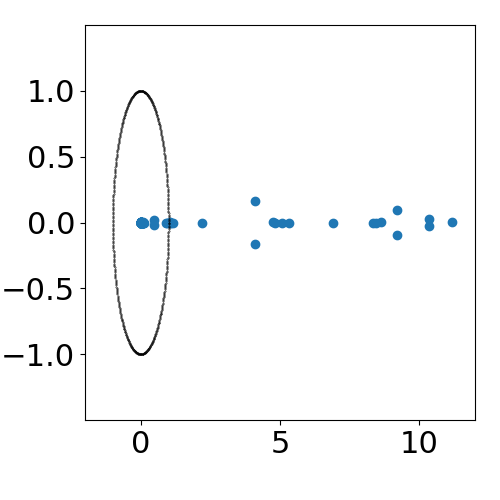}&
\includegraphics[scale=0.17]{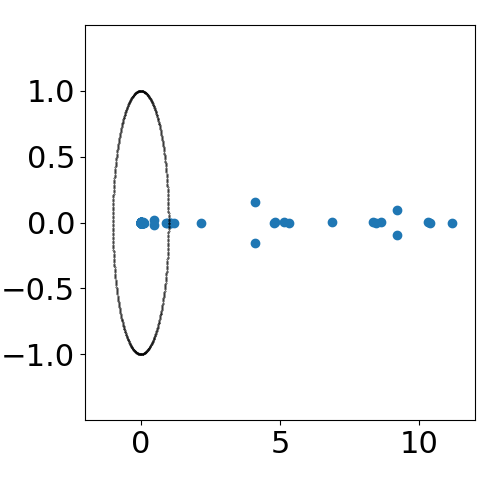}\vspace{-.3cm}\\

\vspace{-2cm}KDMD&
\includegraphics[scale=0.17]{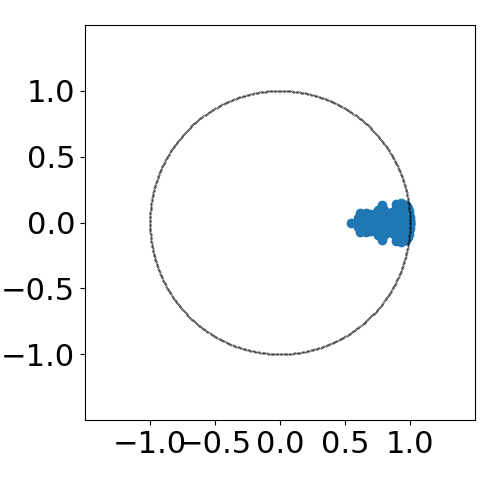}&
\includegraphics[scale=0.17]{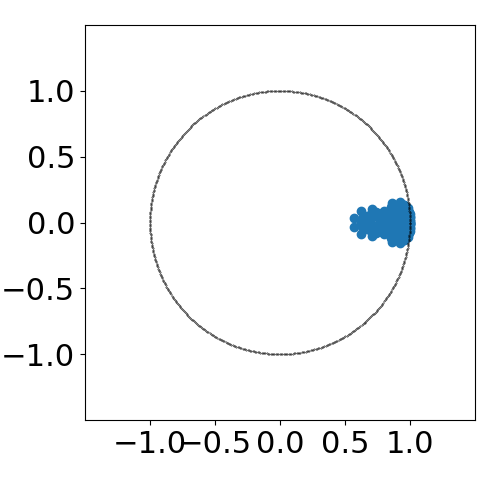}&
\includegraphics[scale=0.17]{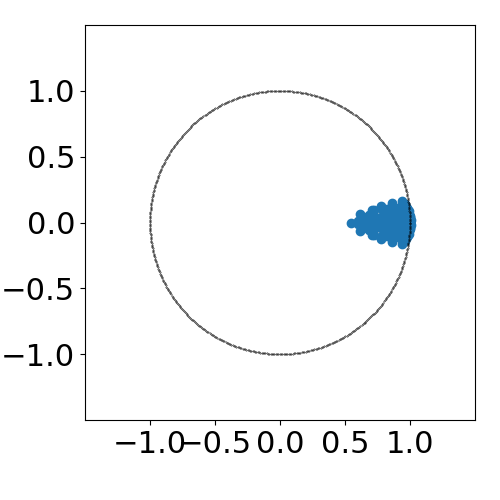}&
\includegraphics[scale=0.17]{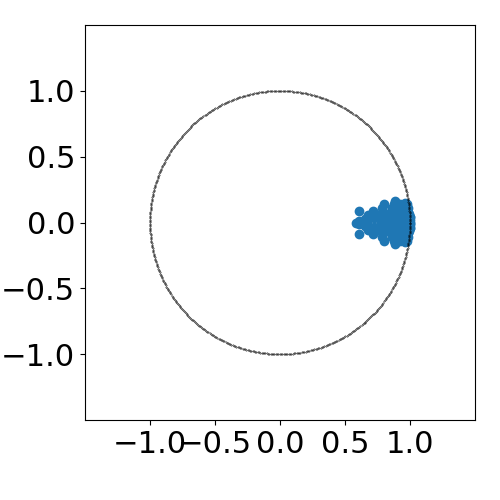}&
\includegraphics[scale=0.17]{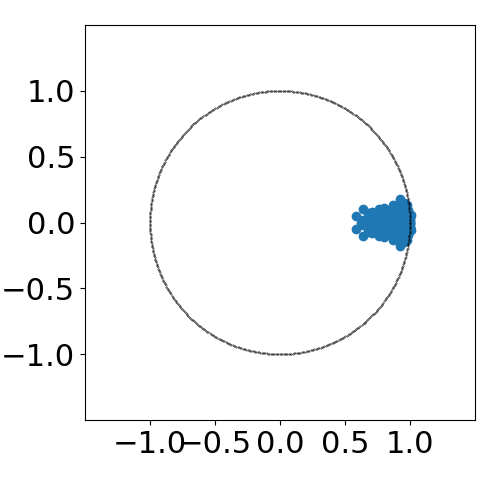}\vspace{-.1cm}\\
&$j=1$&$j=2$&$j=3$&$j=4$&$j=5$
\end{tabularx}\vspace{-.4cm}
\caption{Eigenvalues of the estimated Koopman operators for the nonautonomous measure preserving system.}\label{fig:vol}\vspace{-.2cm}
\end{figure*}

\begin{figure*}[t]
\setlength{\tabcolsep}{5pt}
\def\arraystretch{0.8}
\newcolumntype{C}{>{\centering\arraybackslash}X}
\begin{center}
\begin{tabularx}{\textwidth}{Cccccc}
\vspace{-2cm}{Deep Koopman-layerd model}&
\includegraphics[scale=0.17]{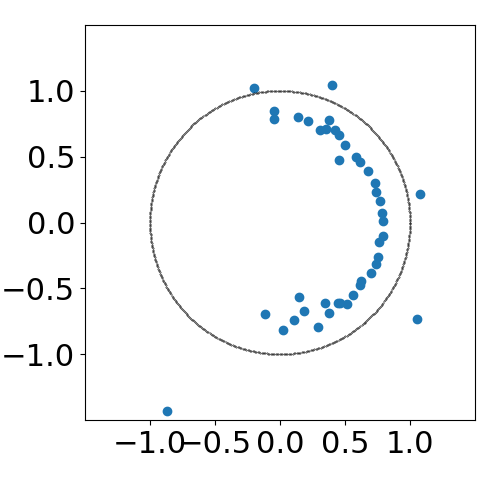}&
\includegraphics[scale=0.17]{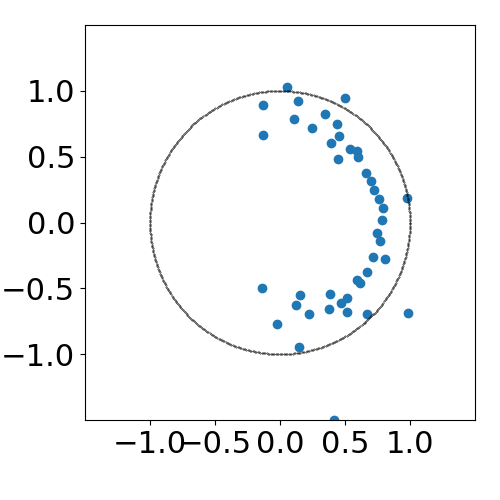}&
\includegraphics[scale=0.17]{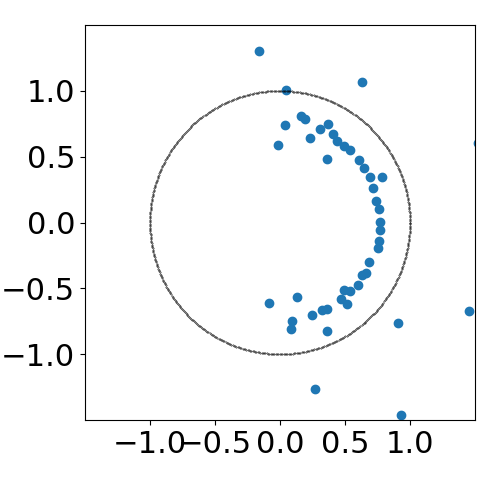}&
\includegraphics[scale=0.17]{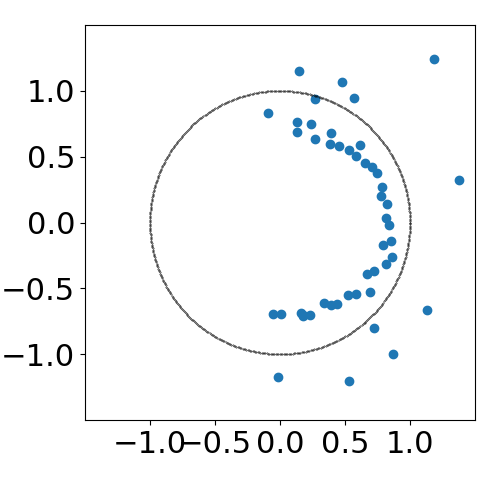}&
\includegraphics[scale=0.17]{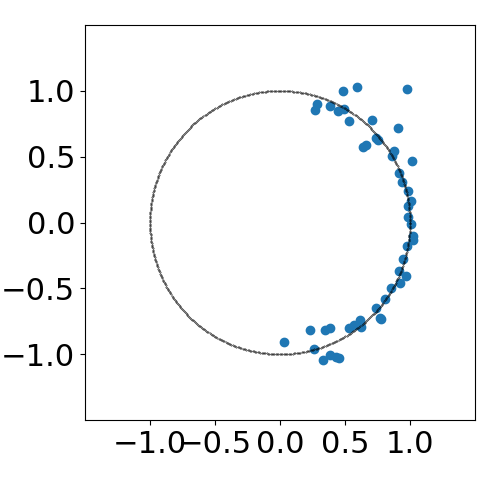}\vspace{-.3cm}\\

\vspace{-2cm}EDMD&
\includegraphics[scale=0.17]{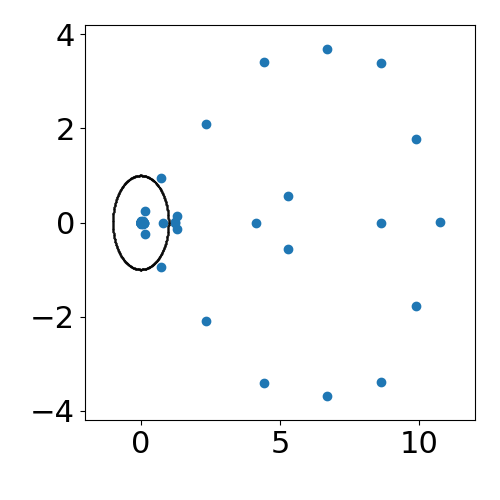}&
\includegraphics[scale=0.17]{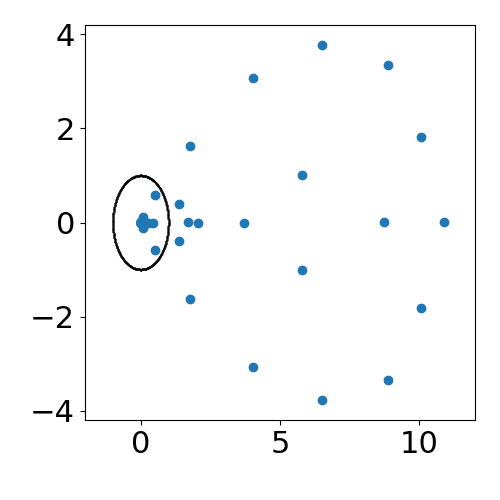}&
\includegraphics[scale=0.17]{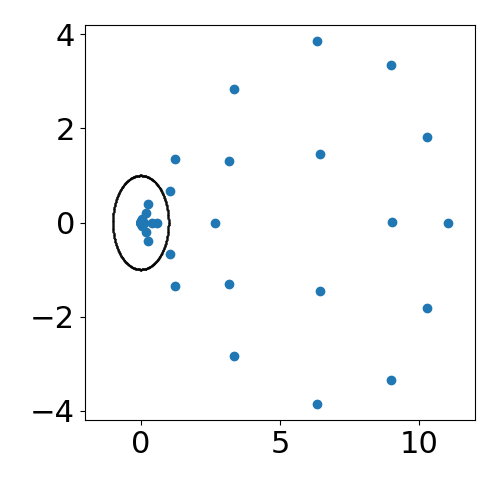}&
\includegraphics[scale=0.17]{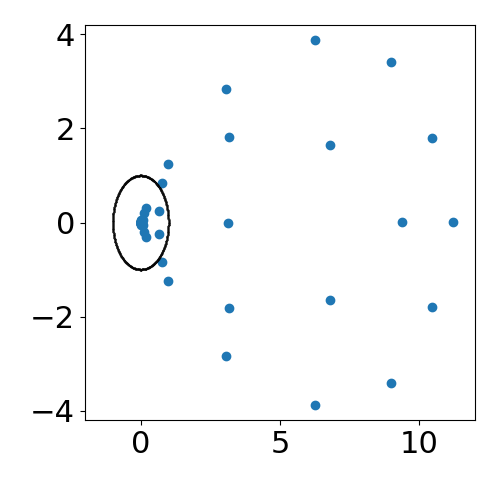}&
\includegraphics[scale=0.17]{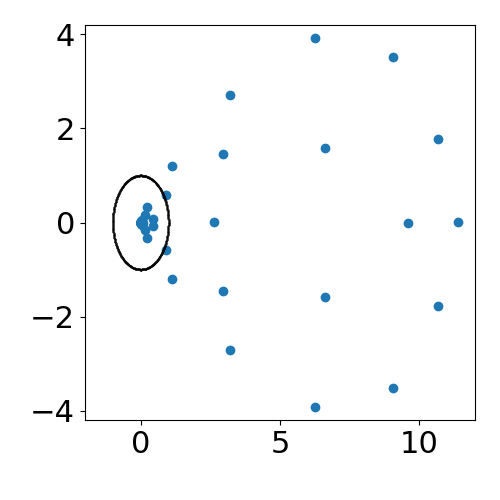}\vspace{-.3cm}\\

\vspace{-2cm}KDMD&
\includegraphics[scale=0.17]{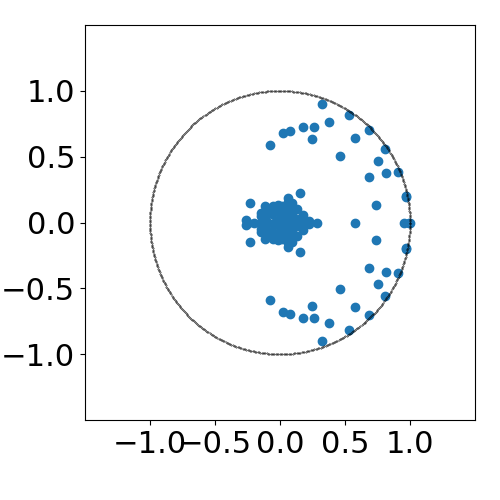}&
\includegraphics[scale=0.17]{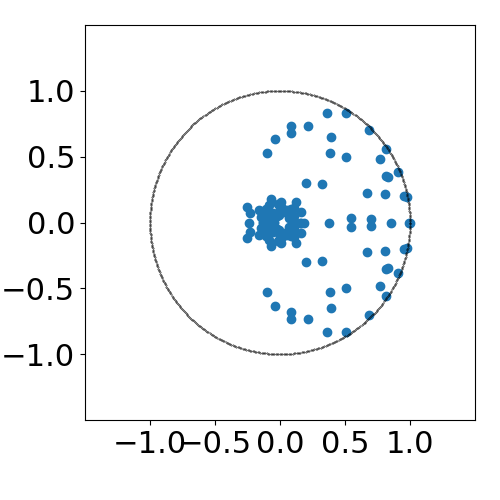}&
\includegraphics[scale=0.17]{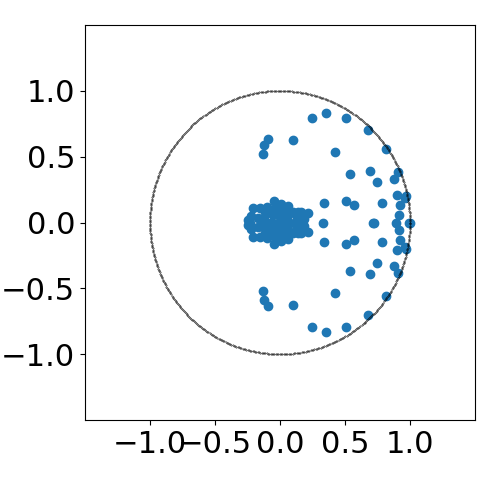}&
\includegraphics[scale=0.17]{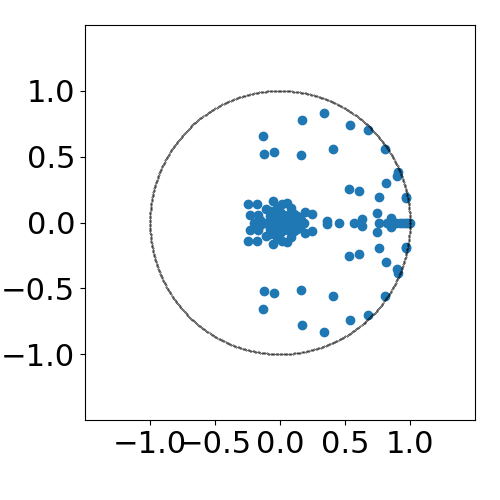}&
\includegraphics[scale=0.17]{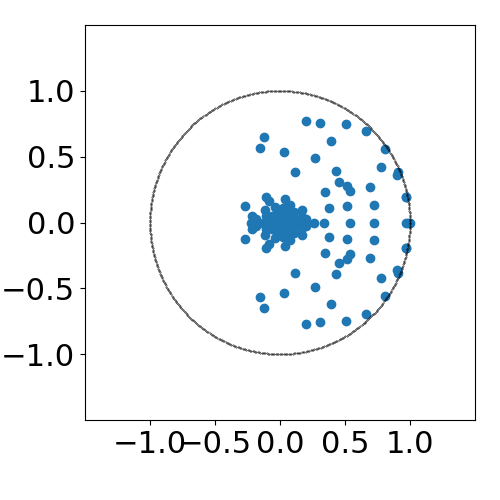}\vspace{-.1cm}\\
&$j=1$&$j=2$&$j=3$&$j=4$&$j=5$
\end{tabularx}\vspace{-.4cm}
\end{center}
\caption{Eigenvalues of the estimated Koopman operators for the nonautonomous damping oscillator. Full results are in Appendix~\ref{ap:exp_eig_addition}.}\label{fig:force}\vspace{-.3cm}
\end{figure*}

\subsection{Eigenvalues of the Koopman-layers for nonautonomous systems}\label{subsec:numexp_eig}
To confirm that we can extract information about the underlying nonautonomous dynamical systems of time-series data using the proposed model, we computed the eigenvalues of the Koopman layers.

\subsubsection{Measure-preserving dynamical system}\label{subsec:voltex}

Consider the nonautonomous dynamical system on $\mathbb{T}^2$ written as
$\big(\frac{\mr{d}x_1(t)}{\mr{d}t},\frac{\mr{d}x_2(t)}{\mr{d}t}\big)=\big(-\frac{\partial \zeta}{\partial x_2}(t,x(t)),\frac{\partial \zeta}{\partial x_1}(t,x(t))\big)
=:f(t,x),$ 
where $\zeta(t,[x_1,x_2])=\mr{e}^{\kappa(\cos(x_1-t)+\cos x_2)}$.
Since the dynamical system $f(t,\cdot)$ is measure-preserving for any $t\in\mathbb{R}$, the corresponding Koopman operator $K^t$ is unitary for any $t\in\mathbb{R}$.
Thus, the spectrum of $K^t$ is on the unit disk in the complex plane.
We generated time-series data 
$x_{s,0},\ldots x_{s,119}$ 
according to the above dynamical system. 
We split the data into 6 subsets $S_t=\{x_{s,j}\,\mid\,s=1,\ldots,1000,\ j=20t,\ldots,20(t+1)-1\}$ for $t=0,\ldots,5$.
Then, we trained the model with 5 Kooman layers on $\mathbb{T}^3$ 
so that the model $Q_N\mr{e}^{\mathbf{L}_j}\cdots\mr{e}^{\mathbf{L}_5}Q_N^*v$ maps samples in $S_{j-1}$ to $S_5$.
We assumed the continuity of the flow of the nonautonomous dynamical system and added a regularization term $0.01\sum_{j=2}^5\Vert\mr{e}^{\mathbf{L}_j}-\mr{e}^{\mathbf{L}_{j-1}}\Vert$ to make the Koopman layers
next to each other become close. 

After training the model sufficiently (after 3000 epochs), we computed the eigenvalues of the approximation $\mr{e}^{\mathbf{L}_j}$ of the Koopman operator for each layer $j=1,\ldots,5$.
For comparison, we estimated the Koopman operator $K^{t_j}_j$ using EDMD and KDMD~\citep{kawahara16} with the dataset $S_{j-1}$ and $S_j$ separately for $j=1,\ldots,5$.
For EDMD, we used the same Fourier functions $\{q_j\,\mid\,j\in N\}$
as the deep Koopman-layered model for the dictionary functions.
For KDMD, we used the Gaussian kernel. 
Figure~\ref{fig:vol} illustrates the results.
We can see that the eigenvalues of the Koopman layers are \red{correctly} distributed on the unit circle for $j=1,\ldots,5$, which enables us to observe that the dynamical system is measure-preserving for any time.
On the other hand, the eigenvalues of the estimated Koopman operators with EDMD and KDMD are not on the unit circle, which implies that the separately applying EDMD and KDMD failed to capture the property of the dynamical system.
\red{This is because the proposed model properly incorporates information from multiple Koopman layers.}
We also constructed a model by replacing each layer of the Koopman model with a general matrix for further comparison.
The results are in Appendix~\ref{ap:eig_general_linear}, which show the validity of using Koopman operators in extracting information about the underlying dynamical system.

\subsubsection{Damping oscillator with external force}\label{subsec:damping}
Consider the nonautonomous dynamical system regarding a damping oscillator on a compact subspace of $\mathbb{R}$ written as
$\frac{\mr{d}^2x(t)}{\mr{d}t^2}=-\alpha \frac{\mr{d}x(t)}{\mr{d}t}-x(t)-a\sin (bt),$ 
where $\alpha=0.1$, $a=b=1$.
By setting ${\mr{d}x}/{\mr{d}t}$ as a new variable, we regard the above system as a first-ordered system on the two-dimensional space.
We generated data, constructed the deep Koopman-layered model, and applied EDMD and KDMD for comparison in the same manner as Subsection~\ref{subsec:voltex}.
However, in this case, we also implemented the EDMD and its variant with learned dictionary functions.
Figure~\ref{fig:force} and Figure~\ref{fig:additional} in Appendix~\ref{ap:exp_eig_addition} illustrate the results.
In this case, since the dynamical system is not measure preserving, it is reasonable that the estimated Koopman operators have eigenvalues inside the unit circle.
We can see that many eigenvalues for the deep Koopman-layered model are distributed inside the unit circle, and the distribution changes along the layers.
Since the external force becomes large as $t$ becomes large, the damping effect becomes small as $t$ becomes large (corresponding to $j$ becoming large).
Thus, the number of eigenvalues distributed inside the unit circle becomes small as $j$ becomes large.
On the other hand, we cannot obtain this type of observation from the separate estimation of the Koopman operators by EDMD and KDMD.
\begin{wraptable}{r}[0pt]{0.6\textwidth}\vspace{-.3cm}
    \centering
    \caption{Relative squared error of the prediction with $\Delta T=192$. (The Average $\pm$ standard deviation for three independent runs.) Full results are documented in Apendix~\ref{ap:full_exp_forecast}.}
    \label{tab:forecast}
    \centering
    \begin{tabular}{c|c|c}
    \hline
    Dataset  & Koopman-layered
    & Fourier filter\\
    \hline
    ETT & {\bf 0.195$\pm$0.0155}  & 0.223$\pm$0.00252\\
    Electricity & {\bf 0.158$\pm$0.000567} & 0.166$\pm$0.000848 \\
    Exchange & {\bf 0.223$\pm$0.00518} & 0.666$\pm$0.100\\
    Traffic & {\bf 0.246$\pm$0.0168} &  0.348$\pm$0.000625 \\
    Weather & 0.188$\pm$0.00322 & {\bf 0.106$\pm$0.000294}\\
    ILI & {\bf 0.259$\pm$0.0109}  & 0.911$\pm$0.311 \\
    \hline
    \end{tabular}\vspace{-.9cm}
\end{wraptable}

\subsection{Application to time-series forecasting}\label{subsec:exp_forecast}
We can also apply the proposed method to time-series forecasting.
Applying the idea of \citet{wang23_koopman,liu23}, 
we decompose the Koopman operator \red{$K^{t_j}$} for $j=1,\ldots,J$ into the time-invariant part $K_{\opn{inv}}$ and the time-variant part and \red{$K_{j,\opn{var}}$}.
We can estimate $K_{\opn{inv}}$ as $K_{\opn{inv}}=\sum_{i=1}^{n_{\opn{inv}}}\sigma_iu_iv_i^*$, where we consider a set of $n_{\opn{inv}}$ singular values $\{\sigma_{i,j}\}_{i=1}^{n_{\opn{inv}}}$ and vectors $\{u_{i,j},v_{i,j}\}_{i=1}^{n_{\opn{inv}}}$ of the approximated Koopman operators for each $j=1,\ldots,J$ that satisfy $(u_{j,i},v_{j,i})\approx (u_{l,i},v_{l,i})$ for $j\neq l$ and set $\sigma_i=\sum_{j=1}^J\sigma_{j,i}/J$, $u_i=\sum_{j=1}^J u_{j,i}/J$, and $v_i=\sum_{j=1}^Jv_{j,i}/J$.
Since these singular vectors are invariant over time, we can extract the time-invariant property of $t\le t_J$ and can forecast time-series well even for $t>t_J$.
For forecasting the value at $t$, we use a short time-series before $t$ to construct $K_{j,\opn{var}}$.
For more details, see Appendix~\ref{ap:forecast}.

We used real-world benchmark datasets ETT, Electricty,
Exchange, Traffic, Weather, and ILI\footnote{\url{https://drive.google.com/drive/folders/1ZOYpTUa82_jCcxIdTmyr0LXQfvaM9vIy}},
which are also used in \citet{liu23, wu21autoformer} and other related papers, to confirm the applicability of the deep Koopman-layered model to time-series forecasting.
To construct $K_{\opn{inv}}$, we applied the deep Koopman-layered model with $J=9$.
We constructed $K_{j,\opn{var}}$ using the KDMD with the Laplacian kernel.
For forecasting $x_{t_{J}+1},\ldots,x_{t_J+1+\Delta T}$, we used the given time-series $x_{t_J-2\Delta T},\ldots,x_{t_J}$ with $\Delta T=48,96,144,192$, in the same manner as \citet{liu23}.
For comparison, we also applied the Fourier filter to split the time-series data into time-invariant and variant parts and applied the EDMD with learned dictionary functions, which is proposed by~\citet{liu23}.
They compare their approach to standard approaches for time-series forecasting and conclude that their approach outperforms these standard approaches. 
Therefore, we chose this approach as a counterpart to our approach.
The result is shown in Table~\ref{tab:forecast}.
We can see that the deep Koopma-layered model outperformed the Fourier filter in many cases.
To simply observe the ability to extract time-invariant features, we used the KDMD to construct $K_{\opn{var}}$ for both the deep Koopman-layered and Fourier filter approach.
Note that for both of these approaches, we can apply a more advanced method e.g., the one proposed by~\citep{liu23}, or the deep Koopman-layered model again, to construct $K_{\opn{var}}$ and obtain better performances.
The results show the potential power of the proposed model for the application to time-series forecasting.

We also conducted experiments with the presence of noise, different values of the number $J$ of layers, and different numbers of Krylov iterations.
The results are in Appendix~\ref{ap:time_forecast_additional}.
The proposed approach is more robust than or as robust as the Fourier filter approach with respect to the noise.
We also emphasize that it is robust with respect to the choice of $J$ and the number of Krylov iterations.

\section{Conclusion and limitation}\label{sec:conclusion}
In this paper, we proposed deep Koopman-layered models based on the Koopman operator theory combined with Fourier functions and Toeplitz matrices.
We showed that the Fourier basis forms a proper approximation space of the Koopman operators in the sense of the universal and generalization properties of the model.
In addition to the theoretical solidness, the flexibility of the proposed model allows us to train the model to fit time-series data coming from nonautonomous dynamical systems.
One difficulty for practical computations is that the index subset $N$
can become large when the input dimension $d$ is large, and the computational cost becomes expensive.
We can alleviate the cost by using neural networks (See Appendix~\ref{ap:exp_detail_forecast}), but further investigations are future work.

\subsection*{Acknowledgements}
We would like to thank Dr. Isao Ishikawa for a constructive discussion.

\bibliography{koopmanbib}
\bibliographystyle{icml2025}


\clearpage
\appendix

\section*{Appendix}

\section{Proofs}\label{ap:proof}
We provide the proofs of statements in the main text.
\begin{mythm}[Proposition~\ref{prop:representation_matrix}]
The $(n,l)$-entry of the representation matrix $Q_N^*L_jQ_N$ of the approximated operator is
\begin{align*}
&\sum_{k=1}^d\sum_{n_{R_j}-l\in M^j_{R_j}}\sum_{n_{R_j-1}-n_{R_j}\in M^j_{R_j-1}}\cdots \sum_{n_{2}-n_3\in M^j_{2}}\sum_{n-n_2\in M^j_{1}}\nn\\
&\qquad\qquad\qquad\qquad  a^{j,k}_{n_{R_j}-l,R_j}a^{j,k}_{n_{R_j-1}-n_{R_j},R_{j-1}}\cdots a^{j,k}_{n_2-n_3,2}a^{j,k}_{n-n_2,1} \mr{i}l_k,
\end{align*}
where $l_k$ is the $k$th element of the index $l\in\mathbb{Z}^d$.
Moreover, we set $n_r=m_{R_j}+\cdots +m_r+l$, thus $n_1=n$, $m_r=n_{r}-n_{r+1}$ for $r=1,\ldots,R_j-1$, and $m_{R_j}=n_{R_j}-l$.
\end{mythm}
\begin{proof}
We have
\begin{align*}
\bracket{q_n,L_jq_l}
&=\Bbracket{q_n,\sum_{k=1}^d\sum_{m_{R_j}\in M^j_{R_j}}a^{j,k}_{m_{R_j},R_j}q_{m_{R_j}}\cdots \sum_{m_1\in M^j_1}a^{j,k}_{m_1,1}q_{m_1}\mr{i}l_kq_l}\\
&=\Bbracket{q_n,\sum_{k=1}^d\sum_{m_{R_j}\in M^j_{R_j}}\cdots \sum_{m_1\in M^j_1}a^{j,k}_{m_{R_j},R_j}\cdots a^{j,k}_{m_1,1} q_{m_{R_j}+\cdots +m_1+l}\mr{i}l_k}\\
&=\sum_{k=1}^d\sum_{\substack{m_{R_j}+\cdots +m_1+l=n\\ m_{R_j}\in M^j_{R_j}\cdots m_1\in M^j_1}}a^{j,k}_{m_{R_j},R_j}\cdots a^{j,k}_{m_1,1} \mr{i}l_k\\
&=\sum_{k=1}^d\sum_{n_{R_j}-l\in M^j_{R_j}}\sum_{n_{R_j-1}-n_{R_j}\in M^j_{R_j-1}}\cdots \sum_{n_{2}-n_3\in M^j_{2}}\sum_{n-n_2\in M^j_{1}}\\
&\qquad\qquad\qquad\qquad  a^{j,k}_{n_{R_j}-l,R_j}a^{j,k}_{n_{R_j-1}-n_{R_j},R_{j-1}}\cdots a^{j,k}_{n_2-n_3,2}a^{j,k}_{n-n_2,1} \mr{i}l_k.
\end{align*}
\end{proof}

\begin{mythm}[Corollary~\ref{cor:universality_time_series}]
Assume $v\in L_0^2(\mathbb{T}^d)$ and $v\neq 0$.
For any sequence $g_1(t_1,\cdot),\ldots, g_{\tilde{J}}(t_{\tilde{J}},\cdot)$ of flows that satisfies $v\circ g_{\tilde{J}}(t_{\tilde{J}},\cdot)\circ\cdots \circ g_{j}(t_{j},\cdot)\in L^2_0(\mathbb{T}^d)$ and $v\circ g_{\tilde{J}}(t_{\tilde{J}},\cdot)\circ\cdots \circ g_{j}(t_{j},\cdot)\neq 0$ for $j=1\ldots \tilde{J}$, and for any $\epsilon>0$, there exist a finite set $N\subset \mathbb{Z}\setminus \{0\}$, integers $0<J_1 <\cdots < J_{\tilde{J}}$, and matrices $\mathbf{L}_1,\ldots,\mathbf{L}_{J_{\tilde{J}}}\in T(N,\mathbb{C})$ such that $\Vert v\circ g_{\tilde{J}}(t_{\tilde{J}},\cdot)\circ \cdots \circ g_j(t_j,\cdot)-\mathbf{G}_j\Vert\le \epsilon$ and $\mathbf{G}_j=Q_N\mr{e}^{\mathbf{L}_{J_{j-1}+1}}\cdots \mr{e}^{\mathbf{L}_{J_{\tilde{J}}}}Q_N^*v$ for $j=1,\ldots,\tilde{J}$, where $J_0=1$. 
\end{mythm}
\begin{proof}
Since $v\circ g_{\tilde{J}}(t_{\tilde{J}},\cdot)\circ\cdots \circ g_{j}(t_{j},\cdot)\in L^2_0(\mathbb{T}^d)$ and $v\circ g_{\tilde{J}}(t_{\tilde{J}},\cdot)\circ\cdots \circ g_{j}(t_{j},\cdot)\neq 0$, there exist finite $N_{j}\subset\mathbb{Z}^d\setminus \{0\}$ and $\mathbf{G}_{j}\in V_{N_j}$, $\mathbf{G}_{j}\neq 0$ such that $\Vert v\circ g_{\tilde{J}}(t_{\tilde{J}},\cdot)\circ\cdots \circ g_{j}(t_{j},\cdot)-\mathbf{G}_j\Vert\le \epsilon$ for $j=1,\ldots,\tilde{J}$.
Since $v\in L_0^2(\mathbb{T}^d)$ and $v\neq 0$, there exist finite $N_{\tilde{J}+1}\subset\mathbb{Z}^d\setminus \{0\}$ such that $Q_{N_{\tilde{J}+1}}^*v\neq 0$.
Let $N=\bigcup_{j=1}^{\tilde{J}+1}N_j$.
By Lemma~\ref{lem:transitive}, since $Q_N^*v\neq 0$, there exist  $J_{\tilde{J}-1},J_{\tilde{J}}\in\mathbb{N}$ and $\mathbf{L}_{J_{\tilde{J}-1}+1},\ldots,\mathbf{L}_{J_{\tilde{J}}}\in T(N,\mathbb{C})$ such that $\mathbf{G}_{\tilde{J}}=Q_N\mr{e}^{\mathbf{L}_{J_{\tilde{J}-1}+1}}\cdots \mr{e}^{\mathbf{L}_{J_{\tilde{J}}}}Q_N^*v$.
Since $\mathbf{G}_{\tilde{J}}\neq 0$, again by Lemma~\ref{lem:transitive}, 
there exist  
$J_{\tilde{J}-2}\in\mathbb{N}$ and $\mathbf{L}_{J_{\tilde{J}-2}+1},\ldots,\mathbf{L}_{J_{\tilde{J}-1}}\in T(N,\mathbb{C})$ such that $\mathbf{G}_{\tilde{J}-1}=Q_N\mr{e}^{\mathbf{L}_{J_{\tilde{J}-2}+1}}\cdots \mr{e}^{\mathbf{L}_{J_{\tilde{J}-1}}}\mr{e}^{\mathbf{L}_{J_{\tilde{J}-1}+1}}\cdots \mr{e}^{\mathbf{L}_{J_1}}Q_N^*v=Q_N\mr{e}^{\mathbf{L}_{J_{\tilde{J}-2}+1}}\cdots \mr{e}^{\mathbf{L}_{J_{\tilde{J}-1}}}Q_N^*\mathbf{G}_{\tilde{J}}$.
We continue to apply Lemma~\ref{lem:transitive} to obtain the result.
\end{proof}

\begin{mythm}[Lemma~\ref{lem:toeplitz_diagonal}]
Assume $N\subset \mathbb{Z}^d\setminus\{0\}$.
Then, we have $\mathbb{C}^{N\times N}=T(N,\mathbb{C})$.
\end{mythm}
\begin{proof}
We show $\mathbb{C}^{N\times N}\subseteq T(N,\mathbb{C})$.
The inclusion $\mathbb{C}^{N\times N}\supseteq T(N,\mathbb{C})$ is trivial.
Since $N\subset \mathbb{Z}^d\setminus\{0\}$, for any $n=[n_1,\ldots,n_d]\in N$, there exists $k\in\{1,\ldots,d\}$ such that $\mr{i}n_k=(D_k)_{n,n}\neq 0$.
We denote by $k_{\opn{min}}(n)$ the minimal index $k\in\{1,\ldots,d\}$ that satisfies $(D_k)_{n,n}\neq 0$.
Let $B\in \mathbb{C}^{N\times N}$.
We decompose $B$ as $B=B_1+\ldots+ B_d$, where $(B_k)_{:,n}=B_{:,n}$ if $k=k_{\opn{min}}(n)$ and $(B_k)_{:,n}=\mathbf{0}$ otherwise.
Here, $(B_k)_{:,n}$ is the $n$th column of $B_k$.
Then, we have $(B_k)_{:,n}=\mathbf{0}$ if $(D_k)_{n,n}=0$.
Let $D_k^+$ be the diagonal matrix defined as $(D_k^+)_{n,n}=1/(D_k)_{n,n}$ if $(D_k)_{n,n}\neq 0$ and $(D_k^+)_{n,n}=0$ if $(D_k)_{n,n}= 0$.
In addition, let $C_k=B_kD_k^+$.
Then, we have $B=\sum_{k=1}^dC_kD_k$.
Applying Propostion~\ref{prop:toeplitz}, we have $B\in T(N,\mathbb{C})$, and obtain $\mathbb{C}^{N\times N}\subseteq T(N,\mathbb{C})$.
\end{proof}

\begin{mythm}[Lemma~\ref{lem:transitive}]
For any $\mathbf{u},\mathbf{v}\in\mathbb{C}^N\setminus \{0\}$, there exists $A\in GL(N,\mathbb{C})$ such that $\mathbf{u}=A\mathbf{v}$.
\end{mythm}
\begin{proof}
Let $n_0\in N$ and let
$B\in\mathbb{N\times N}$ be defined as $B_{n,:}=1/\Vert \mathbf{v}\Vert^2\mathbf{v}^*$ for $n=n_0$ and so that $B_{n,:}$ and $B_{m,:}$ becoming orthogonal if $n\neq m$.
Then, the $n$th element of $B\mathbf{v}$ is $1$ for $n=n_0$ and is $0$ for $n\neq n_0$.
Let $C\in\mathbb{N\times N}$ be defined as $C_{n,:}=\mathbf{u}$ for $n=n_0$ and so that $C_{n,:}$ and $C_{m,:}$ becoming orthogonal if $n\neq m$.
Then, $B,C\in GL(N,\mathbb{C})$ and $CB\mathbf{v}=\mathbf{u}$.
\end{proof}


\section{Details of Remark~\ref{rmk:equiv_transform} (Reduction to the analysis on $\mathbb{T}^d$)}\label{ap:equiv_transform}
Let $B_d$ be the unit ball in $\mathbb{T}^d$.
If $\mcl{X}$ is diffeomorphic to $B_d$, then we can construct a dynamical system $\check{f}_j$ on $\mathbb{T}^d$ that satisfies $\check{f}_j(x)=\tilde{f}_j(x)$ for $x\in B_d$, where $\tilde{f}_j$ is the equivalent dynamical system on $B_d$ with $f_j$.
Indeed, let $B_d=\{x\in\mathbb{R}^d\,\mid\,\Vert x\Vert\le 1\}$ be the unit ball.
Let $\psi:\mcl{X}\to B_d$ be the diffeomorphism, and let $y=\psi(x)$.
Then, the dynamical system $\frac{\mathrm{d}x}{\mathrm{d}t}(t)=f_j(x(t))$ is equivalent to $\frac{\mathrm{d}y}{\mathrm{d}t}(t)=J{\psi}(y(t))^{-1}f_j(y(t))$ since $J\psi(y)$ is invertible for any $y\in B_d$, where $J\psi$ is the Jacobian of $\psi$.
Note that since $J\psi$ does not depend on $j$, the transition of $\tilde{f}_j$ over $j$ depends only on that of $f_j$ over $j$.
Let $\tilde{f}_j(y)=J{\psi}(y)^{-1}f_j(y)$.
Instead of considering the dynamical system $f_j$ on $\mcl{X}$, we can consider the dynamical system $\tilde{f}_j$ on $B_d$.
Let $a$ be a positive real number satisfying $1<a<\pi$.
Then, we can smoothly extend $\tilde{f}_j$ on $B_d$ to a map $\hat{f}_j$ on $aB_d$ as $\hat{f}_j(x)=\tilde{f}_j(x)\ (x\in B_d)$, $\hat{f}_j(x)=0\ (\Vert x\Vert=a)$.
For example, we can construct $\hat{f}_j$ in the same manner as a smooth bump function~\citep{tu08}.
Finally, we extend $\hat{f}_j$ on $aB_d$ to a map $\check{f}_j$ on $[-\pi,\pi]^d$ as $\check{f}_j(x)=\hat{f}_j(x)\ (x\in aB_d)$, $\check{f}_j(x)=0\ (x\notin aB_d)$.
Then, since $\check{f}_j([-\pi,\ldots,-\pi])=\check{f}_j([\pi,\ldots,\pi])$, we can regard $\check{f}_j$ as a dynamical system on $\mathbb{T}^d$.

In practical computations, we do not need to identify $\mathcal{X}$ or transform the dynamical systems. 
The transformation of the dynamical system is to align the model to the setting that can be more easily theoretically analyzed. 
In addition, the only part of the practical algorithm that needs the assumption of the existence of $\mathcal{X}$ is the computation of the inner product between the Fourier function and the fixed function $v$ (See line 1 in Algorithm~\ref{al:deep_koopman}). 
The computation of the inner product requires the integral of a function on $\mathcal{X}$. 
However, we can approximate the integral using a discrete sum over given training samples. In this case, we do not need to determine $\mathcal{X}$ explicitly.

\section{Connection with other methods}\label{sec:connection}
\paragraph{Deep Koopman-layered model as a neural ODE-based model}
The model~\eqref{eq:model_flows} can also be regarded as a model with multiple neural ODEs~(\citealp{teshima20_neuralode}; \citealp[Section 3.3]{li23}).
From this perspective, we can also apply the model to standard tasks with ResNet.
For existing neural ODE-based models, we solve ODEs for the forward computation and solve adjoint equations for backward computation~\citep{chen18,sholokhov23}.
In our framework, solving the ODE corresponds to computing $\mr{e}^{\mathbf{L}_j}u$ for a matrix $\mathbf{L}_j$ and a vector $u$.
As we stated in Subsection~\ref{subsec:approximation_koopman}, we use Krylov subspace methods to compute $\mr{e}^{\mathbf{L}_j}u$.
In this sense, our framework provides a numerical linear algebraic way to solve neural ODE-based models. 
In Appendix~\ref{ap:exp_neuralODE_addition}, we have numerical results that show the proposed model can be an alternative to neural ODE-based approaches.

\paragraph{Connection with neural network-based Koopman approaches}
In the framework of neural network-based Koopman approaches, we train an encoder $\phi$ and a decoder $\psi$ that minimizes $\Vert x_{t+1}-\psi(K\phi(x_t))\Vert$ for the given time-series
$x_0,x_1,\ldots$~\citep{lusch17,li17,azencot20,shi22}.
Here, $K$ is a linear operator, and we can construct $K$ using EDMD or can train $K$ simultaneously with $\phi$ and $\psi$.
Physics-informed frameworks of neural network-based Koopman approaches for incorporating the knowledge of dynamics have also been proposed~\citep{liu24_physics_informed}.
For neural network-based Koopman approaches, since the encoder $\phi$ changes along the learning process, the approximation space of the operator $K$ also changes.
Thus, the theoretical analysis of these approaches is challenging.
On the other hand, our deep Koopman-layered approach fixes the approximation space using the Fourier functions and learns only the linear operators corresponding to Koopman generators, which enables solid theoretical analyses. 

\section{Details of Remark~\ref{rmk:convergence_rate}}\label{ap:convergence_rate}
In the same manner as Theorem~\ref{thm:universality}, we can show that we can represent any function in $V_N=\operatorname{Span}\{q_n\,\mid\,n\in N\}$ exactly using the deep Koopman-layered model.
Thus, if the decay rate of the Fourier transform of the target function $h$ is $\alpha$, i.e., if there exist $0<\alpha<1$ such that $h$ is represented as $h=\sum_{n\in\mathbb{Z}^d}c_nq_n$ with some $c_n\in\mathbb{C}$ satisfying $\vert c_n\vert \le \alpha^{n_1+\cdots +n_d}$ for sufficiently large $n$, then the convergence rate with respect to $N$ is $O(({1-\alpha^2})^{-d/2})$.
Indeed, for sufficiently large $N$, we have 
\begin{align*}
\min_{\tilde{h}\in V_N}\Vert h-\tilde{h}\Vert&=\bigg\Vert \sum_{n\notin N}c_nq_n\bigg\Vert =\bigg(\sum_{n\notin N}\vert c_n\vert^2\bigg)^{1/2}\\
&\le \bigg(\sum_{n\notin N}\alpha^{2(n_1+\cdots +n_d)}\bigg)^{1/2}=O\bigg(\bigg(\frac{1}{1-\alpha^2}\bigg)^{d/2}\bigg).
\end{align*}

\section{Details on deriving the generalization bound}\label{ap:generalization_details}
\subsection{Background: Reproducing kernel Hilbert space}
Let $\kappa:\mathcal{X}\times \mathcal{X}\to\mathbb{C}$ be a positive definite kernel, which satisfies the following two conditions: 

\begin{enumerate}[leftmargin=*,nosep]
    \item ${\kappa}(x,y)=\overline{\kappa(y,x)}$\; for $x,y\in\mathcal{X}$,\quad
    \item  $\sum_{n,m=1}^{N}\!\overline{c_n}c_m\kappa(x_n,x_m)\!\ge\! 0$\; for $N\in\mathbb{N}$, $c_n\in\mathbb{C}$, $x_n\in\mathcal{X}$.
\end{enumerate}

Let ${\phi}$ be a feature map defined as $\phi(x)=\kappa(\cdot,x)$.
The reproducing kernel Hilbert space (RKHS) $\hil_{\kappa}$ is the Hilbert space spanned by $\{\phi(x)\,\mid\,x\in\mathcal{X}\}$.
The inner product $\bracket{\cdot,\cdot}:\hil_{\kappa}\times \hil_{\kappa}\to\mathbb{C}$ in $\hil_{\kappa}$ is defined as
\begin{equation*}
\Bbracket{\!\sum_{n=1}^{N}c_n{\phi}(x_n),\sum_{m=1}^{M}d_m{\phi}(y_m)\!}\!=\!\sum_{n=1}^{N}\sum_{m=1}^{M}\overline{c_n}d_m\kappa(x_n,y_m)
\end{equation*}
for $c_n,d_n\in\mathbb{C}$ and $x_n,y_n\in\mathcal{X}$.
Note that by the definition of $\kappa$, $\bracket{\cdot,\cdot}$ is well-defined and satisfies the axiom of inner products.
An important property for RKHSs is the reproducing property. 
For $x\in \mathcal{X}$ and $v\in \hil_{\kappa}$, we have $\bracket{\phi(x),v}=v(x)$, which is useful for deriving a generalization bound.

\subsection{Derivation of the generalization bound}
We use the Rademacher complexity to derive Proposition~\ref{prop:generalization}. 
For this purpose, we regard the model~\eqref{eq:model_flows} as a function in an RKHS.
For $j\in\mathbb{Z}^d$ and $x\in\mathbb{T}^d$, let $\tilde{q}_j(x)=\mr{e}^{-\tau\Vert j\Vert_1}\mr{e}^{\mr{i}j\cdot x}$, where $\tau>0$ is a fixed parameter.
Let $\kappa(x,y)=\sum_{j\in\mathbb{Z}^d}\overline{\tilde{q}_j(x)}\tilde{q}_j(y)$, and consider the RKHS $\hil_{\kappa}$ associated with the kernel $\kappa$.
Note that $\kappa$ is a positive definite kernel, and $\{\tilde{q}_j\,\mid\,j\in\mathbb{Z}^d\}$ is an orthonormal basis of $\hil_{\kappa}$.
\citet{giannakis22} and \citet{das21} used this kind of RKHSs for simulating dynamical systems on a quantum computer based on Koopman operator theory and for approximating Koopman operators by a sequence of compact operators.
Here, we use the RKHS $\hil_{\kappa}$ for deriving a generalization bound.
To regard the function $\mathbf{G}\in V_N=\opn{Span}\{q_j\,\mid\,j\in N\}\subset L^2(\mathbb{T}^d)$ as a function in $\hil_{\kappa}$, we define an inclusion map $\iota_N:V_N\to\hil_{\kappa}$ as $\iota_N q_j=\mr{e}^{\tau\Vert j\Vert_1}\tilde{q}_j$ for $j\in N$.
Then, the operator norm of $\iota_N$ is $\Vert\tau_N\Vert=\max_{j\in N}\mr{e}^{\tau\Vert j\Vert_1}$.

Let $S\in\mathbb{N}$ be the sample size, $\sigma_1,\ldots,\sigma_S$ be i.i.d. Rademacher variables (i.e., random variables that follow uniform distribution over $\{\pm 1\}$), and $x_1,\ldots,x_S$ be given samples.
Then, the empirical Rademacher complexity $\hat{R}_S(\mcl{G}_N)$ is bounded as follows.
\begin{lemma}\label{lem:rademacher}
We have
\begin{align*}
\hat{R}_S(\mcl{G}_N)\le \frac{\alpha}{\sqrt{S}}\max_{j\in N}\mr{e}^{\tau\Vert j\Vert_1}\!\!\!\!\!\!\!\!\!\!\!\!\sup_{\mathbf{L}_1,\ldots,\mathbf{L}_J\in T(N,\mathbb{C})}\!\!\!\!\Vert \mr{e}^{\mathbf{L}_1}\Vert\cdots \Vert\mr{e}^{\mathbf{L}_J}\Vert\,\Vert v\Vert.
\end{align*}
\end{lemma}
\begin{proof}
\begin{align*}
\hat{R}_S(\mcl{G}_N)&=\frac{1}{S}\mr{E}\bigg[\sup_{\mathbf{G}\in\mcl{G}_N}\sum_{s=1}^S\mathbf{G}(x_s)\sigma_s\bigg]
=\frac{1}{S}\mr{E}\bigg[\sup_{\mathbf{G}\in\mcl{G}_N}\sum_{s=1}^S\iota_N\mathbf{G}(x_s)\sigma_s\bigg]\\
&=\frac{1}{S}\mr{E}\bigg[\sup_{\mathbf{G}\in\mcl{G}_N}\Bbracket{\sum_{s=1}^S\sigma_s\phi(x_s),\iota_N\mathbf{G}}\bigg]
\le \frac{1}{S}\sup_{\mathbf{G}\in\mcl{G}_N}\Vert \iota_N\mathbf{G}\Vert_{\hil_K} \bigg(\sum_{s=1}^SK(x_s,x_s)\bigg)^{1/2}\\
&\le \frac{\alpha}{\sqrt{S}}\sup_{\mathbf{G}\in\mcl{G}_N}\Vert \iota_N\Vert \Vert\mathbf{G}\Vert_{L^2(\mathbb{T}^d)}
\le \frac{\alpha}{\sqrt{S}}\max_{j\in N}\mr{e}^{\tau\Vert j\Vert_1}\sup_{\mathbf{L}_1,\ldots,\mathbf{L}_J\in T(N,\mathbb{C})}\Vert Q_N\mr{e}^{\mathbf{L}_1}\cdots \mr{e}^{\mathbf{L}_J}Q_N^*v\Vert\\
&\le \frac{\alpha}{\sqrt{S}}\max_{j\in N}\mr{e}^{\tau\Vert j\Vert_1}\sup_{\mathbf{L}_1,\ldots,\mathbf{L}_J\in T(N,\mathbb{C})}\Vert \mr{e}^{\mathbf{L}_1}\Vert\cdots \Vert\mr{e}^{\mathbf{L}_J}\Vert\,\Vert v\Vert,
\end{align*}
where $\alpha=\sum_{j\in\mathbb{Z}^d}\mr{e}^{-2\tau\Vert j\Vert_1}$.
\end{proof}

Combining Lemma 4.2 in~\citet{mohri18} and Lemma~\ref{lem:rademacher}, we derive Proposition~\ref{prop:generalization}.

\section{Details of Remark~\ref{rmk:regularization}}\label{ap:boundedness}
In our setting, we assume that the flow $g(t,\cdot)$ is invertible and the Jacobian $Jg_t^{-1}$ of $g_t^{-1}$ is bounded for any $t$. Here, we denote $g_t=g(t,\cdot)$.
In this case, the Koopman operator $K^t$ is bounded.
Indeed, we have
\begin{align*}
\Vert K^th\Vert^2=\int_{\mathbb{T}^d}\vert h(g(t,x))\vert^2 dx = \int_{\mathbb{T}^d}\vert h(x)\vert^2 \vert \operatorname{det}Jg_t^{-1}(x)\vert dx \le \Vert h\Vert^2 \sup_{x\in \mathbb{T}^d}\vert\operatorname{det}Jg_t^{-1}(x)\vert.
\end{align*}

\section{Algorithmic details of training deep Koopman-layered model}~\label{ap:pseudocode}
{Based on Corollary~\ref{cor:universality_time_series}, we train the deep Koopman-layered model using time-series data as follows:
We first fix the auxiliary transform $v$ in the model $\mathbf{G}$ taking Remark~\ref{rmk:d+1} into account, the number of layers $\tilde{J}$, and the index sets $N$, $M_r$.
We input a family of time-series data $\{(x_{s,0},\ldots,x_{s,\tilde{J}})\}_{s=1}^S$ to $\mathbf{G}$.
For obtaining the output of $\mathbf{G}$, we first compute $Q_N^*v=[\bracket{q_n,v}]_n$, where $\bracket{\cdot,\cdot}$ is the inner product in $L^2(\mathbb{T}^d)$, and compute $\mr{e}^{\mathbf{L}_J}Q_N^*v$ using the Krylov subspace method, where $J=J_{\tilde{J}}$, $\mathbf{L}_J=t_J\sum_{k=1}^dA_1^{k}\cdots A_R^kD_k$, and $A_r^k=[a^{k,J}_{n-l,r}]_{n,l}$ is the Toeplitz matrix.
In the same manner, we compute $\mr{e}^{\mathbf{L}_{J-1}}(\mr{e}^{\mathbf{L}_J}Q_N^*v)$. 
We continue that and finally obtain the output $\mathbf{G}(x)=Q_Nu(x)=\sum_{n\in N}q_n(x)u_n$, where $u=[u_1,\ldots,u_n]^T=\mr{e}^{\mathbf{L}_{1}}\cdots \mr{e}^{\mathbf{L}_J}Q_N^*v$.}
We learn the parameter $a_{m,r}^{k}$ for each layer in $\mathbf{G}$ by minimizing $\sum_{s=1}^S\ell(v(x_{s,\tilde{J}}),\mathbf{G}_j(x_{s,j-1}))$ for $j=1,\ldots,\tilde{J}$ using an optimization method.
For example, we can set an objective function $\sum_{j=1}^{\tilde{J}}\sum_{s=1}^S\ell(v(x_{s,\tilde{J}}),\mathbf{G}_j(x_{s,j-1}))$.
Here $\ell:\mathbb{C}\times \mathbb{C}\to\mathbb{R}$ is a loss function.
For example, we can set $\ell$ as the squared error.

We provide a pseudocode of the algorithm of training the deep Koopman-layered model in Algorithm~\ref{al:deep_koopman}.
Let $q_n$ be the Fourier function defined as $q_n(z)=\mr{e}^{\mr{i}n\cdot z}$ for $n\in\mathbb{Z}^d$ and $z\in\mathbb{T}^d$, and let $\bracket{\cdot,\cdot}$ be the inner product in $L^2(\mathbb{T}^d)$.
Thus, $\bracket{q_n,v}$ means the $n$th Fourier coefficient of a function $v$.
We fix an auxiliary nonlinear map $v\in L^2(\mathbb{T}^d)$ in the model $\mathbf{G}$.
We also fix the finite index set $N\subseteq \mathbb{N}^d$ determining the approximation space of the Koopman generators, number of layers $J\in\mathbb{N}$, the number $R_j\in\mathbb{N}$ of Toeplitz matrices, index sets $M^j_1,\ldots,M^j_{R_j}\subseteq \mathbb{Z}^d$ determining the sparseness of the Toeplitz matrices for the $j$th layer, and the loss function $\ell:\mathbb{C}\times \mathbb{C}\to\mathbb{R}_+$.
They determine the model architecture.
Let $A_{r}^{k,j}=[a^{k,j}_{n-l,r}]_{n,l\in N, n-l\in M_r^j}$ be the Toeplitz matrix with learnable parameters $a^{k,j}_{n,r}$ and $D_k$ be the diagonal matrix with $(D_k)_{l,l}=\mr{i}l_k$ for $l\in\mathbb{Z}^d$. 
In addition, we put all the learnable parameters $A=[a_{k,j}^{n,r}]_{k=1,\ldots,d,n\in N\bigcap M_r^j,r=1,\ldots,R_j,j=1,\ldots\tilde{J}}$.
For simplicity, we focus on the case of the number of layers $J$ is equal to the time step $\tilde{J}$.
We note that the time $t$ in the definition of $\mathbf{L}$ in Subsection~\ref{subsec:approximation_koopman} do not need for practical learning algorithm since it is just regarded as the scale factor of the learnable parameter $A_1^k$.

\subsection{Determining an optimal number $J$ of layers}
Although providing thorough discussion of determining an optimal number $J$ of layers is future work, we provide examples of heuristic approaches to determining $J$. 
Heuristically, we can use validation data to determine an optimal number of layers.
For example, we begin by one layer and compute the validation loss.
Then, we set two layers and compute the validation loss, and continue with more layers.
We can set the number of layers as the number that achieves the minimal validation loss.
Another way is to set a sufficiently large number of layers and train the model with the validation data.
As we discussed in Section 6.2, we can add a regularization term to the loss function so that the Koopman layers next to each other become close.
After the training, if there are Koopman layers next to each other and sufficiently close, then we can regard them as one Koopman layer and determine an optimal number of layers.
In Appendix~\ref{ap:time_forecast_additional}, we empirically show the proposed model is robust with respect to the choice of $J$.

\section{Application to time-series forecasting}\label{ap:forecast}
\begin{figure}
    \centering
    \includegraphics[width=0.9\linewidth]{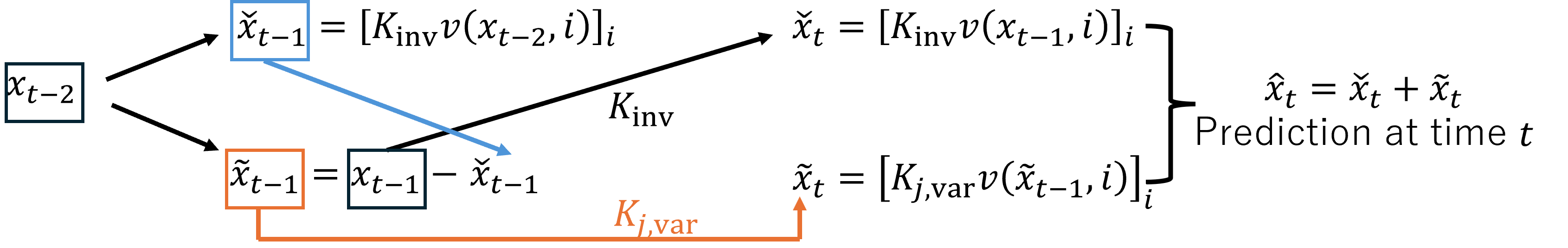}
    \caption{Overview of the computation of the prediction $\hat{x}_t$ using $x_{t-2}$ and $x_{t-1}$.}
    \label{fig:overview_forecasting}
\end{figure}

Assume that we have a time-series $x_0,\ldots,x_{t_J}\in\mathcal{X}$.
We first construct $K_{\opn{inv}}$ as explained in Subsection~\ref{subsec:exp_forecast}.
Note that since we construct the approximation of the Koopman generator, we can obtain $K_{\opn{inv}}=K_{\opn{inv}}^{\Delta t}$ for any $\Delta t$.
Indeed, let $t_j$ the time interval between the $j-1$th and $j$th Koopman layers.
Then, we obtain the Koopman operator for time interval $\Delta t$ by $\mr{e}^{\mathbb{L}_j\Delta t/t_j}$.
Let $s<t_J$ be a time index that is close to $t_J$.
We set $\tilde{x}_{t,i}=x_t-K_{\opn{inv}}v(x_{t-1},i)$ for $t=s,\ldots t_J$, where $x_{t,i}$ is the $i$th element of $x_t$ and $v$ is a map that satisfies $v(x_t,i)=x_{t,i}$.
We estimate the Koopman operator underlying the time-series $\{\tilde{x}_t\}_{t=s}^{t_J}$.
Given $x_{t-2}$ and $x_{t-1}$, we first compute $\tilde{x}_{t-1,i}=x_{t-1,i}-K_{\opn{inv}}v(x_{t-2},i)$.
Then, $x_t$ is estimated by $K_{j,\opn{var}}v(\tilde{x}_{t-1},i)+K_{\opn{inv}}v(x_{t-1},i)$.
Figure~\ref{fig:overview_forecasting} schematically shows how to compute the prediction $\hat{x}_t$ of $x_t$ using the previous time data $x_{t-2}$ and $x_{t-1}$.

\section{Experimental details}\label{ap:exp_detail}
We present details of the experiments in Section~\ref{sec:num_exp}.
All the experiments in this paper were executed with Python 3.9 on an Intel(R) Core(TM) i9-10885H
2.4GHz processor with the Windows 10 operating system.

\subsection{Representation power and generalization}\label{ap:exp_detail_rep}
We discretized the dynamical system with the time-interval $\Delta t=0.01$, and generated 1000 time-series $\{x_{s,0},\ldots x_{s,100}\}$ for $s=1,\ldots,1000$ with different initial values distributed uniformly on $[-1,1]\times [-1,1]$.
We added a random noise, which was drawn from the normal distribution of mean 0 and standard deviation 0.01, to each $x_{s,j}$ and set it as $\tilde{x}_{s,j}$.
For training, we used the pairs $\{{x}_{s,0},\tilde{x}_{s,100}\}$ for $s=1,\ldots,1000$.
Then, we trained deep Koopman-layered models on $\mathbb{T}^3$ by minimizing the loss $\sum_{s=1}^{1000}\Vert [Q_N\mr{e}^{\mathbf{L}_1}\cdots\mr{e}^{\mathbf{L}_J}Q_N^*v(\tilde{x}_{s,0},\pi/2),Q_N\mr{e}^{\mathbf{L}_1}\cdots\mr{e}^{\mathbf{L}_J}Q_N^*v(\tilde{x}_{s,0},0)]-\tilde{x}_{s,100}\Vert^2$ using the Adam optimizer~\citep{kingma15} with the learning rate $0.001$.
We created data for testing in the same manner as the training dataset.
We set $v(x,y)=\sin(y)x_1+\cos(y)x_2$ for $x=[x_1,x_2]\in\mathbb{T}^2$ and $y\in\mathbb{T}$.
Note that based on Remark~\ref{rmk:d+1}, we constructed Kooman-layers on $\mathbb{T}^{d+1}$ for the input dimension $d$, and we {designed the function $v$ so that it} recovers $x_1$ by $v(x,\pi/2)$ and $x_2$ by $v(x,0)$.
{We used the sine and cosine functions for designing $v$ since the approximation space is constructed with the Fourier functions.}
We set $N=\{n=[n_1,n_2,n_3]\in\mathbb{Z}^3\,\mid\,-5\le n_1,n_2,n_3\le 5\}\setminus\{0\}$, $R=1$, and $M_1=\{n=[n_1,n_2,n_3]\in\mathbb{Z}^3\,\mid\,-2\le n_1,n_2\le 2, -1\le n_3\le 1\}\setminus\{0\}$ for all the layers.
We applied the Arnoldi method~\citep{saad92} to compute the exponential of $\mathbf{L}_j$.

For the experiment on the generalization property (Figure~\ref{fig:layers} (b)), we generated training data as above, but the sample size was 30, and the standard deviation of the noise was 0.03.
We used the test data without the noise.
The sample size of the test data was 1000.

\subsection{Eigenvalues of the Koopman-layers for
nonautonomous systems}\label{ap:exp_detail_eig}
We discretized the dynamical systems with the time-interval $\Delta t=0.01$, and generated 1000 time-series $\{x_{s,0},\ldots x_{s,119}\}$ for $s=1,\ldots,1000$ for training with different initial values distributed uniformly on $[-1,1]\times [-1,1]$.
We split the data into 6 subsets $S_t=\{x_{s,j}\,\mid\,s\in\{1,\ldots,1000\},\ j\in\{20t,\ldots,20(t+1)-1\}\}$ for $t=0,\ldots,5$.
Then, we trained the model with 5 Kooman-layers on $\mathbb{T}^3$ by minimizing the loss $$\sum_{j=1}^5\sum_{s=1}^{1000}\sum_{l=0}^{19}\Vert [Q_N\mr{e}^{\mathbf{L}_j}\cdots\mr{e}^{\mathbf{L}_5}Q_N^*v(x_{s,20(j-1)+l},\pi/2),\ Q_N\mr{e}^{\mathbf{L}_j}\cdots\mr{e}^{\mathbf{L}_5}Q_N^*v(x_{s,20(j-1)+l},0)]-x_{s,100+l}\Vert^2$$ 
using the Adam optimizer with the learning rate $0.001$.
In the same manner as Subsection~\ref{subsec:vanderPol}, we set $v(x,y)=\sin(y)x_1+\cos(y)x_2$ for $x=[x_1,x_2]\in\mathbb{T}^2$ and $y\in\mathbb{T}$.
We set $N=\{n=[n_1,n_2,n_3]\in\mathbb{Z}^3\,\mid\,-5\le n_1,n_2\le 5, -2\le n_3\le 2\}$, $R=1$, and $M_1=\{n=[n_1,n_2,n_3]\in\mathbb{Z}^3\,\mid\,-2\le n_1,n_2\le 2, -1\le n_3\le 1\}$ for all the layers.
We applied the Arnoldi method to compute the exponential of $\mathbf{L}_j$.

For KDMD, we transformed $[x_1,x_2]\in\mathbb{T}^2$ into $\tilde{x}=[\mr{e}^{\mr{i}x_1},\mr{e}^{\mr{i}x_2}]\in\mathbb{C}^2$ and applied the Gaussian kernel $\kappa(x,y)=\mr{e}^{-0.1\Vert \tilde{x}-\tilde{y}\Vert^2 }$. 
For estimating $K^{t_j}_j$, we applied the principal component analysis to the space spanned by $\{\kappa(\cdot,x)\,\mid\,x\in S_{j-1}\}$ to obtain $\vert N\vert$ principal vectors $p_1,\ldots,p_{\vert N\vert}$.
We estimated $K^{t_j}_j$ by constructing the projection onto the space spanned by $p_1,\ldots,p_{\vert N\vert}$.

\subsection{Application to time-series forecasting}\label{ap:exp_detail_forecast}
We set $J=9$ and split the training data into 10 sub-datasets as $\{x_0,\ldots,x_{T-1}\},\{x_{T},\ldots, x_{2T-1}\},\ldots,\{x_{9T},\ldots,x_{10T-1}\}$.
We constructed the family of input time-series data explained in Subsection~\ref{subsec:training} by setting $x_{s,j}$ in Subsection~\ref{subsec:training} as $x_{jT+s}$.
For the deep Koopman-layered model, we set $N=\{(n,\ldots,n)\,\mid\,-10\le n\le 10\}$, $R=1$, and $M=\{(n,\ldots,n)\,\mid\,-5\le n\le 5\}$ for all layers.
In addition, we set $v(x,i)=x_i$ for $x=(x_1,\ldots,x_d)\in\mcl{X}$ and $i=1,\ldots,d$.
We trained the deep Koopman-layered model and constructed $K_{\opn{inv}}$ with $n_{\opn{inv}}=5$.
For Traffic, since the dimension of each sample is large (862-dimensional), we constructed functions $\alpha:\mcl{X}\to\tilde{\mcl{X}}\subset\mathbb{R}^7$ and $\beta:\tilde{\mcl{X}}\to\mcl{X}$ using fully connected 2-layered ReLU neural networks.
Then, we transformed the input data $x\in\mcl{X}$ into $\alpha(x)\in \tilde{\mcl{X}}$ and constructed the deep Koopman-layered model on $\tilde{\mcl{X}}$.
We transformed the output $y\in \tilde{\mcl{X}}$ of the deep Koopman-layered model into $\beta(y)\in\mcl{X}$, and set the loss function as the difference between $\beta(y)$ and the real observation.
The width of the first layer is 256.
We trained $\alpha$ and $\beta$ simultaneously with the learnable parameters in the deep Koopman-layered model. 
For the Fourier filter approach, we calculated
the averaged value of the Fourier components of the 10 training sub-dataset.
Then, we extracted the top 5 components and regard them as time-invariant components and constructed $K_{\opn{inv}}$.
We constructed $K_{\opn{var}}$ using the EDMD with learned dictionary functions and
the KDMD with the Laplacian kernel $\kappa(x_1,x_2)=\mr{e}^{-0.1\sum_{k=1}^d\vert x_{1,k}-x_{2,k}\vert}$.
For the EDMD, we used a fully connected 2-layered ReLU neural network to learn dictionary functions. 
The widths of the first and second layers
are $4\vert N\vert$ and $8\vert N\vert$ for Traffic and $\lfloor \vert N\vert (d+1)/2\rfloor$ and $\vert N\vert (d+1)$ for the other datasets so that the number of dictionary functions becomes equal to that of the deep Koopman-layered model.
For each dataset, we used the 70\% of the whole time-series data for the training.
In addition, for ETT, Electricity, Exchange, Weather, and ILI, since the scales of the values are different for each dimension of samples, we normalized the each dimension of samples so that it becomes mean 0 and standard deviation 1.
For the deep Koopman-layered model and the EDMD, we used the Adam optimizer with a learning rate of 0.001.
For both methods, the results are obtained after 3000 epochs of the training.

\section{Additional numerical results}~\label{ap:experiments}

\begin{figure}[t]
    \centering
    \subfigure[Neural ODE with one time step.]{\includegraphics[scale=0.3]{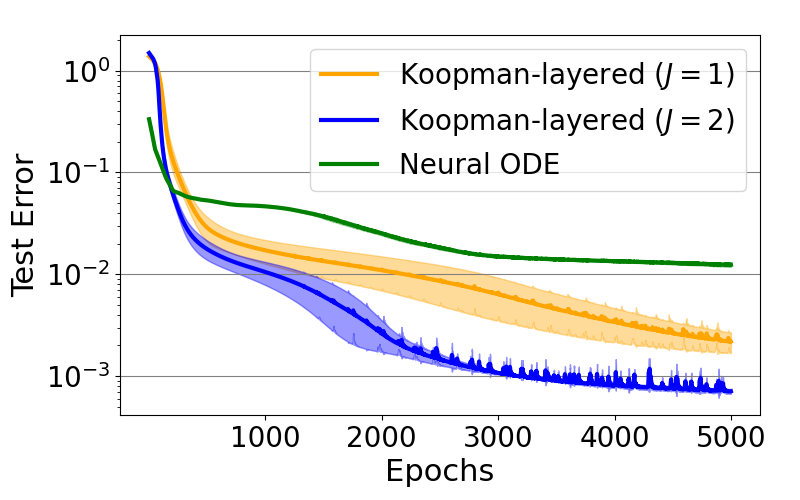}}
    \subfigure[Neural ODE with two time step data.]{\includegraphics[scale=0.3]{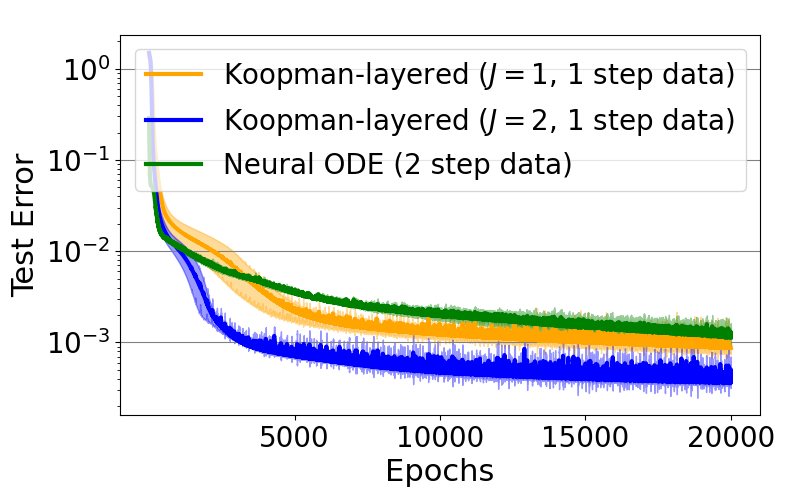}}
    \caption{Test errors for the deep Koopman layered models and the neural ODE. The result is the average $\pm$ the standard deviation of three independent runs.}
    \label{fig:neural_ode}
\end{figure}

\begin{figure}[t]
\begin{center}
\setlength{\tabcolsep}{3pt}
\subfigure[Common dictionary functions for five Koopman generators]{\begin{tabular}{ccccc}
\includegraphics[scale=0.2]{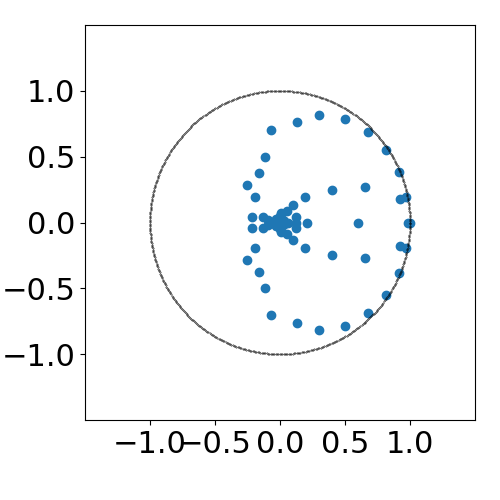}&
\includegraphics[scale=0.2]{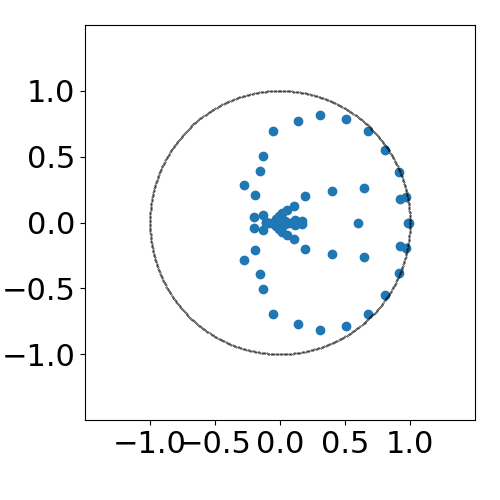}&
\includegraphics[scale=0.2]{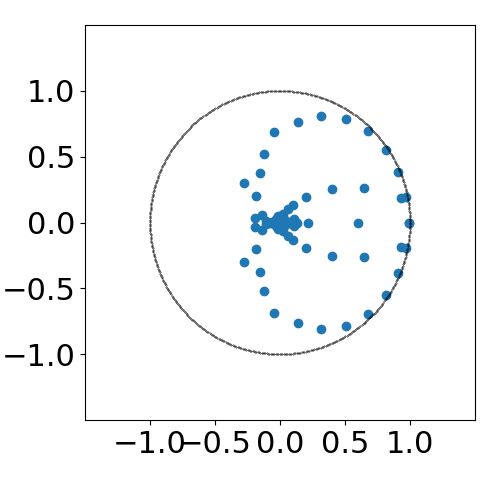}&
\includegraphics[scale=0.2]{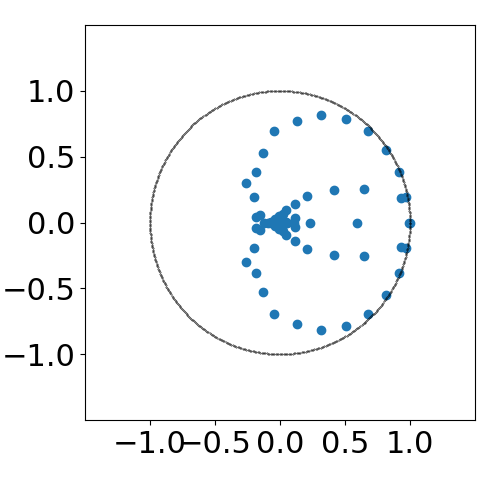}&
\includegraphics[scale=0.2]{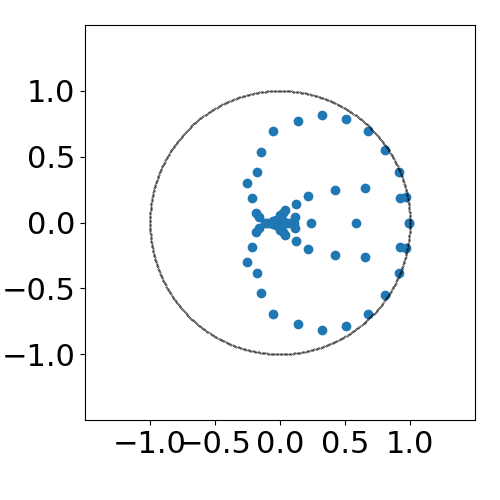}\\
$j=1$&$j=2$&$j=3$&$j=4$&$j=5$
\end{tabular}}

\subfigure[Separate dictionary functions for each Koopma generator]{\begin{tabular}{ccccc}
\includegraphics[scale=0.2]{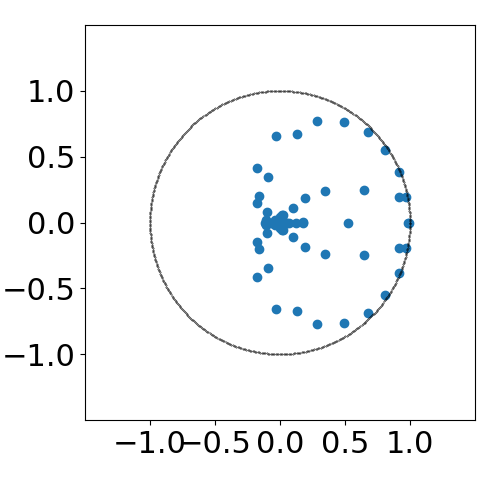}&
\includegraphics[scale=0.2]{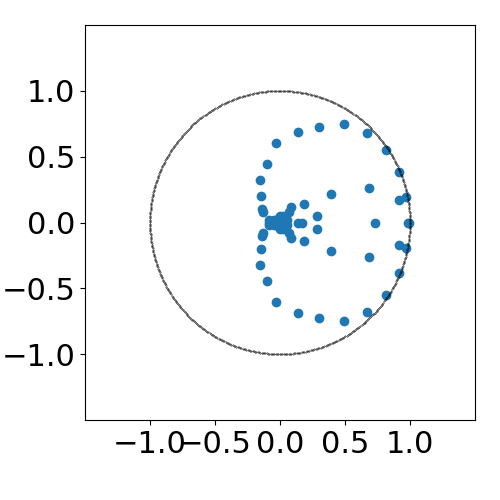}&
\includegraphics[scale=0.2]{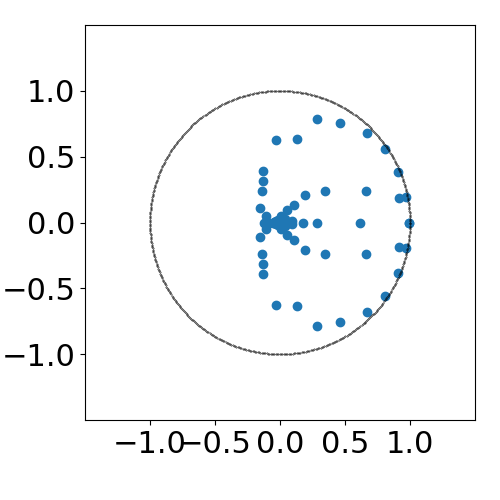}&
\includegraphics[scale=0.2]{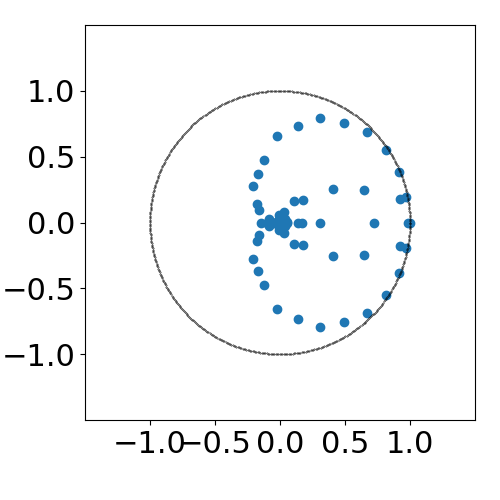}&
\includegraphics[scale=0.2]{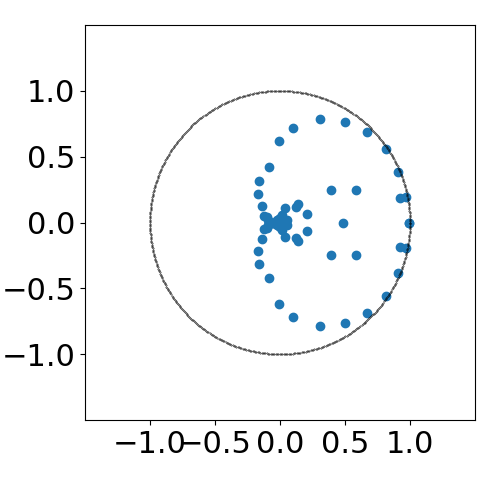}\\
$j=1$&$j=2$&$j=3$&$j=4$&$j=5$
\end{tabular}}

\subfigure[Separate dictionary functions for each Koopma generator with forward-backward extended DMD]{\begin{tabular}{ccccc}
\includegraphics[scale=0.2]{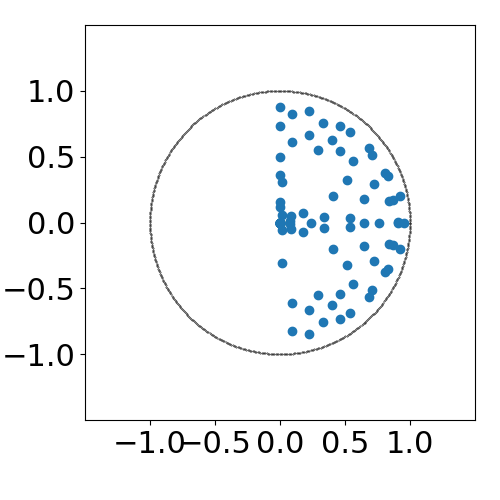}&
\includegraphics[scale=0.2]{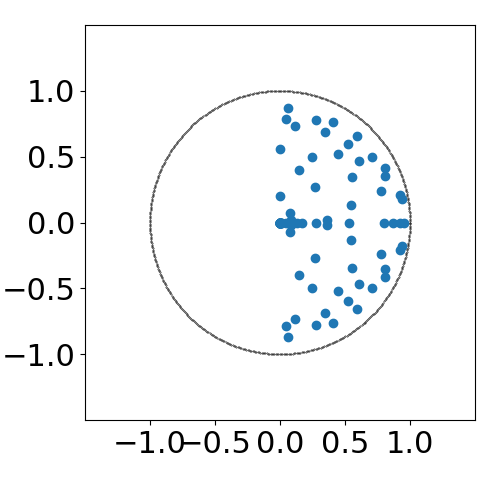}&
\includegraphics[scale=0.2]{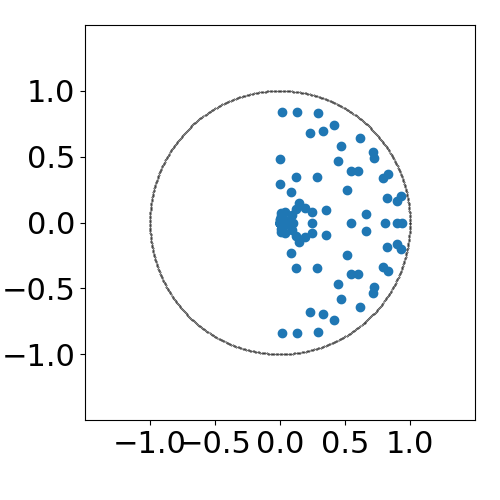}&
\includegraphics[scale=0.2]{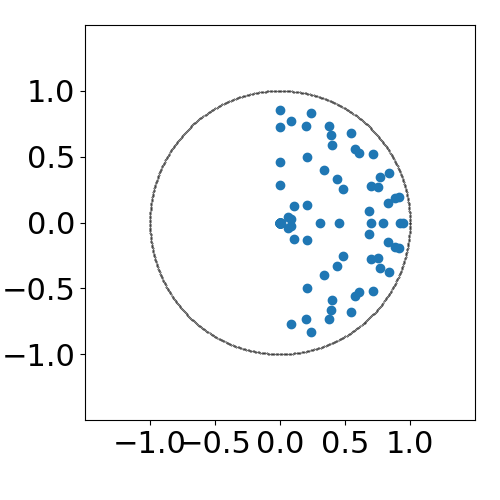}&
\includegraphics[scale=0.2]{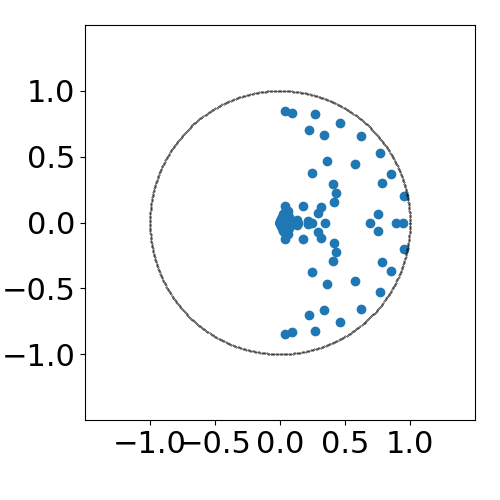}\\
$j=1$&$j=2$&$j=3$&$j=4$&$j=5$
\end{tabular}}
\caption{Eigenvalues of the estimated Koopman operators with learned approximation spaces for the nonautonomous damping oscillator.}\label{fig:additional}
\end{center}
\end{figure}

\subsection{Comparison to neural ODE}\label{ap:exp_neuralODE_addition}
To show that the proposed model can be an alternative to neural ODE-based approaches, we conducted additional experiments.
We applied a neural ODE~\citep{chen18} to the same problem as in Subsection~\ref{subsec:vanderPol} (the van der Pol oscillator).
The neural ODE is composed of the fully connected two-layer neural network with the hyperbolic tangent activation function whose width of the first layer is $55$.
The forward process is solved by the Runge-Kutta method.
We note that the number of parameters of this model is $2\times 55+55\times 2=220$, which is almost the same as the number of parameters of the Koopman-layered model considered in Subsection~\ref{subsec:vanderPol}, which is $222$ for the case of $J=2$.
To compare the basic performance of the two models, we used one time step data for training the neural ODE.
Note that in Subsection~\ref{subsec:vanderPol}, we also used only one time step data for the deep Koopman-layered model.
In the same manner as the deep Koopman-layered model, we used the Adam optimizer with a learning rate of 0.001.
The result is shown in Figure~\ref{fig:neural_ode} (a).
We can see that the deep Koopman-layered model outperforms the neural ODE model.

We can also use multi time step data for training the above neural ODE model.
Thus, we also used two time steps $\{{x}_{s,0},\tilde{x}_{s,50},\tilde{x}_{s,100}\}_{s=1}^{1000}$ to train the same neural ODE model and compared the performance with the deep Koopman-layered model.
We used the Adam optimizer with the learning rate 0.01.
The result is shown in Figure~\ref{fig:neural_ode} (b).
We can see that even if we use two time stap data for the neural ODE model, the deep Koopman-layered model with one time step data outperformed the neural ODE model.
The numerical methods applied in the neural ODE model perform well with multi time step data.
Thus, the performance of the neural ODE model with the two time step data is better than that with the one time step data.
On the other hand, for the Koopman-layered model, even with one time step data, we can estimate the Koopman operator well.
These results show that the Koopman-layered model has a potential power of being an alternative to neural ODE-based approaches.

\subsection{Comparison to Koopman-based approach with learned approximation spaces}\label{ap:exp_eig_addition}
We show the results of additional experiments with the Koopman-based approach with learned approximation spaces (see the second paragraph of Section~\ref{sec:connection}).
We considered the following two settings for the same example in Subsection~\ref{subsec:damping}.
\begin{enumerate}
    \item Learn a set of dictionary functions to construct the approximation space of five Koopman generators (learning a common set of dictionary functions is also considered by~\citep{liu23}).
    \item Learn five sets of dictionary functions each of which is for each Koopman generator.
\end{enumerate}
We used a 3-layered fully connected ReLU neural network to learn dictionary functions.
The widths of the first and second layers are 1024 and 121.
We applied the EDMD with the learned dictionary functions.
The result is illustrated in Figure~\ref{fig:additional} (a,b).
We cannot capture the transition of the distribution of the eigenvalues through $j=1,\ldots,5$ even though we learned the dictionary functions.
We can also see that there are some eigenvalues equally spaced on the unit circle.
This behavior is typical for autonomous systems with a constant frequency.
Since the dynamical system is nonautonomous and the frequency of the system changes over time, the above behavior is not suitable for this example.
This result implies that DMD-based methods try to capture the system as an autonomous system, which is not suitable for nonautonomous systems.
To obtain more stable eigenvalues, we also implemented the forward-backward extended DMD~\citep{lortie24} with the second setting.
The result is shown in Figure~\ref{fig:additional} (c), and it is similar to the above two cases.

\subsection{Comparison to models with general linear operators}\label{ap:eig_general_linear}
\begin{figure}[t]
    \centering
    \setlength{\tabcolsep}{3pt}
    \begin{tabular}{ccccc}
    \includegraphics[scale=0.2]{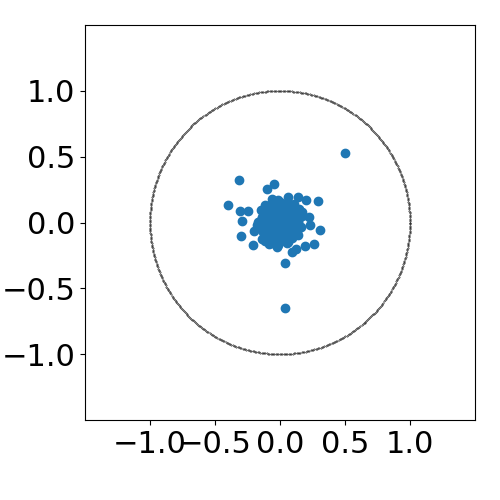}&
    \includegraphics[scale=0.2]{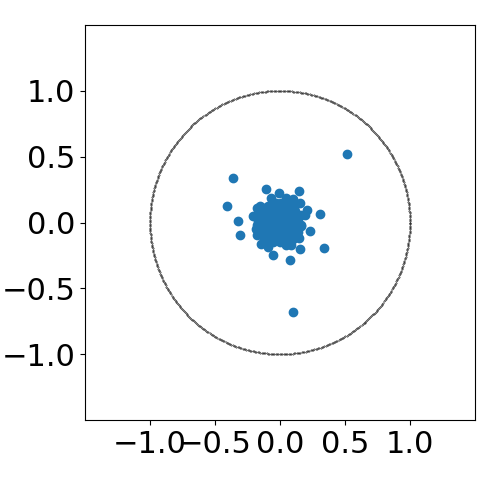}&
    \includegraphics[scale=0.2]{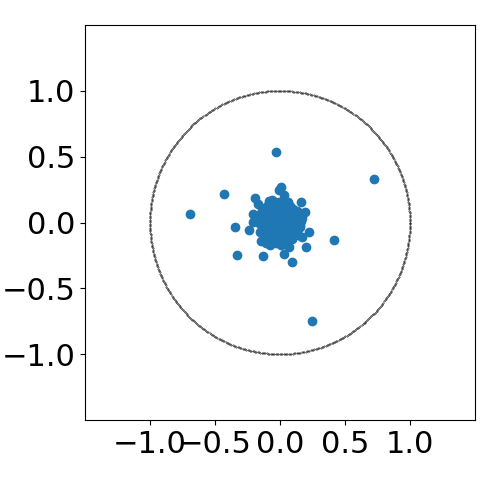}&
    \includegraphics[scale=0.2]{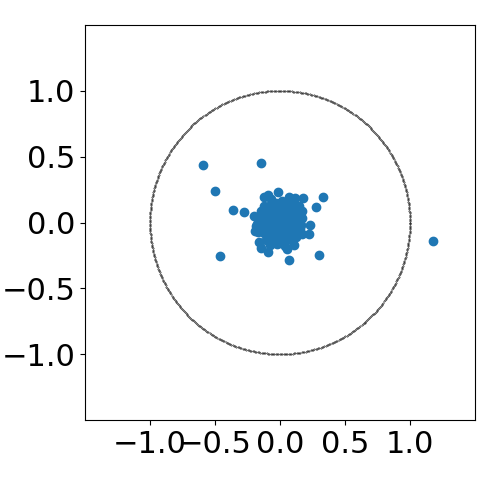}&
    \includegraphics[scale=0.2]{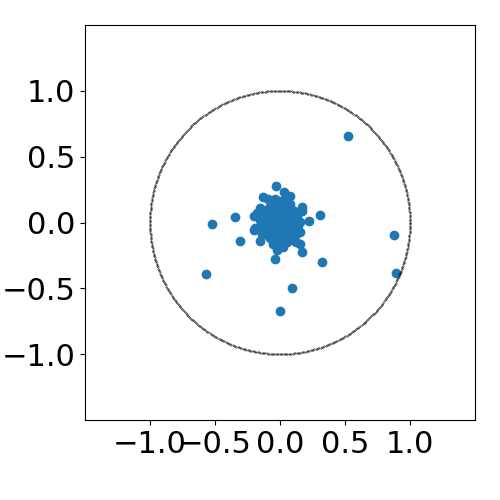}\\
    $j=1$&$j=2$&$j=3$&$j=4$&$j=5$
    \end{tabular}
\caption{Eigenvalues of the learned general matrices for the nonautonomous measure preserving system.}\label{fig:vol_general_linear}
\end{figure}

\begin{figure}[t]
    \centering
    \setlength{\tabcolsep}{3pt}
    \begin{tabular}{ccccc}
    \includegraphics[scale=0.2]{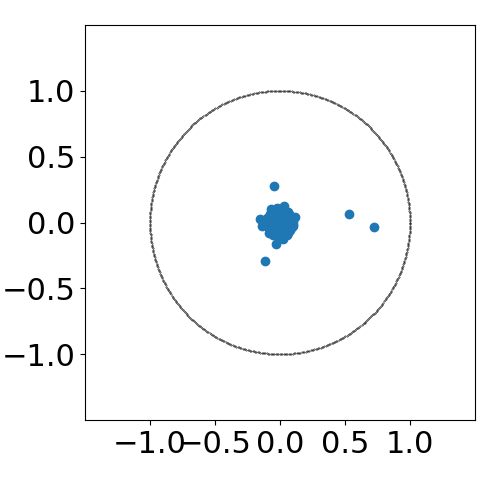}&
    \includegraphics[scale=0.2]{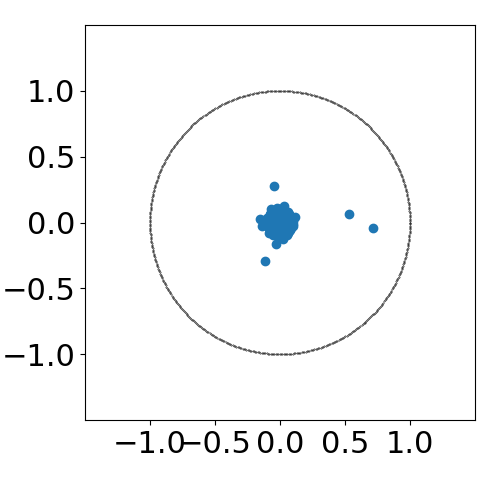}&
    \includegraphics[scale=0.2]{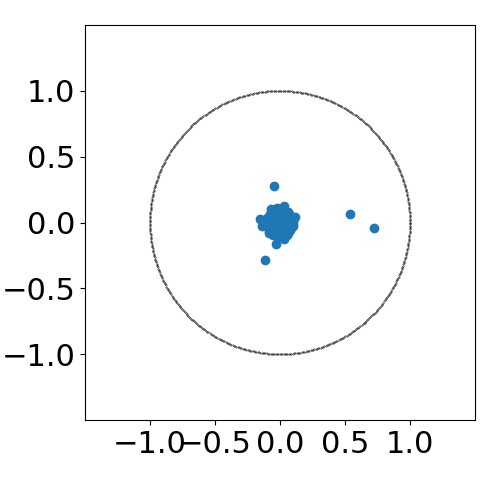}&
    \includegraphics[scale=0.2]{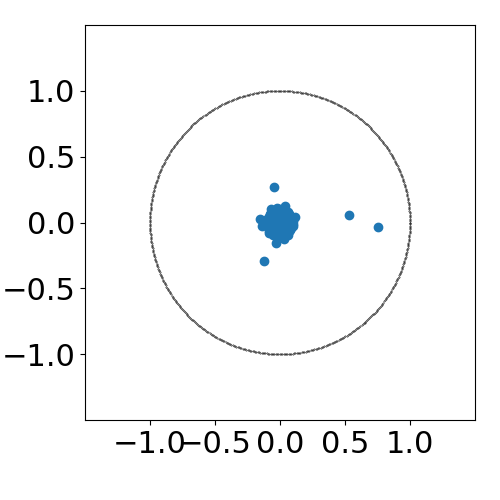}&
    \includegraphics[scale=0.2]{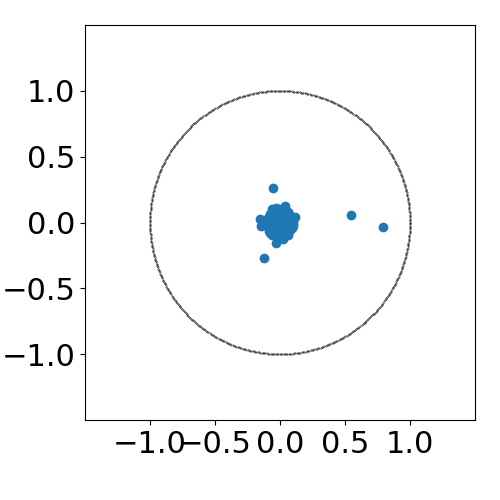}\\
    $j=1$&$j=2$&$j=3$&$j=4$&$j=5$
    \end{tabular}
\caption{Eigenvalues of the learned general matrices for the nonautonomous damping oscillator. } \label{fig:damping_general_linear}
\end{figure}

We can just use general linear operators instead of the Koopman operators to construct the model. However, in this case, we cannot relate that operator to the dynamical system, and cannot extract properties of the dynamical systems.
To empirically investigate the advantage of using the approximation \eqref{eq:generator_ap} over using general linear operators to represent the dynamical system, we conducted an additional experiment. 
We constructed a model by replacing each Koopman layer with a general matrix, and applied it to the same problems in Subsection~\ref{subsec:numexp_eig}.
The results are in Figures~\ref{fig:vol_general_linear} and \ref{fig:damping_general_linear}.
The figures show the eigenvalues of the matrix for the $j$th layer. 
Unlike the results of the proposed approach, it is difficult to interpret the eigenvalues of the learned operators, and they do not reflect the underlying dynamical system.


\subsection{Full results of the experiment in Subsection~\ref{subsec:exp_forecast}}\label{ap:full_exp_forecast}
We show the full results of the experiment in Subsection~\ref{subsec:exp_forecast} in Table~\ref{tab:forecast_full}.
ETT means ETTh2.
Since the dimension of each sample of Traffic is large, we could not implement the KDMD due to the memory shortage.

\begin{table}[t]
    \centering
    \caption{Relative squared error of the prediction with $\Delta T=48,96,144,196$. (The Average $\pm$ standard deviation for three independent runs for the deep Koopman layered model. We do not have any randomness for the Fourier filter with KDMD approach.) EDMD dict. means the EDMD with learned dictionary functions.}\vspace{.2cm}
    \label{tab:forecast_full}
    \begin{center}
    \centering
    $\Delta T=48$\vspace{.2cm}\\
    \begin{tabular}{c|c|c|c}
    \hline
    Dataset  &  Koopman-layered
    & Fourier filter (w. KDMD) & Fourier filter (w. EDMD dict.)\\
    \hline
    ETT & {\bf 0.0900$\pm$0.00924} & 2.25 & 0.0908$\pm$0.000935\\
    Electricity & {\bf 0.270$\pm$0.00143} & 0.717& 0.286$\pm$0.00151\\
    Exchange & {\bf 0.0560$\pm$0.00197}  &2.35 & 0.915$\pm$0.290\\
    Traffic & 0.482$\pm$0.0442 & --  & {\bf 0.433$\pm$0.000459} \\
    Weather & 0.569$\pm$0.00722 & 1.19 & {\bf 0.358$\pm$0.00130}\\
    ILI & {\bf 0.0858$\pm$0.0103} & 1.09 &  0.659$\pm$0.0257 \\
    \hline
    \end{tabular}\vspace{.5cm}\\
    \centering
    $\Delta T=96$\vspace{.2cm}\\
    \begin{tabular}{c|c|c|c}
    \hline
    Dataset  & Koopman-layered
    & Fourier filter (w. KDMD) & Fourier filter (w. EDMD dict.)\\
    \hline
    ETT & 0.137$\pm$0.0113 & 2.41 & {\bf 0.134$\pm$0.000631}\\
    Electricity & {\bf 0.217$\pm$0.00114} & 0.733 & {\bf 0.217$\pm$0.00149} \\
    Exchange & {\bf 0.0495$\pm$0.00103} & 2.94 & 1.03$\pm$0.217\\
    Traffic & {\bf 0.366$\pm$0.0303}  & --  & 0.401$\pm$0.00144 \\
    Weather &  0.189$\pm$0.00178 & 0.845 & {\bf 0.131$\pm$0.00106} \\
    ILI & {\bf 0.0896$\pm$0.00724} &0.971 & 0.662$\pm$0.0426 \\
    \hline
    \end{tabular}\vspace{.5cm}\\
    \centering
    $\Delta T=144$\vspace{.2cm}\\
    \begin{tabular}{c|c|c|c}
    \hline
    Dataset  & Koopman-layered
    & Fourier filter (w. KDMD) & Fourier filter (w. EDMD dict.)\\
    \hline
    ETT & {\bf 0.139$\pm$0.00902} & 2.27 &  0.141$\pm$0.000942\\
    Electricity & {\bf 0.179$\pm$0.00132} & 0.731 & 0.186$\pm$0.000671 \\
    Exchange & {\bf 0.155$\pm$0.00361} &3.02 & 0.890$\pm$0.0972\\
    Traffic & {\bf 0.274$\pm$0.0186} & -- & 0.377$\pm$0.000925 \\
    Weather & 0.161$\pm$0.00484  & 0.860 & {\bf 0.107$\pm$0.000266}\\
    ILI & {\bf 0.414$\pm$0.00787} &1.15 & 0.899$\pm$0.0869\\
    \hline
    \end{tabular}\vspace{.5cm}\\
    \centering
    $\Delta T=192$\vspace{.2cm}\\
    \begin{tabular}{c|c|c|c}
    \hline
    Dataset  & Koopman-layered
    & Fourier filter (w. KDMD) & Fourier filter (w. EDMD dict.)\\
    \hline
    ETT & {\bf 0.195$\pm$0.0155}  & 2.38 & 0.223$\pm$0.00252\\
    Electricity & {\bf 0.158$\pm$0.000567} & 0.725 & 0.166$\pm$0.000848 \\
    Exchange & {\bf 0.223$\pm$0.00518} & 2.96 & 0.666$\pm$0.100\\
    Traffic & {\bf 0.246$\pm$0.0168} & -- &  0.348$\pm$0.000625 \\
    Weather & 0.188$\pm$0.00322 & 0.898 & {\bf 0.106$\pm$0.000294}\\
    ILI & {\bf 0.259$\pm$0.0109}  &0.884 & 0.911$\pm$0.311 \\
    \hline
    \end{tabular}
    \end{center}
\end{table}

\subsection{Additional results of the experiment in Subsection~\ref{subsec:exp_forecast}}\label{ap:time_forecast_additional}

\begin{figure}[t]
\setlength{\tabcolsep}{0pt}
\def\arraystretch{0.8}
\newcolumntype{C}{>{\centering\arraybackslash}X}
\begin{center}
\begin{tabularx}{\textwidth}{Ccccc}
&$\Delta t=48$ & $\Delta t=96$ & $\Delta t=144$ & $\Delta t=192$\\
\vspace{-1.5cm}ETT & \includegraphics[width=0.21\textwidth]{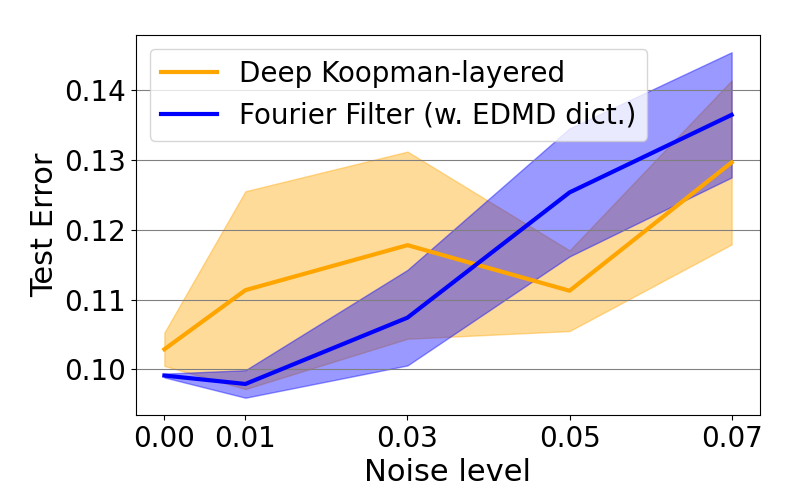} & \includegraphics[width=0.21\textwidth]{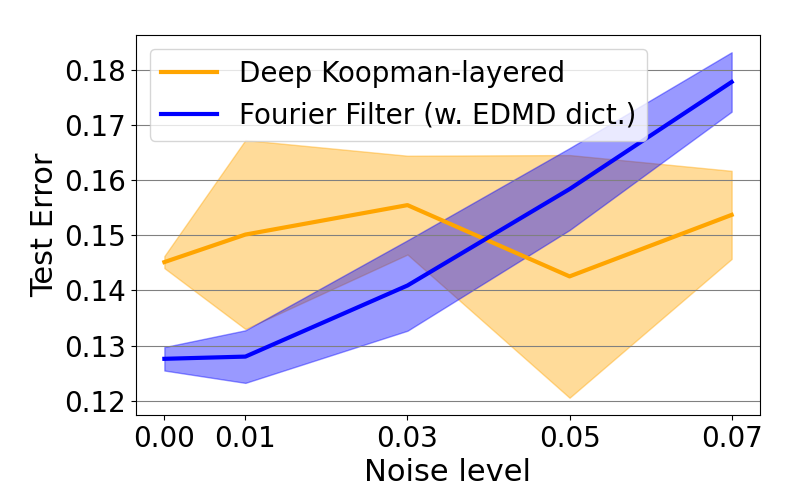} & \includegraphics[width=0.21\textwidth]{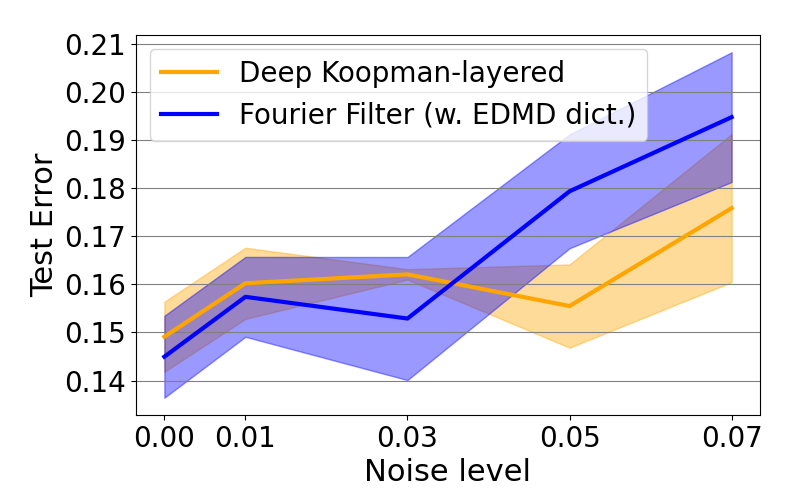} & \includegraphics[width=0.21\textwidth]{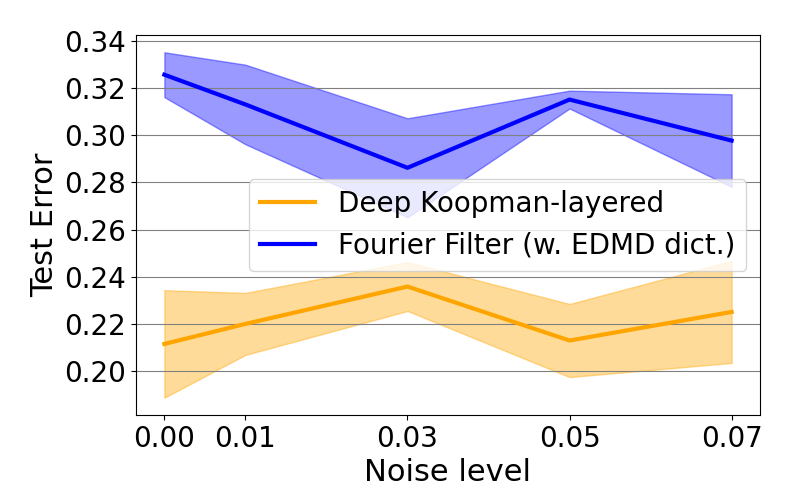}\\
\vspace{-1.5cm}Electricity & \includegraphics[width=0.21\textwidth]{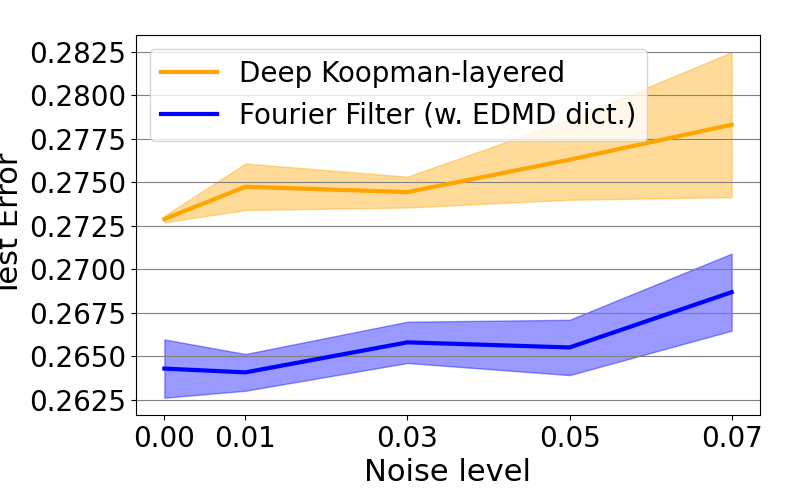} & \includegraphics[width=0.21\textwidth]{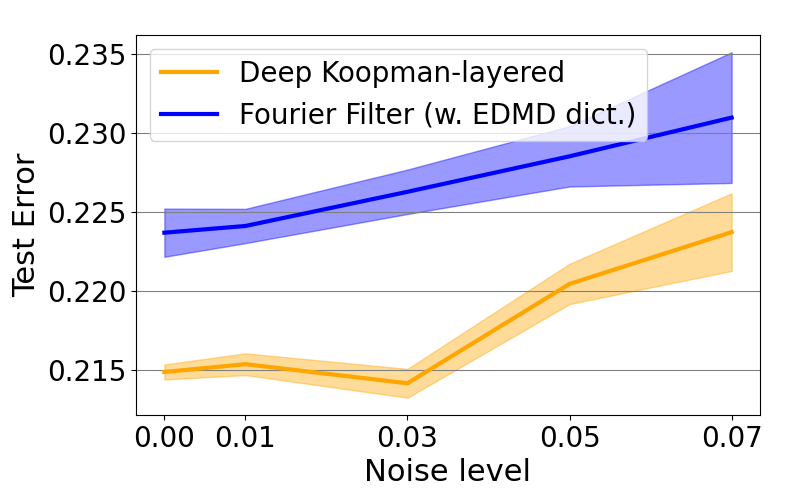} & \includegraphics[width=0.21\textwidth]{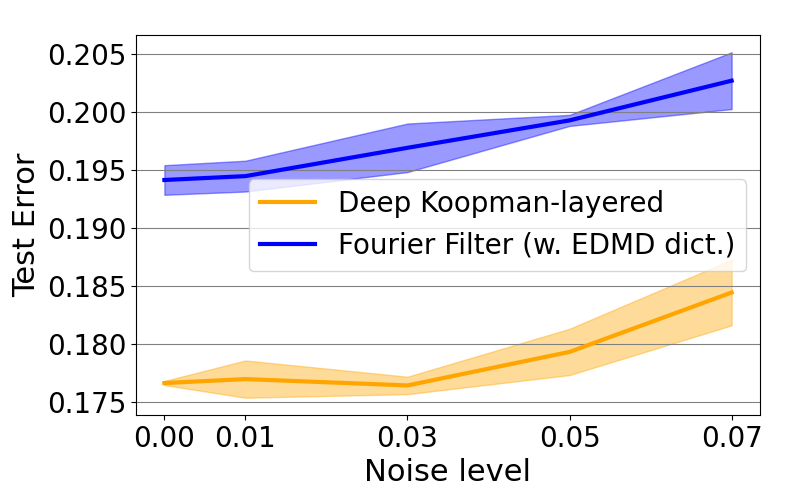} & \includegraphics[width=0.21\textwidth]{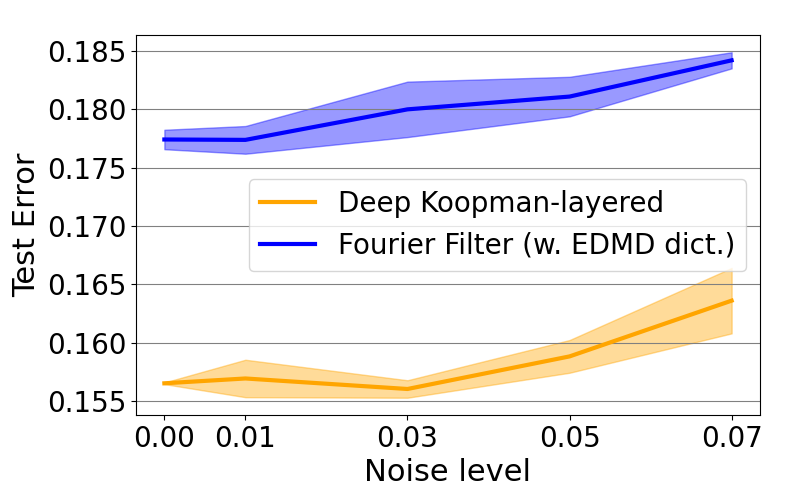}\\
\vspace{-1.5cm}Exchange & \includegraphics[width=0.21\textwidth]{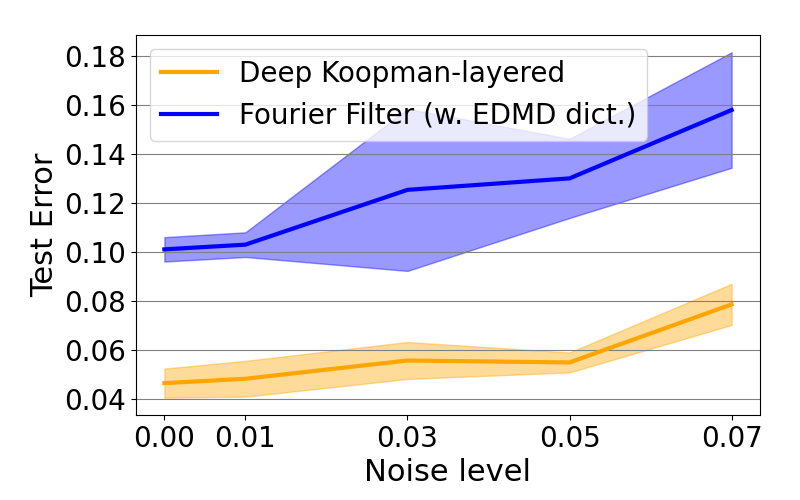} & \includegraphics[width=0.21\textwidth]{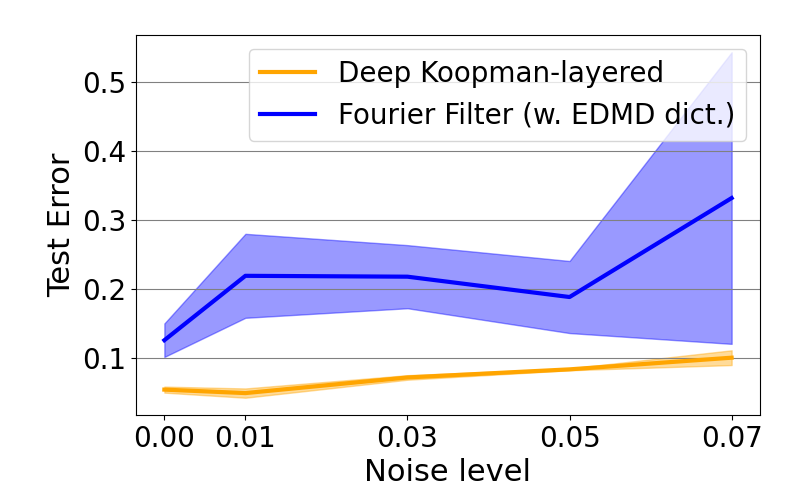} & \includegraphics[width=0.21\textwidth]{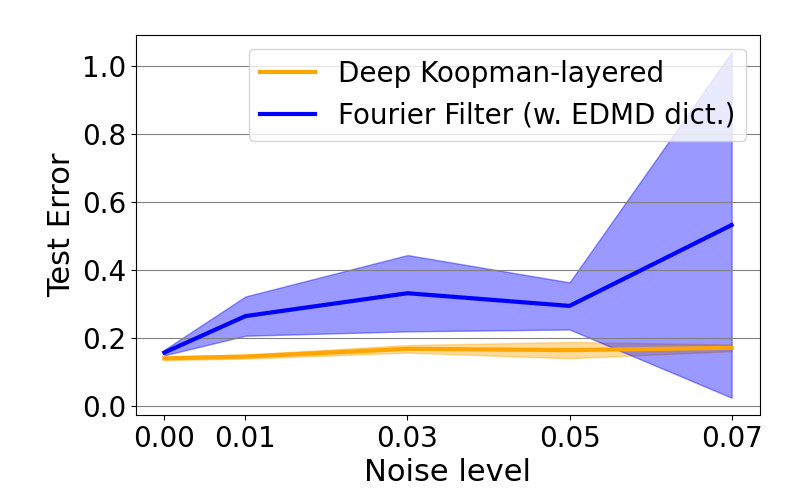} & \includegraphics[width=0.21\textwidth]{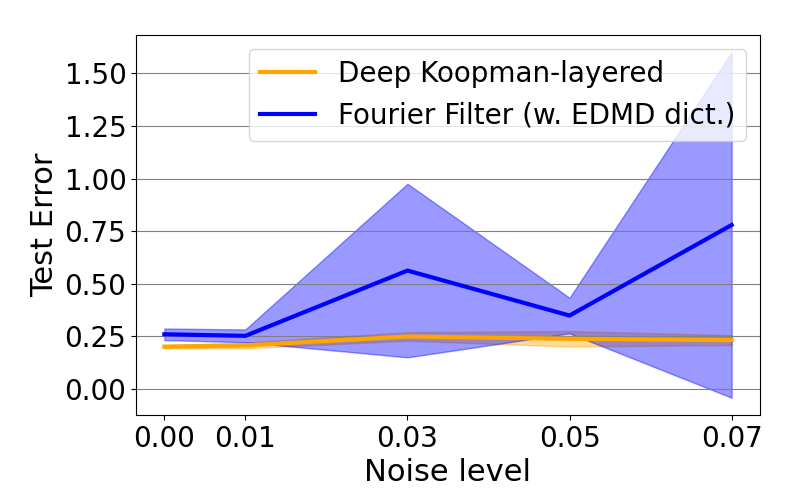}\\
\vspace{-1.5cm}Weather & \includegraphics[width=0.21\textwidth]{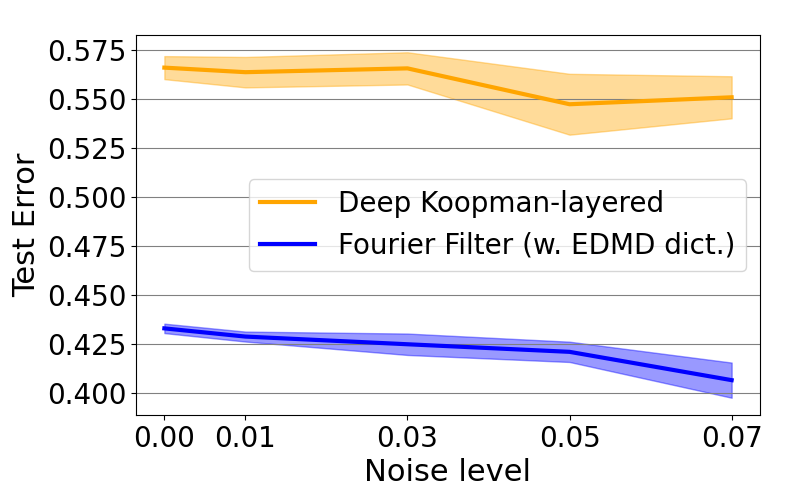} & \includegraphics[width=0.21\textwidth]{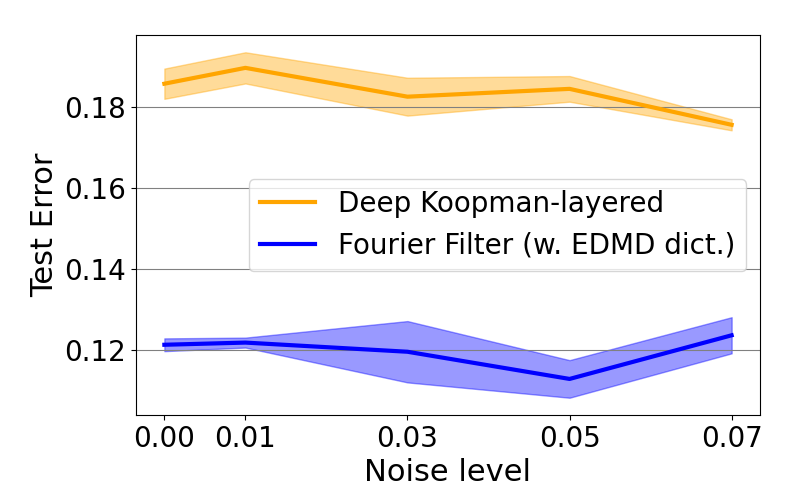} & \includegraphics[width=0.21\textwidth]{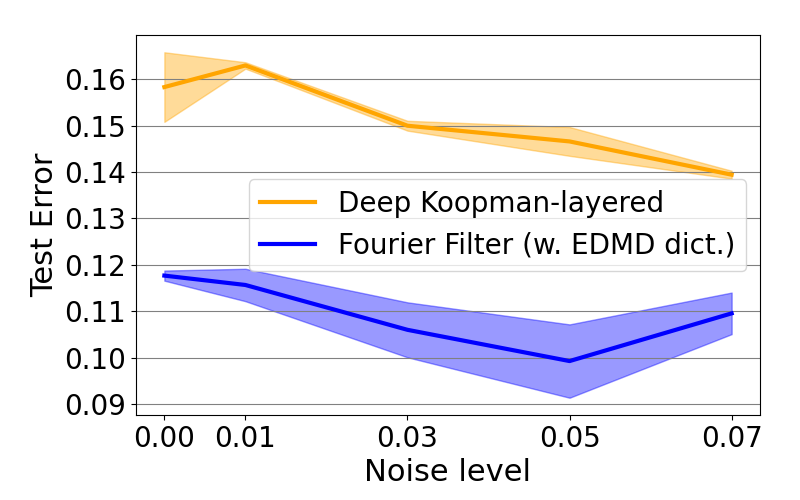} & \includegraphics[width=0.21\textwidth]{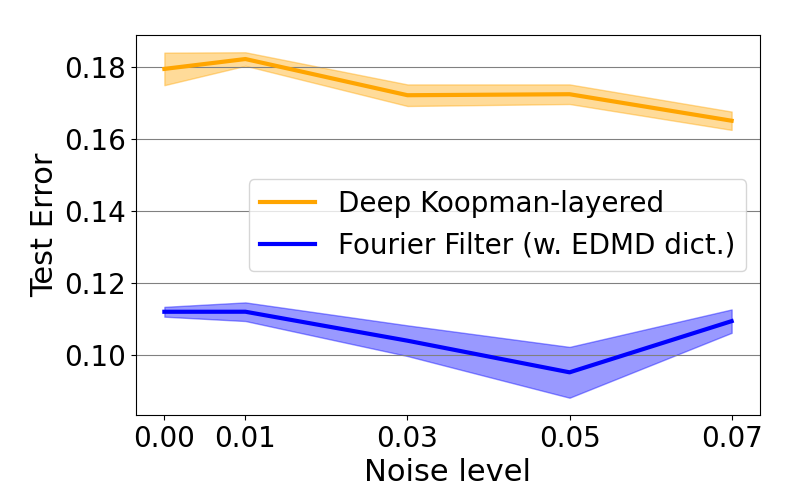} \\
\vspace{-1.5cm}ILI & \includegraphics[width=0.21\textwidth]{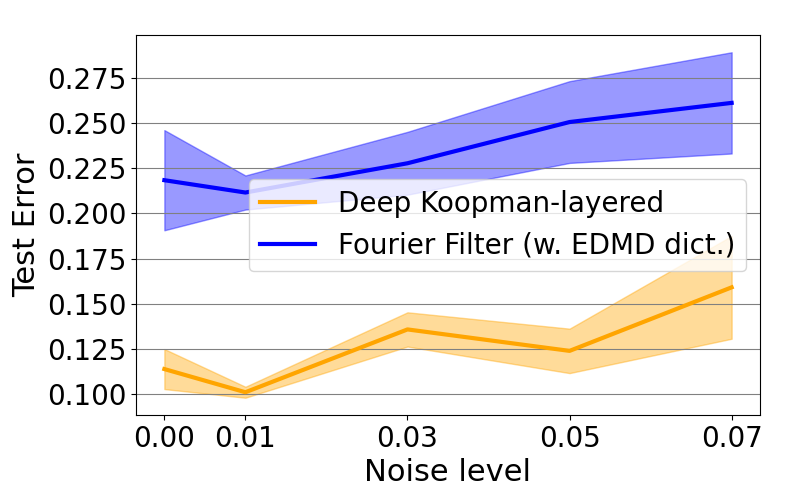} & \includegraphics[width=0.21\textwidth]{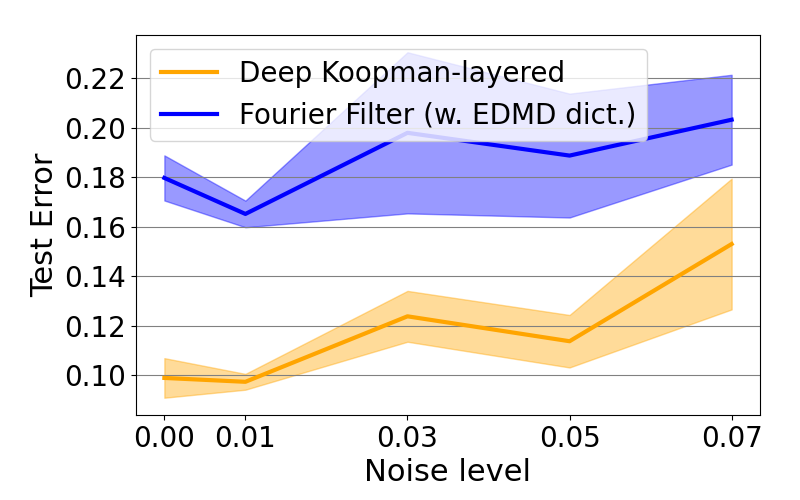} & \includegraphics[width=0.21\textwidth]{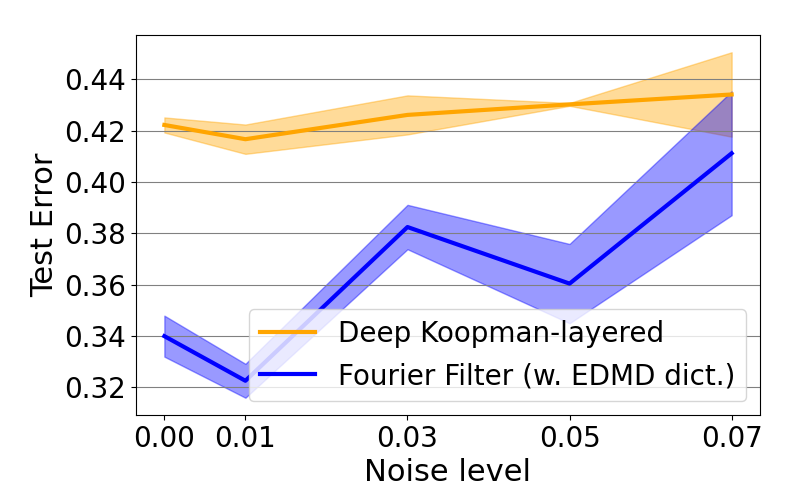} & \includegraphics[width=0.21\textwidth]{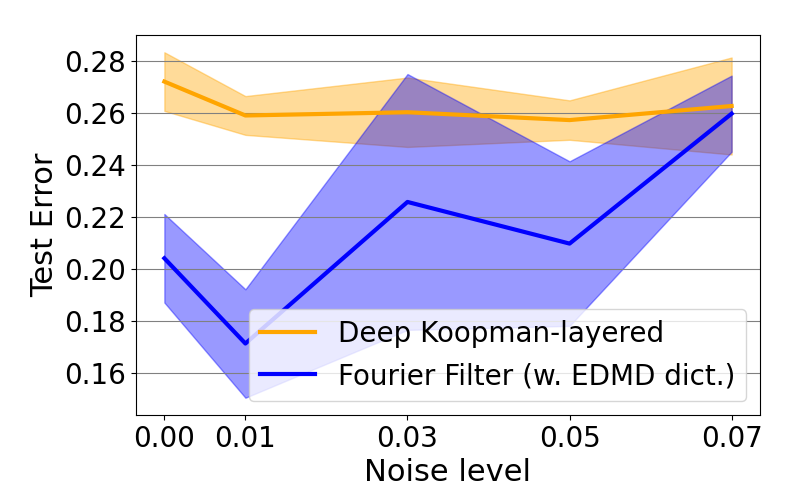}
\end{tabularx}
\end{center}
    \caption{Test error for the same time forecasting task as Subsection~\ref{subsec:exp_forecast} with the presence of noise. We set $J=5$.}
    \label{fig:noise_all}
\end{figure}

We conducted an additional experiment to investigate the sensitivity of the proposed method with respect to noise. We considered the same problem as that in Subsection~\ref{subsec:exp_forecast}, but added noise, generated from the normal distribution with mean 0 and different values of standard deviation, to the input data. 
The results are shown in Figures~\ref{fig:noise_all}.
We can see that the proposed approach is more robust than or as robust as the Fourier filter approach.
Since the model for Traffic dataset is combined with neural networks, we only conducted experiments for the other 5 datasets to investigate the properties of pure and basic Koopman-layered models.

\begin{figure}[t]
\setlength{\tabcolsep}{0pt}
\def\arraystretch{0.8}
\newcolumntype{C}{>{\centering\arraybackslash}X}
\begin{center}
\begin{tabularx}{\textwidth}{Ccccc}
&$\Delta t=48$ & $\Delta t=96$ & $\Delta t=144$ & $\Delta t=192$\\
\vspace{-1.5cm}ETT & \includegraphics[width=0.21\textwidth]{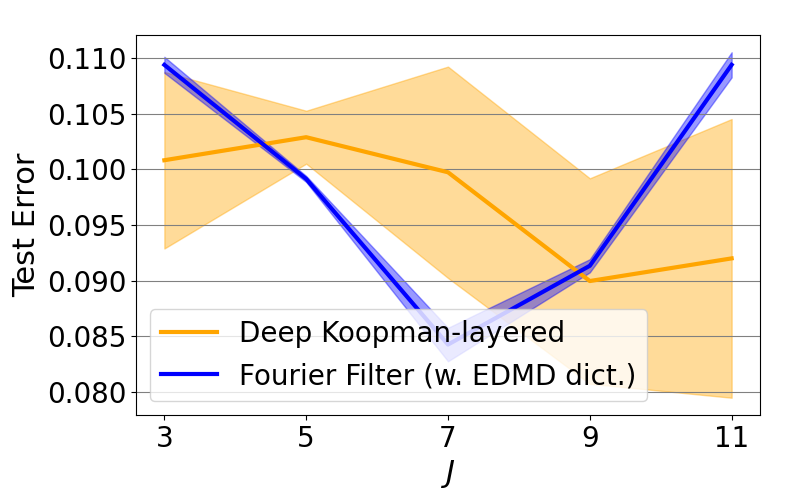} & \includegraphics[width=0.21\textwidth]{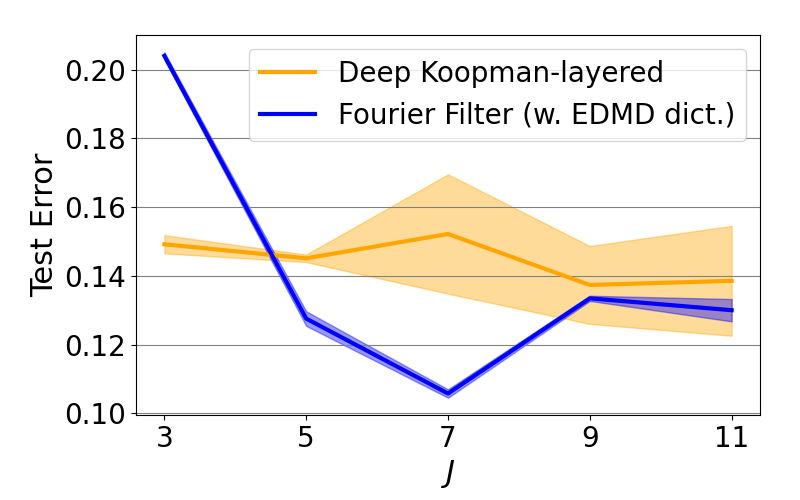} & \includegraphics[width=0.21\textwidth]{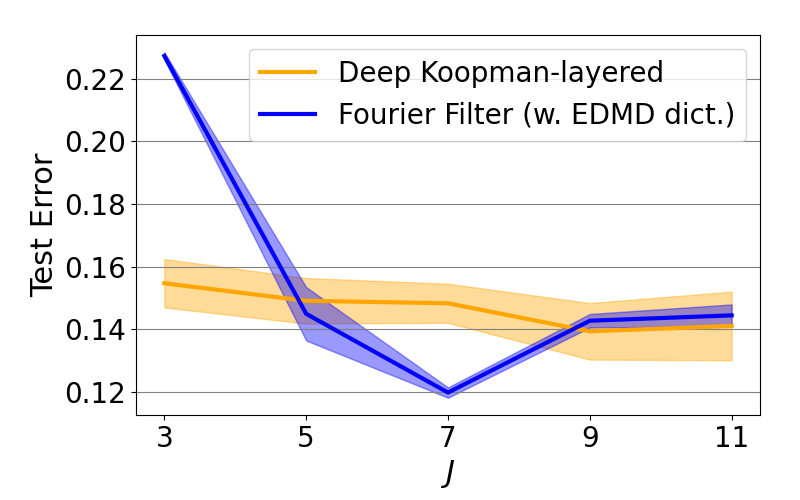} & \includegraphics[width=0.21\textwidth]{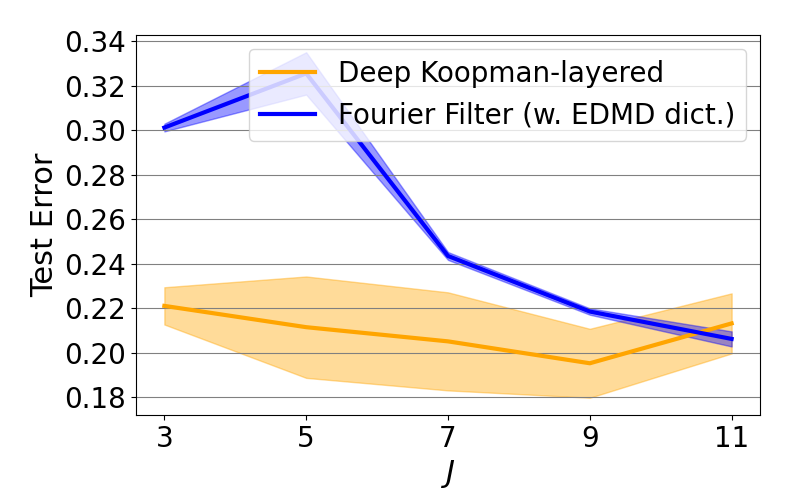}\\
\vspace{-1.5cm}Electricity & \includegraphics[width=0.21\textwidth]{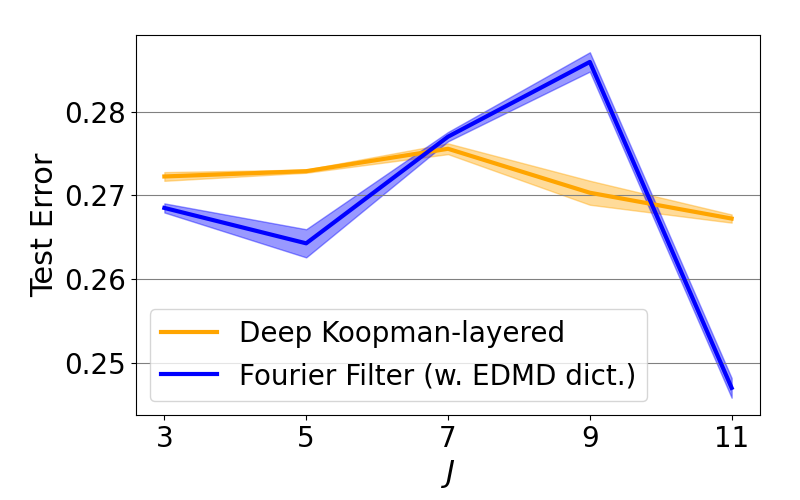} & \includegraphics[width=0.21\textwidth]{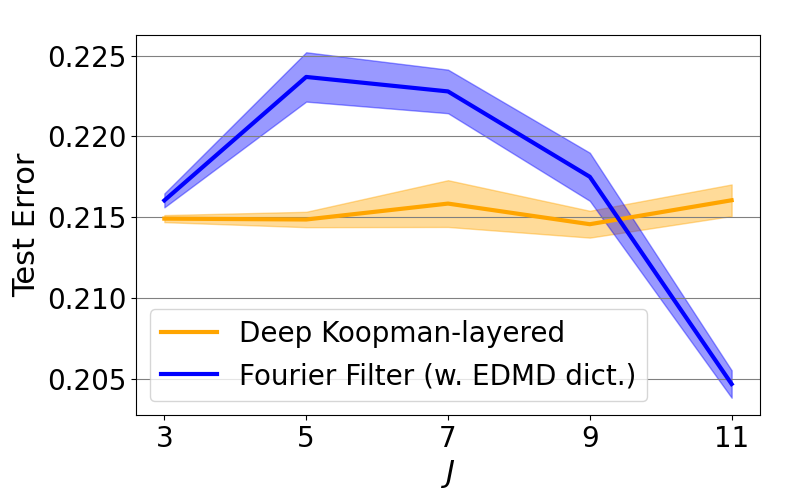} & \includegraphics[width=0.21\textwidth]{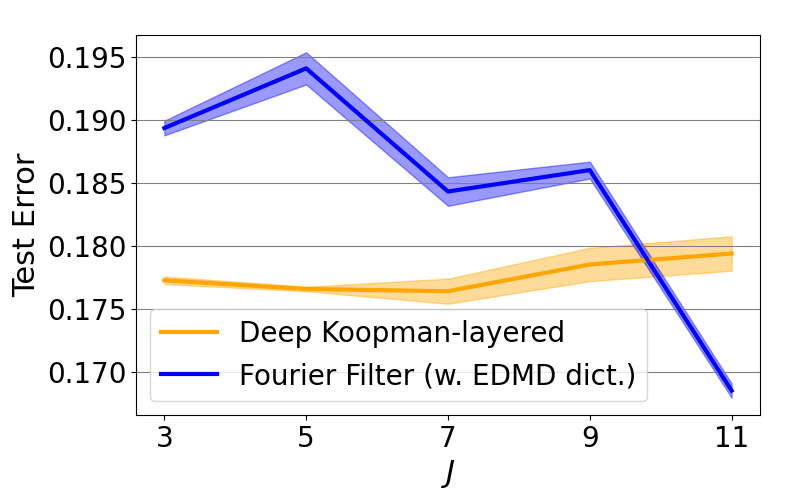} & \includegraphics[width=0.21\textwidth]{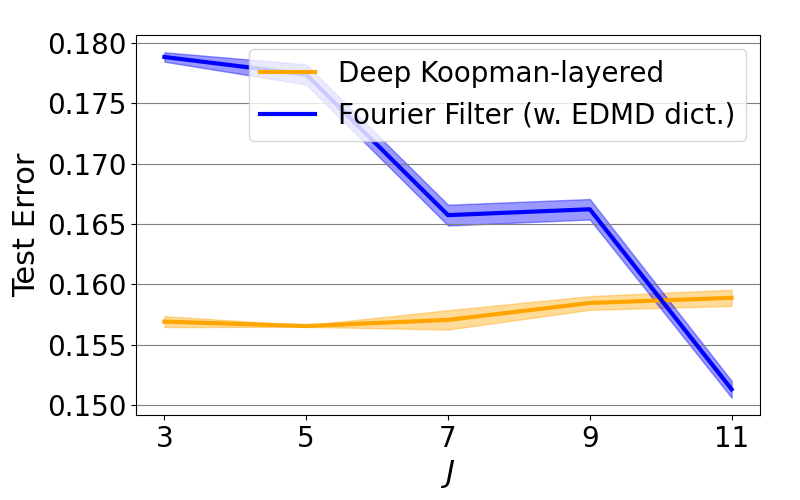}\\
\vspace{-1.5cm}Exchange & \includegraphics[width=0.21\textwidth]{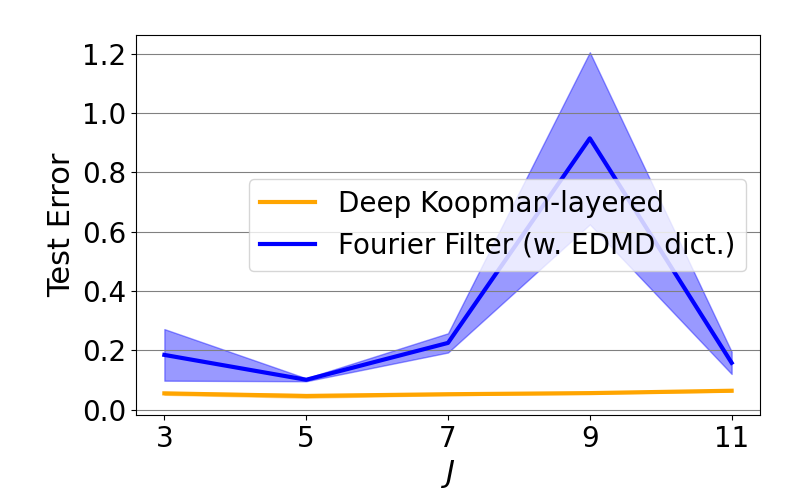} & \includegraphics[width=0.21\textwidth]{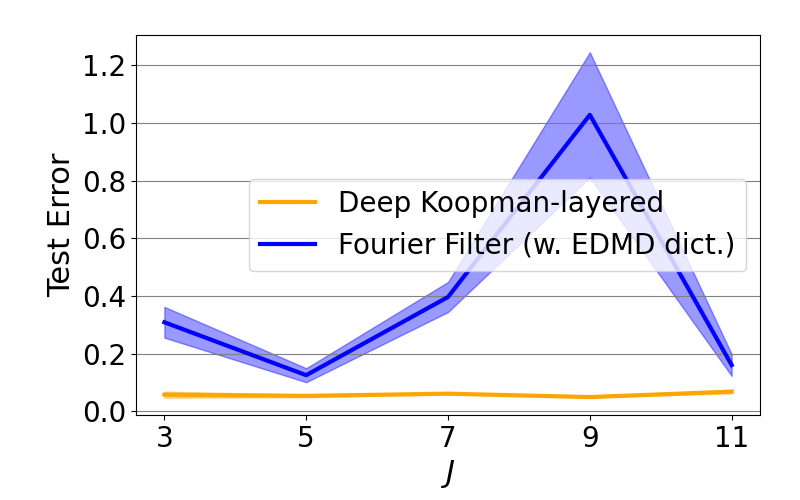} & \includegraphics[width=0.21\textwidth]{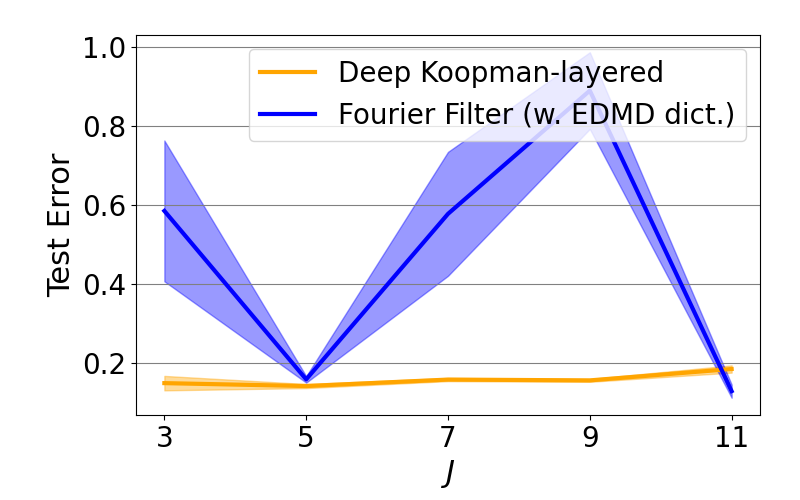} & \includegraphics[width=0.21\textwidth]{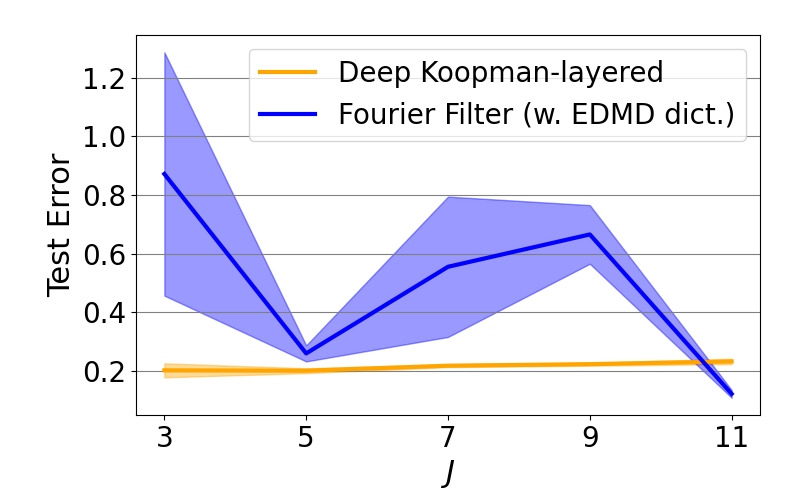}\\
\vspace{-1.5cm}Weather & \includegraphics[width=0.21\textwidth]{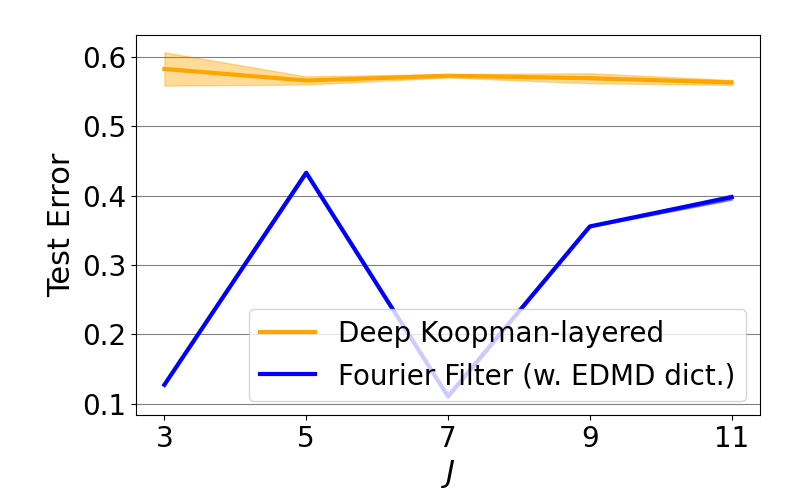} & \includegraphics[width=0.21\textwidth]{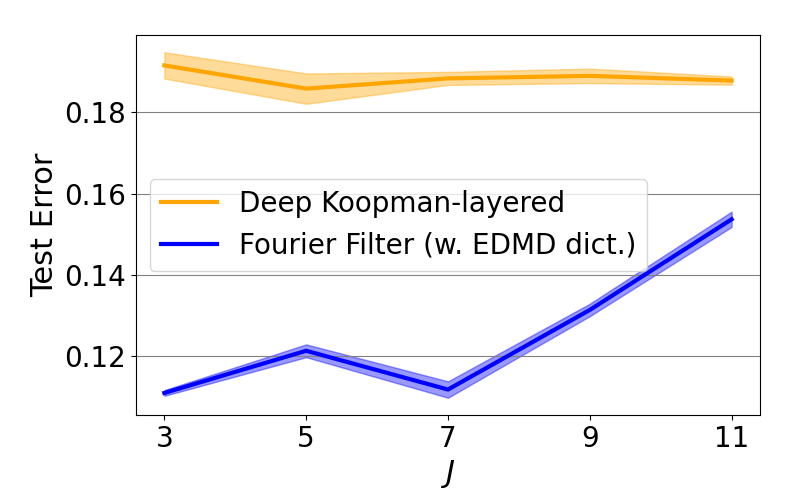} & \includegraphics[width=0.21\textwidth]{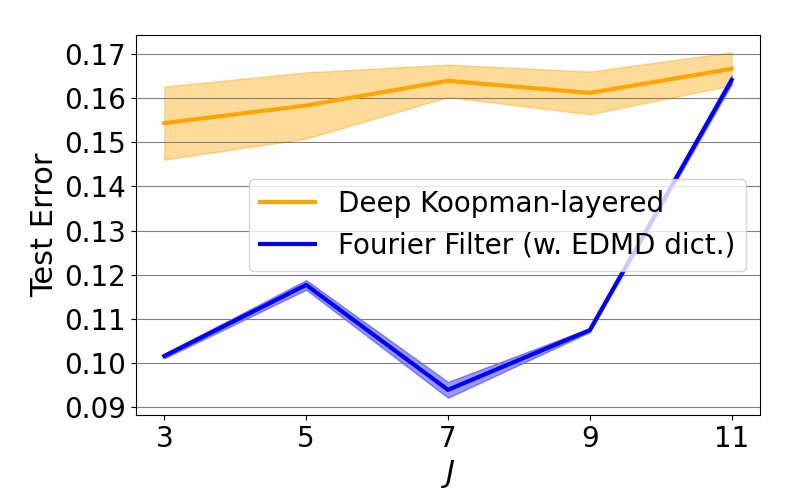} & \includegraphics[width=0.21\textwidth]{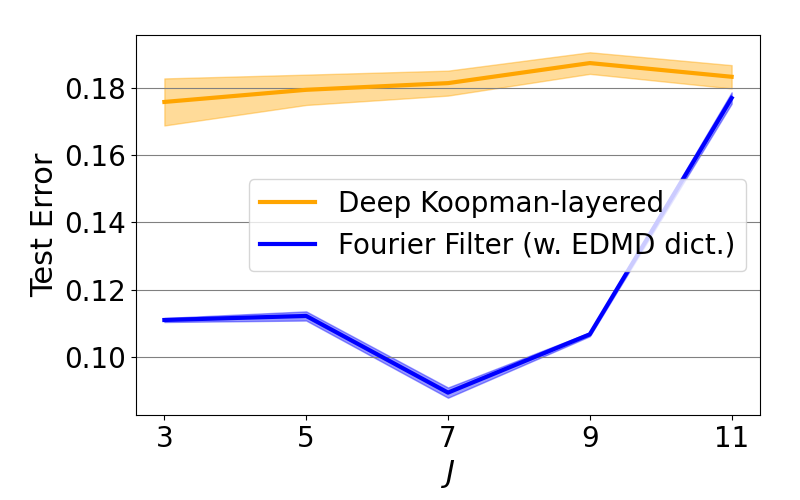} \\
\vspace{-1.5cm}ILI & \includegraphics[width=0.21\textwidth]{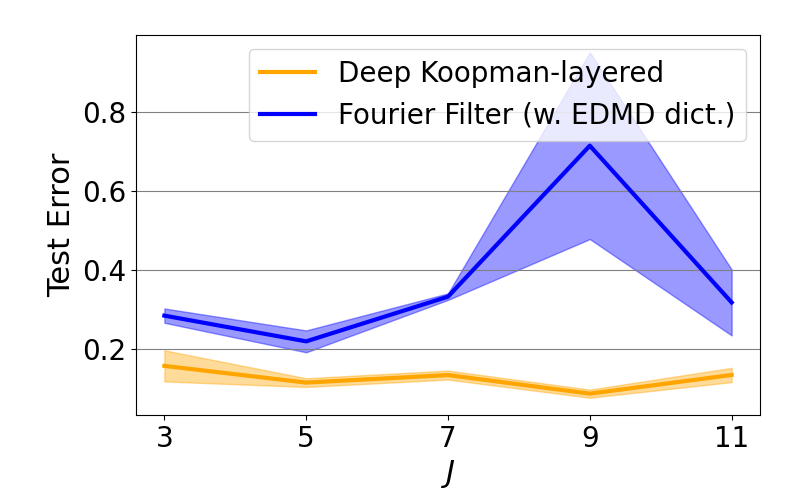} & \includegraphics[width=0.21\textwidth]{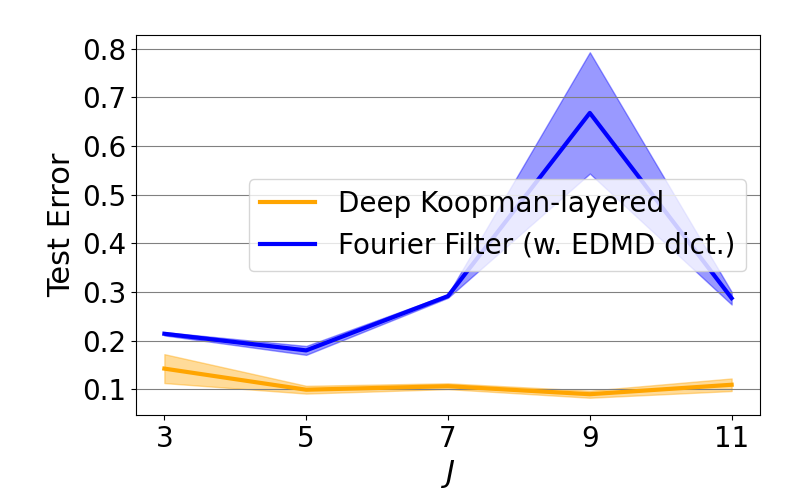} & \includegraphics[width=0.21\textwidth]{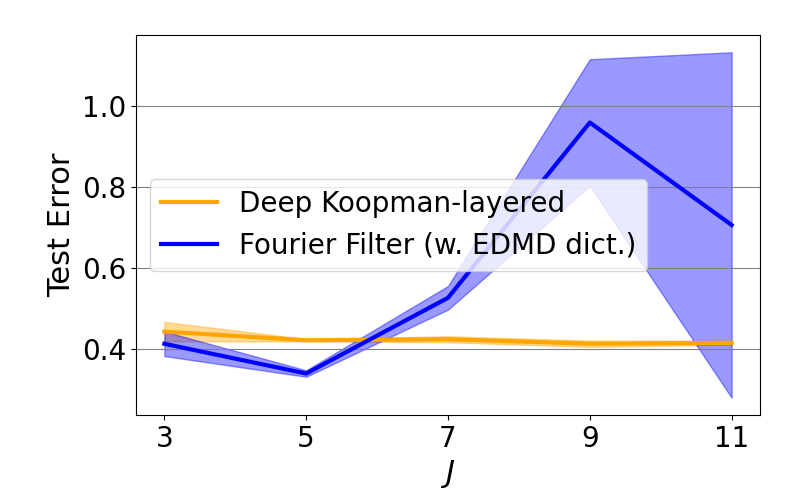} & \includegraphics[width=0.21\textwidth]{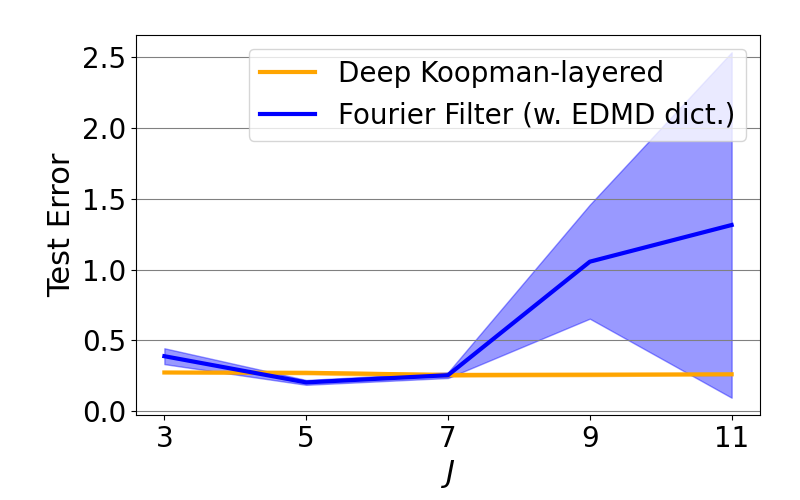}
\end{tabularx}
\end{center}
    \caption{Test error for the same time forecasting task as Subsection~\ref{subsec:exp_forecast} with different values of $J$.}
    \label{fig:J_all}
\end{figure}

In addition, we conducted an additional experiment and observed that our proposed method is robust with respect to the choice of $J$. 
Theoretically, if $J$ is large, then the model can represent more functions. However, in practice, there is a tradeoff between $J$ and the number of samples necessary to extract local information of each Koopman layer. 
Indeed, to train a $J$-layered model, we split the dataset into $J+1$ sub-datasets. 
Then, we train the model so that the $j$th layer transforms the $(j-1)$th sub-dataset into $j$th sub-dataset. 
When $J$ is large, the size of each sub-dataset becomes small, which makes the approximation of the corresponding Koopman operator more difficult. 
In addition, if $J$ is large, the model can easily overfit to training data due to the large amount of learnable parameters.
On the other hand, when $J$ is small, since we try to represent the dynamics with a small number of Koopman operators, we cannot capture the nonautonomous dynamics well.
The empirical results are shown in Figure~\ref{fig:J_all}. 
We computed the test error with different values of $J$. 
To obtain the best performance, $J$ should not be too small or too large. For comparison, we also conducted an experiment with the approach with the combination of the Fourier filter and EDMD. 
The Fourier filter was applied to each sub-dataset, and we extracted time-invariant features by averaging the value of the Fourier components (See Appendix~\ref{ap:exp_detail_forecast} for more details). 
The proposed approach is more robust with respect to the choice of $J$. 
This is because even with an inappropriate choice of $J$, since the proposed model learns all $J$ Koopman layers simultaneously, the layers share their local information. 
On the other hand, since the Fourier filter is applied separately to each sub-dataset, the model cannot share the local information, which results in the performance significantly depending on the choice of $J$.

\begin{figure}[t]
\setlength{\tabcolsep}{0pt}
\def\arraystretch{0.8}
\newcolumntype{C}{>{\centering\arraybackslash}X}
\begin{center}
\begin{tabularx}{\textwidth}{Ccccc}
&$\Delta t=48$ & $\Delta t=96$ & $\Delta t=144$ & $\Delta t=192$\\
\vspace{-1.5cm}ETT & \includegraphics[width=0.21\textwidth]{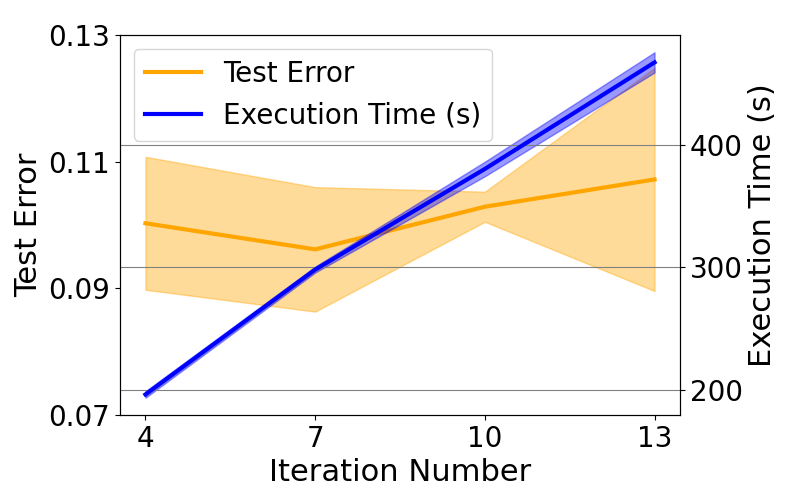} & \includegraphics[width=0.21\textwidth]{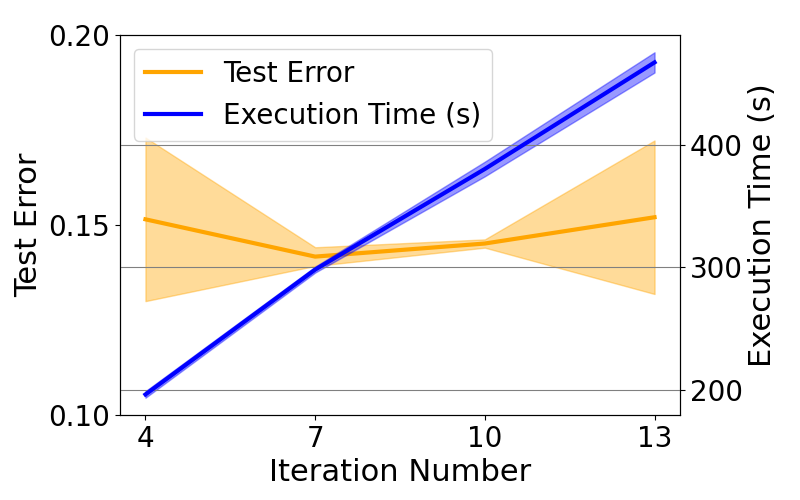} & \includegraphics[width=0.21\textwidth]{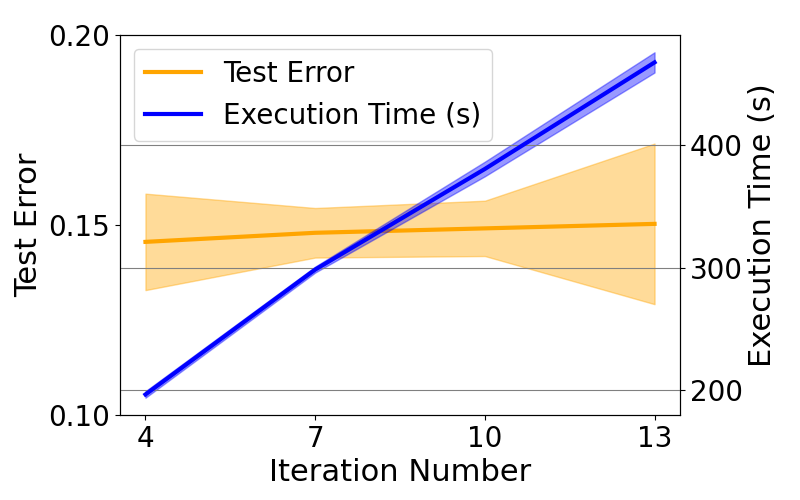} & \includegraphics[width=0.21\textwidth]{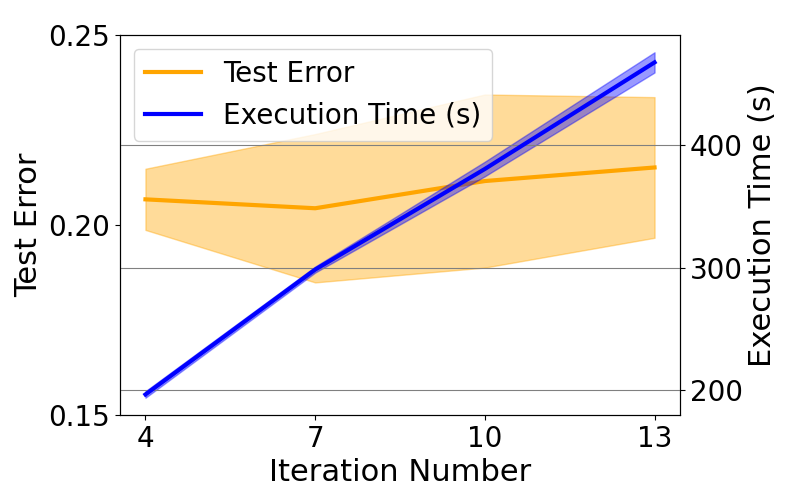}\\
\vspace{-1.5cm}Electricity & \includegraphics[width=0.21\textwidth]{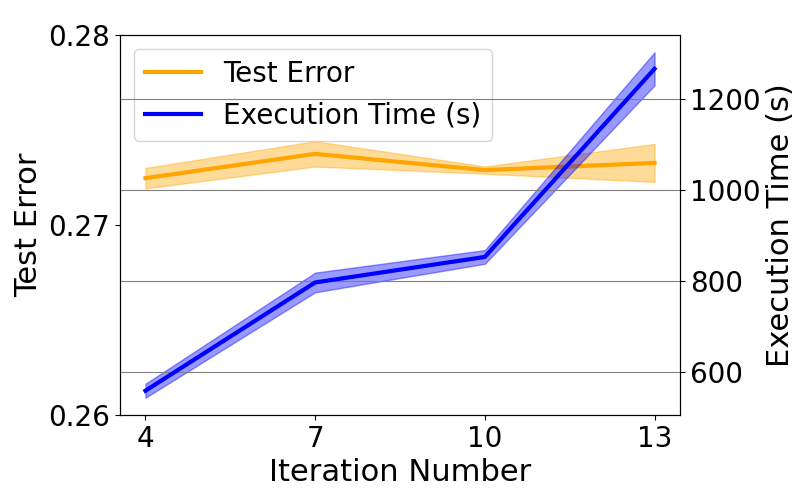} & \includegraphics[width=0.21\textwidth]{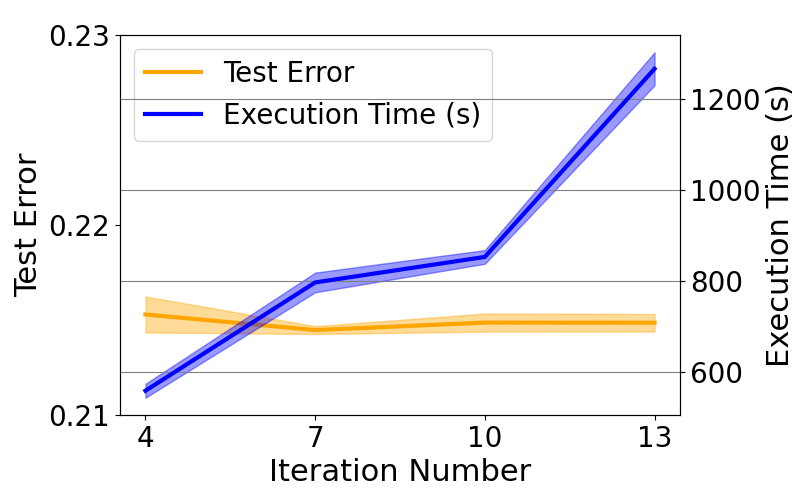} & \includegraphics[width=0.21\textwidth]{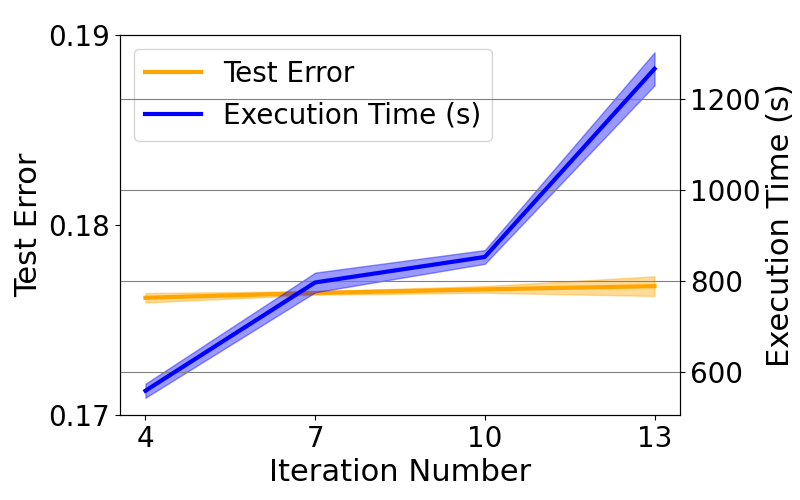} & \includegraphics[width=0.21\textwidth]{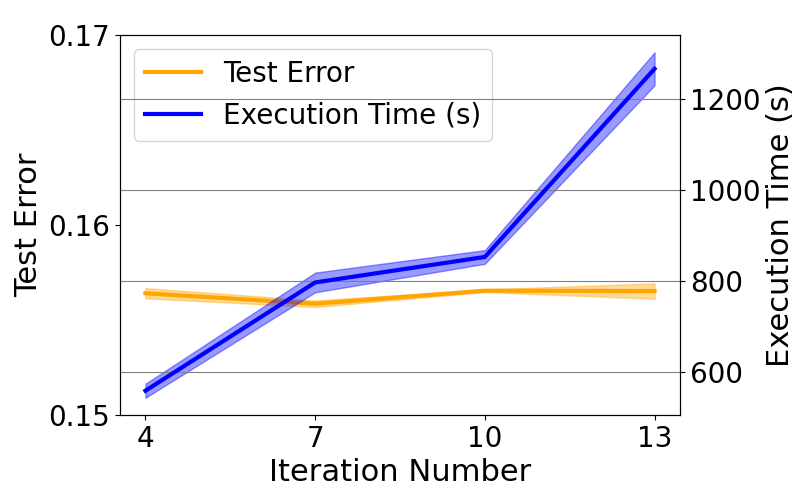}\\
\vspace{-1.5cm}Exchange & \includegraphics[width=0.21\textwidth]{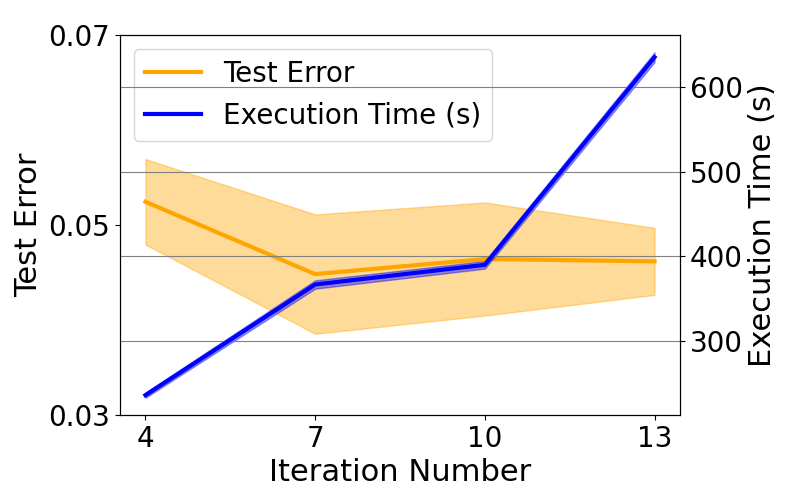} & \includegraphics[width=0.21\textwidth]{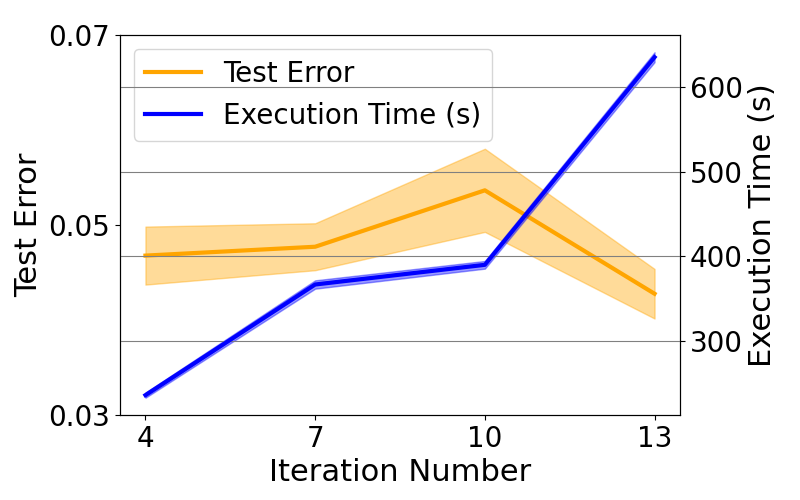} & \includegraphics[width=0.21\textwidth]{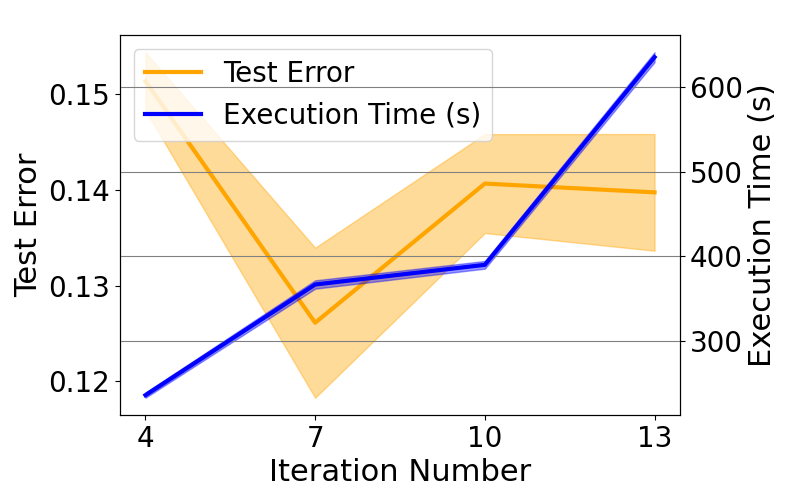} & \includegraphics[width=0.21\textwidth]{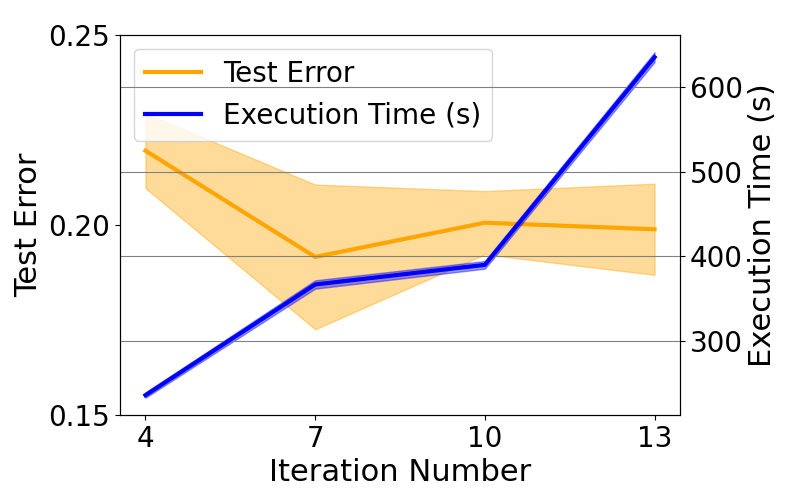}\\
\vspace{-1.5cm}Weather & \includegraphics[width=0.21\textwidth]{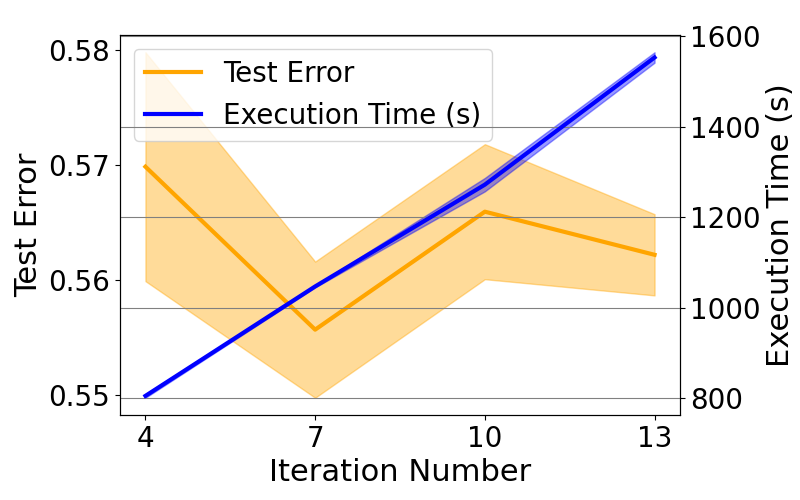} & \includegraphics[width=0.21\textwidth]{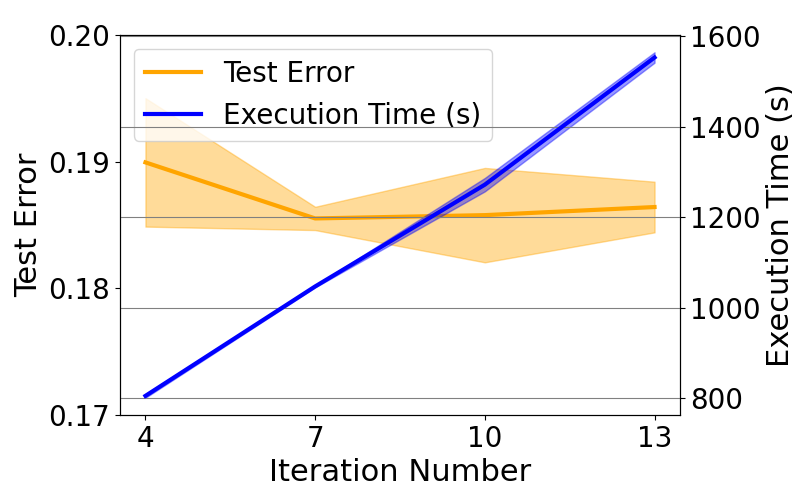} & \includegraphics[width=0.21\textwidth]{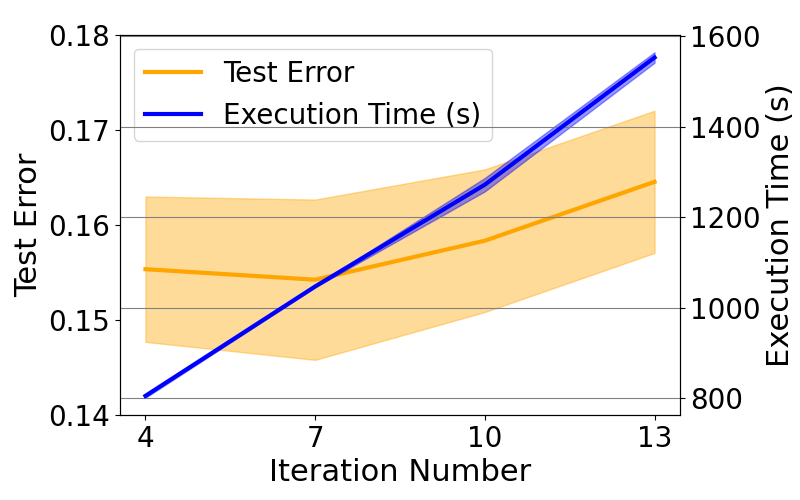} & \includegraphics[width=0.21\textwidth]{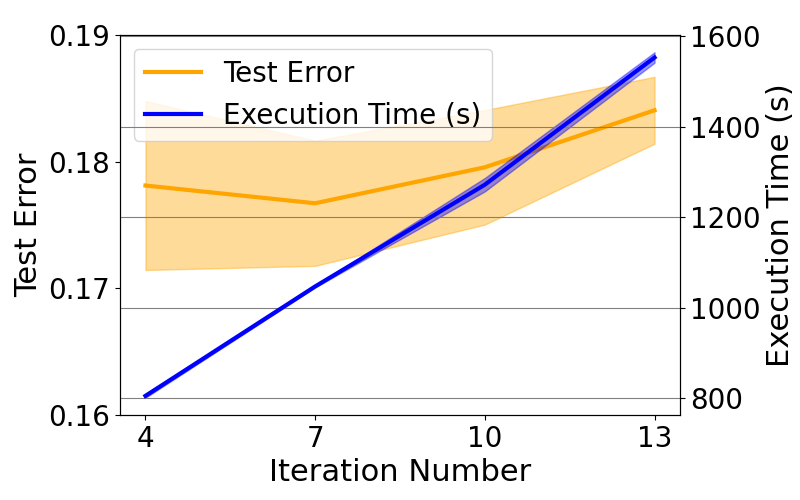} \\
\vspace{-1.5cm}ILI & \includegraphics[width=0.21\textwidth]{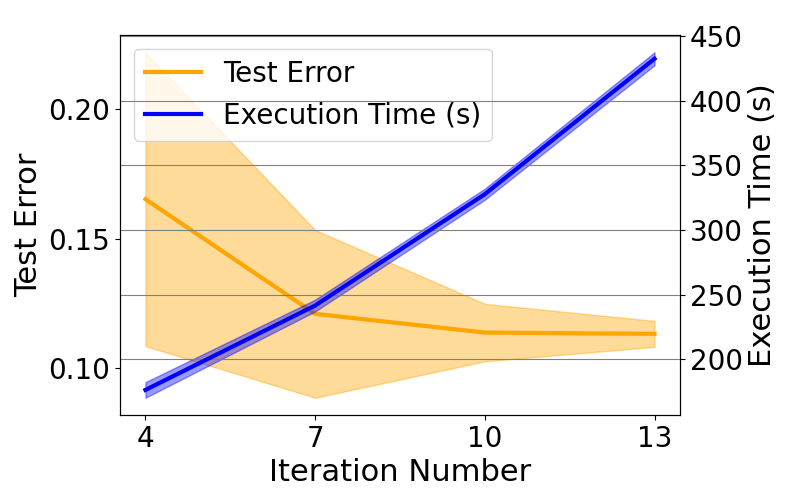} & \includegraphics[width=0.21\textwidth]{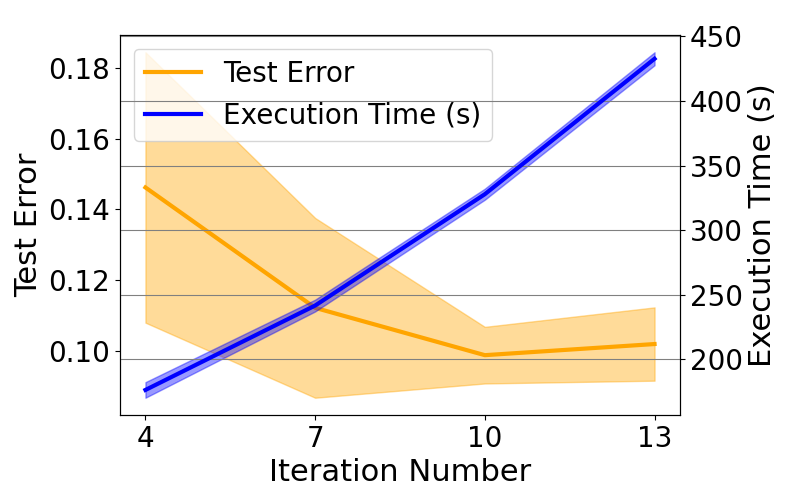} & \includegraphics[width=0.21\textwidth]{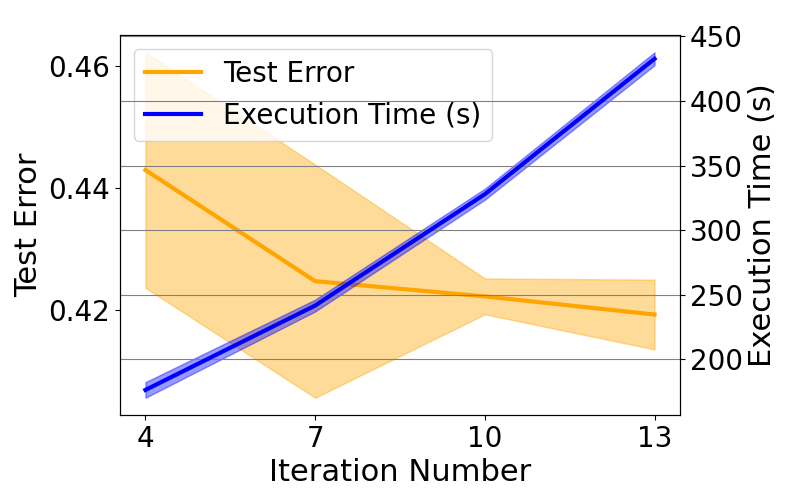} & \includegraphics[width=0.21\textwidth]{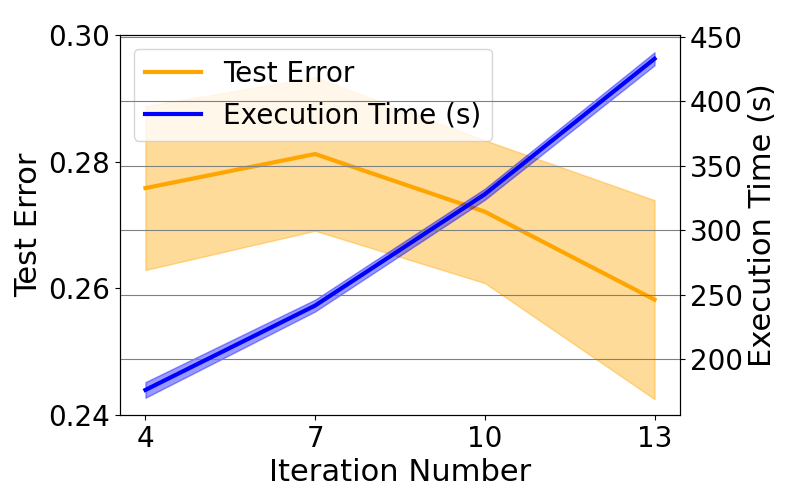}
\end{tabularx}
\end{center}
    \caption{Execution time and test error for the same time forecasting task as Subsection~\ref{subsec:exp_forecast} with the deep Koopman-layered approach with different values of the iteration number of Krylov subspace method.}
    \label{fig:itr_all}
\end{figure}

Furthermore, we investigated the relationship among the number of Krylov iterations, execution time, and test error for the proposed approach. 
The results are shown in Figure~\ref{fig:itr_all}. 
While the execution time increases linearly with respect to the iteration number, the test error does not change so much, which implies that the Krylov approximation of the matrix exponential is reasonable even with small iteration numbers, and a proper choice of the iteration number can reduce the computational cost.

\end{document}